\documentclass[pdflatex]{sn-jnl}%

\usepackage{graphicx}%
\usepackage{multirow}%
\usepackage{amsmath,amssymb,amsfonts}%
\usepackage{amsthm}%
\usepackage{mathrsfs}%
\usepackage[title]{appendix}%
\usepackage{xcolor}%
\usepackage{textcomp}%
\usepackage{manyfoot}%
\usepackage{booktabs}%
\usepackage{algorithm}%
\usepackage{algorithmicx}%
\usepackage{algpseudocode}%
\usepackage{makecell}%
\usepackage{listings}%
\usepackage[numbers]{natbib}
\usepackage{subcaption}
\usepackage{rotating}
\usepackage{tikz}
\usepackage{svg}
\usepackage{microtype}
\usetikzlibrary{shapes,arrows,positioning,backgrounds,fit}

\usepackage{pgfplots}
\pgfplotsset{compat=1.18} %

\definecolor{safetyColor}{HTML}{0d6a82}
\definecolor{biodiversitycolor}{HTML}{fb4d3d}
\definecolor{3dmodelColor}{HTML}{345995}
\definecolor{applicationColor}{HTML}{e7901a}
\definecolor{backgroundColor}{HTML}{2a9d8f}
\definecolor{challengeColor}{HTML}{606c38}
\definecolor{applicationBackground}{HTML}{ffb703}

\tikzset{
    base node/.style={
            minimum width=1cm,
            align=center,
            minimum height=0.75cm,
            inner sep=5pt
        },
    category box/.style={
            rectangle,
            draw=black,
            inner sep=5pt
        },
    arrow/.style={
            -latex,
            ultra thick,
            opacity=0.9
        },
    category passopti/.style={
             draw=blue!40, thick, rounded corners, fill=blue!2
        },
    3dmodels/.style={base node, fill=white, draw=black},
    biodiversity/.style={base node, fill=biodiversitycolor!5, draw=biodiversitycolor},
    safety/.style={base node, fill=safetyColor!5, draw=safetyColor},
    applications/.style={base node, fill=applicationColor!5, draw=applicationColor},
    title/.style={minimum height=0cm, align=center},
    optimized/.style={dashed, draw=safetyColor, thick},
    optimizedbox/.style={optimized, fill=safetyColor!5},
    templ/.style={
            align=left,
            font=\footnotesize,
            execute at begin node=\makebox[\linewidth][l]{\normalsize #1},
        },
    templdraw/.style={templ={#1},
        }
}

\newlength{\simimgw}     
\newlength{\simcolsep}   
\newlength{\simlabw}     
\newlength{\simlabsep}   
\newlength{\simgroupsep} 
\newlength{\simrowsep}   

\newlength{\simspanw}   
\newlength{\simblockw}  

\newcommand{\simrowgap}{\noalign{\vskip\simrowsep}}
\newcommand{\simheadgap}{\noalign{\vskip4pt}}

\newcommand{\simimgs}[3]{\mbox{%
    \includegraphics[width=\simimgw]{#1}\hspace{\simcolsep}%
    \includegraphics[width=\simimgw]{#2}\hspace{\simcolsep}%
    \includegraphics[width=\simimgw]{#3}}}
\newcommand{\simfreqs}[3]{\mbox{\small
    \makebox[\simimgw]{#1}\hspace{\simcolsep}%
    \makebox[\simimgw]{#2}\hspace{\simcolsep}%
    \makebox[\simimgw]{#3}}}
\newcommand{\simbar}[1]{\includegraphics[width=\simspanw]{#1}}

\newcommand{\simblock}[2]{%
    \begin{minipage}[t]{\simblockw}
        \centering
        #1\\[3pt]
        #2
    \end{minipage}%
}

\newcommand{\simheader}[2]{%
    \makebox[\simlabw]{}\hspace{\simlabsep}%
    \makebox[\simblockw]{#1}\hspace{\simgroupsep}%
    \makebox[\simblockw]{#2}%
}

\newsavebox{\simrowbox}
\newsavebox{\simlabbox}

\newcommand{\simscenerow}[2]{%
    \sbox{\simrowbox}{#2}%
    \sbox{\simlabbox}{\rotatebox[origin=c]{90}{\small #1}}%
    \makebox[\simlabw]{%
        \raisebox{\dimexpr 0.5\ht\simrowbox-0.5\dp\simrowbox
            -0.5\ht\simlabbox+0.5\dp\simlabbox\relax}[0pt][0pt]{%
            \usebox{\simlabbox}}}%
    \hspace{\simlabsep}%
    \usebox{\simrowbox}%
}

\newlength{\simpairsep}    
\newlength{\simpairw}      
\newlength{\simpairblockw} 

\newcommand{\simpairimgs}[2]{\mbox{%
    \includegraphics[width=\simimgw]{#1}\hspace{\simcolsep}%
    \includegraphics[width=\simimgw]{#2}}}
\newcommand{\simpairlabels}[2]{\mbox{\footnotesize
    \makebox[\simimgw]{#1}\hspace{\simcolsep}%
    \makebox[\simimgw]{#2}}}
\newcommand{\simpairtop}[2]{\mbox{\small
    \makebox[\simimgw]{#1}\hspace{\simcolsep}%
    \makebox[\simimgw]{#2}}}
\newcommand{\simpairblock}[2]{%
    \begin{minipage}[t]{\simpairblockw}
        \centering
        #1\\[3pt]
        #2
    \end{minipage}%
}
\newcommand{\simpairheader}[3]{%
    \makebox[\simlabw]{}\hspace{\simlabsep}%
    \makebox[\simpairblockw]{#1}\hspace{\simpairsep}%
    \makebox[\simpairblockw]{#2}\hspace{\simpairsep}%
    \makebox[\simpairblockw]{#3}%
}

\usepackage[capitalise]{cleveref}
\usepackage{siunitx}
\usepackage{mathtools}

\newcommand{\minimize}{\operatornamewithlimits{minimize}}
\DeclareMathOperator*{\argmin}{arg\,min}
\newcommand{\norm}[1]{\left\lVert#1\right\rVert_2}
\newcommand{\hatnabla}{\hat{\nabla}}

\theoremstyle{thmstyleone}%

\theoremstyle{thmstyletwo}%

\usepackage{comment}

\theoremstyle{thmstylethree}%

\newcommand{\methodname}{FORGE-SIM}

\begin{document}

\title[Spline-Based Boundary Representations for Sparse View Reconstruction and Simulation Using Isogeometric Analysis]{Spline-Based Boundary Representations for Sparse View Reconstruction and Simulation Using Isogeometric Analysis}

\author*[1,2]{\fnm{Davor} \sur{Dobrota}}\email{davor.dobrota@epfl.ch}

\author[1]{\fnm{Vsevolod} \sur{Skorokhodov}}\email{vsevolod.skorokhodov@epfl.ch}

\author[3]{\fnm{Chenghao} \sur{Xu}}\email{chenghao.xu@epfl.ch}

\author[3]{\fnm{Olga} \sur{Fink}}\email{olga.fink@epfl.ch}

\author*[1]{\fnm{Malcolm} \sur{Mielle}}\email{forge-sim@mielle.dev}

\affil[1]{\orgdiv{Schindler Lab EPFL}, \orgname{Schindler AG}, \orgaddress{\street{EPFL Innovation Park}, \city{Lausanne}, \postcode{1015}, \state{Vaud}, \country{Switzerland}}}

\affil[2]{\orgdiv{Chair of Numerical Modelling and Simulation}, \orgname{École polytechnique fédérale de Lausanne}, \orgaddress{\street{Station 8}, \city{Lausanne}, \postcode{1015}, \state{Vaud}, \country{Switzerland}}}

\affil[3]{\orgdiv{Intelligent Maintenance and Operating Systems Lab}, \orgname{École polytechnique fédérale de Lausanne}, \orgaddress{\street{Station 18}, \city{Lausanne}, \postcode{1015}, \state{Vaud}, \country{Switzerland}}}

\abstract{
    Image-based reconstruction aims to recover three-dimensional geometry from images.
    Recent advances have enabled the recovery of visually detailed models, yet their representations are not well-suited for numerical simulation.
    Simulation frameworks typically require explicit, watertight, and smooth geometries to ensure numerical robustness and accuracy, properties that surfaces extracted from image-based reconstructions lack.
    We propose \methodname{}, a method to directly reconstruct a multi-patch B-spline boundary representation from sparse posed RGB images without manual intervention.
    By optimizing the spline representation itself, our approach produces compact, smooth, and watertight geometries that are natively compatible with both Computer Aided Design and simulation workflows.
    Additionally, we introduce a strategy to project observation-derived fields, such as a thermal state and semantic information, onto the reconstructed models in the same spline basis, enabling immediate use in simulation.
    We demonstrate that the obtained models are of sufficiently high quality to enable thermal simulation and modal analysis.
    By unifying image-based reconstruction and simulation-ready modeling within a single optimization framework, this work removes a long-standing barrier between computer vision and numerical analysis.
    We anticipate that it will enable new workflows for simulation-driven design, inspection, and digital twin applications.
}


\maketitle

\section{Main}

Digital twins are increasingly used to understand, predict, and optimize the behavior of physical systems across engineering, manufacturing, energy, and infrastructure applications.
By enabling physics-based numerical simulation directly on virtual representations of real-world assets, they support tasks ranging from structural assessment~\cite{hadiIntegrityRevitStructural2021,liuVibrationKLShells2022} and heat-transfer analysis~\cite{chassaingThermoxelsVoxelBasedMethod2025} to design optimization~\cite{alvesWindTurbineDesign2025} and predictive maintenance~\cite{henrikMappingDamages2023}.
Despite their growing importance, constructing simulation-ready digital twins of existing physical objects remains largely a manual process.
Engineers typically reconstruct Computer Aided Design (CAD) models from measurements, such as 3D scans, generate conformal meshes that approximate the model, and subsequently perform numerical simulation.
These manual steps are a major bottleneck: they require time and specialized expertise, and introduce opportunities for errors at each step between measurement and the final simulation result.

Recent advances in computer vision and deep learning have made significant progress in obtaining 3D representations of an object from images.
Methods based on Neural Radiance Fields~\cite{hassanThermoNeRFMultimodalNeural2025}, Signed Distance Functions~\cite{sun3QFPEfficientNeural2024a}, Gaussian splatting~\cite{kerbl3DGaussianSplatting2023}, and voxel-based representations~\cite{chassaingThermoxelsVoxelBasedMethod2025} can now recover detailed 3D geometry from sparse image sets captured with ordinary cameras.
While these techniques have proven highly effective in applications where visual fidelity matters most---ranging from autonomous navigation~\cite{sunHighFidelitySLAMUsing2024} to virtual reality~\cite{jiangVRGSPhysicalDynamicsAware2024}---reconstructing appearance and enabling physically consistent reasoning are fundamentally different objectives.
A representation that accurately reproduces image observations may still be unsuitable for numerical simulation, which requires watertight manifold geometry, topological structure, and numerical regularity to support the solution of the governing physical equations.
On the other hand, vision-based methods primarily optimize for photometric accuracy and rendering quality, often using representations that do not satisfy the aforementioned requirements.
An exception is Thermoxels~\cite{chassaingThermoxelsVoxelBasedMethod2025}, which produces voxel-derived tetrahedral meshes with thermal fields directly from images; however, it does not provide the smooth, CAD-compatible boundary representation that downstream modeling and shell-based analysis require, leading to simulations that quickly diverge from the ground truth.

The finite element method (FEM)~\cite{allaireNumericalAnalysisOptimization2007} is the most common approach to the numerical solution of partial differential equations (PDEs) on complex domains.
Converting outputs from computer vision pipelines into models suitable for finite element analysis (FEA) requires extensive post-processing in which every step demands manual expert intervention: segmenting reconstructed geometry into meaningful regions, fitting spline surface patches to these segments \cite{eckAutomaticReconstruction1996, zhuSurfaceReconstrucionPC2022}, repairing topology through sewing and healing operations to achieve watertightness~\cite{chongMeshHealing2007}, and performing simulation-specific modifications to generate valid meshes~\cite{varadayReverseEngineeringCAD1997}.
Furthermore, triangle meshes, especially those obtained from vision-based methods, often contain non-manifold edges, inconsistent face orientations, and small gaps that prevent direct use in simulation.
Even when topology can be repaired, the resulting geometry may lack the smoothness and parametric control that CAD models provide~\cite{karmanMeshSimulationChallenges2017}.
Industrial practice therefore continues to rely on manual CAD modeling despite automated reconstruction capabilities, because vision methods cannot deliver the geometric quality that simulation frameworks expect.

This limitation reflects a deeper incompatibility in how these fields represent geometry.
As FEM and CAD matured together, meshing boundary representation (B-rep) models became standard practice.
These models are built from both simple primitives, such as spheres and cylinders, and free-form surfaces, typically represented by Non-Uniform Rational B-Splines (NURBS)~\cite{pieglNURBSBook1997}.
B-rep solid models incorporate detailed topological information---in the form of faces, edges, and vertices---to ensure piecewise manifoldness with gaps below a specified geometric tolerance, a crucial requirement for precise imposition of boundary conditions and efficacy of meshing algorithms.
Furthermore, spline-based primitives provide flexible continuity control across surface patches and compact parametric descriptions of free-form surfaces.
Modern simulation frameworks, particularly isogeometric analysis (IGA)~\cite{hughesIsogeometricAnalysisCAD2005a}, exploit these properties by using geometric basis functions to represent the discrete solution of the PDE, improving accuracy, robustness, and computational efficiency.

To bridge this gap, we propose a reconstruction paradigm in which simulation compatibility is enforced throughout the reconstruction process rather than recovered through post-processing.
Instead of reconstructing geometry in a vision-oriented representation and subsequently converting it into a simulation model, we directly optimize a spline-based boundary representation from sparse image observations.
This enables three capabilities that are not achieved simultaneously by existing approaches: (i) reconstruction of watertight, simulation-ready geometries directly from images, (ii) compatibility with both isogeometric and conventional finite-element analysis, and (iii) recovery of observation-derived fields---a semantic material-class field, an initial thermal state, or, in general, any single- or multi-channel field---directly on the reconstructed representation.

We introduce \methodname{} (\underline{F}ree-form \underline{O}ptimized \underline{R}econstruction of \underline{G}eometry for \underline{E}ngineering \underline{Sim}ulation), a framework for reconstructing simulation-ready digital twins directly from sparse multi-modal images.
Unlike existing reconstruction approaches that prioritize rendering fidelity, \methodname{} jointly enforces geometric smoothness, topological validity, watertightness, and compatibility with downstream numerical simulation throughout the reconstruction process.
By combining differentiable image-based reconstruction with spline-based geometric modeling, the method directly optimizes closed, watertight boundary representations---following the notion of regular solids in the mathematical theory of CAD~\cite{requichaRigidSolids1980}---from observations while maintaining compatibility with both isogeometric analysis and conventional finite-element workflows.

The central idea is to couple image-based reconstruction with a simulation-native geometric representation.
To make optimization tractable, \methodname{} employs a dual-representation strategy in which a spline-based boundary representation is coupled to an auxiliary mesh and bitmap texture used for efficient differentiable rendering.
Image observations provide gradients through the rendering process, while the spline representation maintains the geometric regularity, continuity, and topological validity required for simulation.
Combined with coarse-to-fine refinement and geometry-aware regularization, this enables shape optimization directly from sparse image observations while preserving simulation compatibility throughout the reconstruction process.
The final reconstructed model can be exported to standard CAD formats (STEP, IGES~\cite{trimmingReviewMaraussig2018}) and integrated seamlessly with established mesh simulation tools---avoiding the complex and error-prone multi-step conversion pipelines commonly used in current practice.

Beyond geometry reconstruction, \methodname{} provides a unified representation for both geometry and simulation-relevant quantities.
Spatially varying fields---a semantic material-class field, an initial thermal state, and in general any single- or multi-channel field that can be supervised through the renderer---are represented in the same spline basis as the geometry itself, with physical parameters then assigned from these recovered fields. This yields simulation-ready models directly from observations.
The resulting representation therefore serves not only as a geometric model, but also as a computational model for downstream physics-based analysis and simulation.

\methodname{} advances the field by establishing a direct pathway from image observations to simulation-ready digital twins.
By unifying image-based reconstruction, CAD-quality geometric modeling, and simulation-compatible representation learning within a single framework, it bypasses the manual CAD reconstruction and geometry-repair workflows that currently separate computer vision from computational simulation.
To the best of our knowledge, \methodname{} is the first pipeline to produce simulation-ready boundary representations from images without manual geometry authoring, repair, or meshing at any stage: segmentation (from a per-scene text prompt), pose estimation, shape and texture optimization, and field inpainting all run automatically, with per-scene inputs reduced to the refinement schedule and physical material parameters.

Since the resulting representation is compatible with both IGA and conventional FEM workflows, the framework provides a general approach for constructing simulation-compatible digital twins of existing physical objects.
These capabilities enable practical workflows where physical objects can be digitized and analyzed without manual CAD reconstruction, with immediate relevance to applications such as reverse engineering, quality control, and structural health monitoring where simulation on as-built geometry is essential.
More broadly, the framework enables physical objects to be reconstructed, enriched with simulation-relevant information, and analyzed through physics-based models directly from observations, opening new opportunities for scalable digital-twin construction across engineering and scientific domains.
The complete pipeline is illustrated in \cref{fig:flowchart}.

\section{Results}

\input{images/flowchart.tex}

The central hypothesis of \methodname{} is that B-spline-based boundary representations can be used for both shape reconstruction and simulation, enabling a complete image-to-simulation pipeline.
If this hypothesis holds, we should be able to obtain realistic 3D reconstructions that yield physically consistent, close-to-ground-truth simulations, while also being compatible with classic computer vision tasks such as novel-view synthesis (NVS).
We test this hypothesis through a series of experiments that evaluate (i) the ability to perform thermal and modal simulations on \methodname{}'s reconstructions, evaluated both against ground-truth simulations and on real-world scenes (\cref{results:heatSimulation,results:modalAnalysis}), and (ii) the performance of \methodname{} for geometric reconstruction and novel-view synthesis (\cref{sec:results:nvs}).

To support controlled evaluation, we use three datasets.
The first dataset includes four textureless meshes traditionally used in computer vision.\footnote{\url{https://github.com/alecjacobson/common-3d-test-models}}
The second dataset includes four realistic building structures (from BuildNet3D~\cite{xuExploitingSemanticScene2025b}) used to evaluate \methodname{}'s ability to reconstruct more complex scenes while removing the influence of real-world data-collection errors such as pose inaccuracies or sensor noise.
Finally, the third dataset comprises a compilation of RGB and thermal images of four real-world scenes: Building A~\cite{hassanThermoNeRFMultimodalNeural2025}, a plushie lion~\cite{thermalmix}, a car, and a woodshed; the car and woodshed are new $360^\circ$ RGB+thermal scenes captured for this paper using the ThermoScenes~\cite{hassanThermoNeRFMultimodalNeural2025} collection setup.
The procedure used to pre-process the real-world scenes and obtain camera poses is detailed in \cref{implementation:datasets}.

\subsection{Simulation}

To evaluate whether the boundary representations reconstructed via \methodname{} are suitable for downstream physical simulations, we consider two distinct tasks: heat-flow simulation and modal analysis.
Together, these tasks demonstrate the applicability of our framework to simulation problems with different modeling requirements.
Heat-flow analysis supports thermography-based condition assessment of concrete
infrastructure~\cite{hiasaExperimentalNumericalStudies2018} and is also relevant in quality assessment as thermal simulations can quantify how defects such as porosity affect component performance and expected lifetime~\cite{evansTransientThermalFinite2015}.
On the other hand, shell-based modal analysis is essential to model the curved, thin-walled structures
encountered in biological~\cite{houShellElementsforCellMechanics2018},
automotive~\cite{zhangLightweightDesignAutomobile2006}, and civil-engineering
systems~\cite{blaauwendraadStructuralShellAnalysis2014}.

\subsubsection{Heat-Flow Simulation}\label{results:heatSimulation}

We first validate the simulation fidelity of the \methodname{} pipeline by comparing IGA-based simulations on our reconstructed boundary representation against FEM-based simulations on ground-truth triangle meshes.
This controlled experiment removes the influence of pose estimation, sensor noise, and texture reconstruction to evaluate the core pipeline of our method.
Heat simulation is performed using spatially varying material properties, as detailed in \cref{tab:scene_thermal_materials}.
For each object, the left side is heated with a constant heat source, while the right (less conductive) side has simple diffusion.
Thermal and semantic information is mapped to vertices via texture coordinates and spatial coordinates---establishing a checkerboard initial temperature distribution and partitioning the domain into aluminum and steel.
This provides a challenging scenario for simulation with sharp material and heat distribution changes.

We quantify end-to-end fidelity through four complementary metrics (\cref{tab:e2e_summary}).
The \emph{initial-energy error} compares the total thermal energy $E_0 = \int_\mathcal{M} d\,C\,u_0\,dS$ at $t=0$ between the reconstructed and ground-truth models, jointly measuring the recovered surface area and the inpainted initial field quality.
Here, $\mathcal{M}$ denotes the manifold (reconstructed or ground truth), $d$ is the shell thickness, $C$ the volumetric heat capacity, and $u_0$ the initial temperature field (see \cref{implementation:heat}).
The \emph{heat generation-rate error} compares the total source power $\int_\mathcal{M} f_M\,dS$, which is proportional to the recovered source-region area, between ground-truth and \methodname{} simulations.
Since the total energy is conserved by construction (Crank--Nicolson scheme) and is therefore blind to heat diffusion, we additionally report the \emph{field discrepancy}, i.e., the area-weighted relative $L^2$ difference between the two temperature fields transferred across surfaces by closest-point projection.
We report it at the start and final simulation time (${FL}^2_{t_0}$ and ${FL}^2_{t_N}$ respectively); sharp-pattern transfer artifacts are removed by a floor correction---see Supplementary, \cref{sup_mat:endtoendheat}.
Finally, the \emph{relaxation-rate error} compares the smallest non-zero eigenvalue of the heat transfer operator $\lambda_1 = 1/T_\text{relax}$---the geometry-intrinsic slowest equilibration rate (Laplace--Beltrami), computed from the spectrum of the assembled operators.

\begin{figure}[t]
    \centering
    \setlength{\tabcolsep}{0pt}
    \begin{tabular}{@{}l@{}}
        \simpairheader
        {\simpairtop{\textbf{GT initial}}{\textbf{Ours initial}}}
        {\simpairtop{\textbf{GT final}}{\textbf{Ours final}}}
        {\simpairtop{\boldmath$f_\text{GT}$}{\boldmath$f_\text{Ours}$}}      \\ \simheadgap
        \simscenerow{Suzanne}{%
            \simpairblock
            {\simpairimgs
                {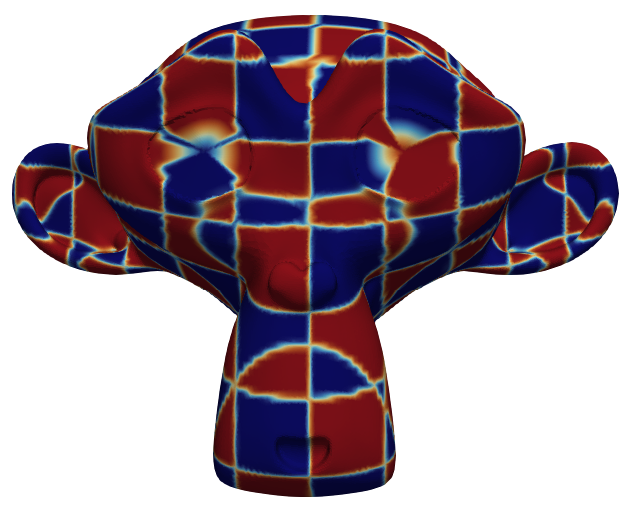}
                {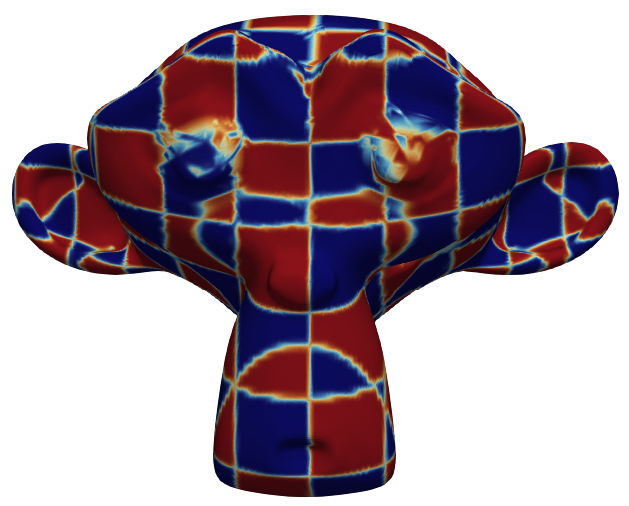}}
            {\simpairlabels{$262.6$ \si{\kilo\joule}}{$262.8$ \si{\kilo\joule}}}%
            \hspace{\simpairsep}%
            \simpairblock
            {\simpairimgs
                {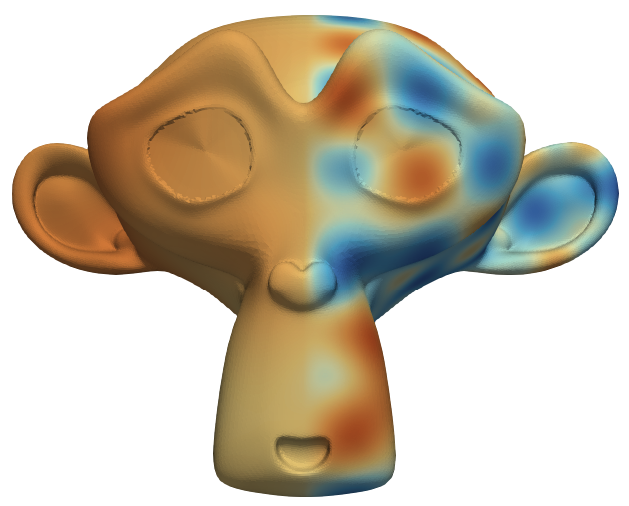}
                {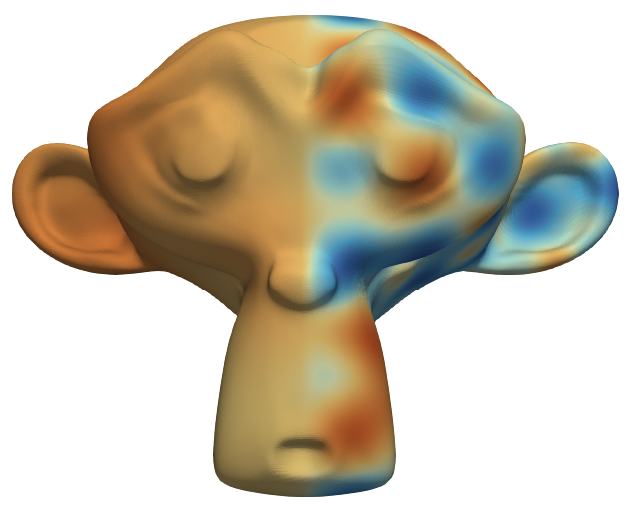}}
            {\simpairlabels{$292.3$ \si{\kilo\joule}}{$291.8$ \si{\kilo\joule}}}%
            \hspace{\simpairsep}%
            \simpairblock
            {\simpairimgs
                {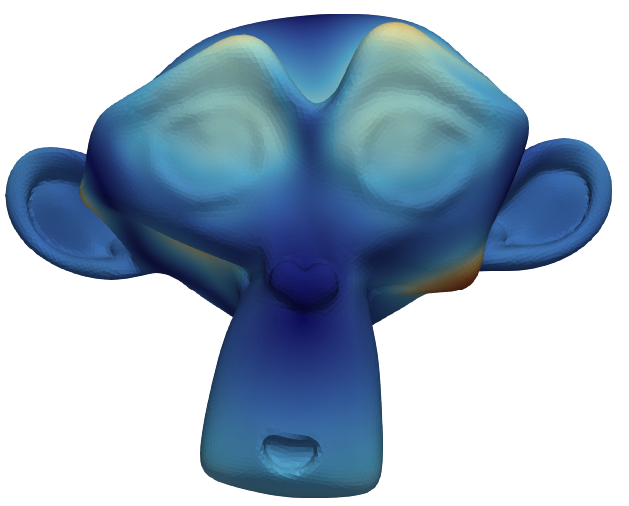}
                {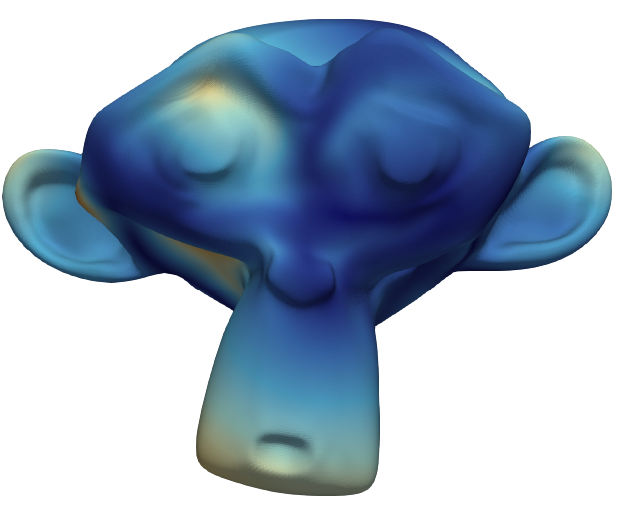}}
            {\simpairlabels{$27.5$ \si{\hertz}}{$29.8$ \si{\hertz}}}%
        }                                                                      \\ \simrowgap
        \simscenerow{Bunny}{%
            \simpairblock
            {\simpairimgs
                {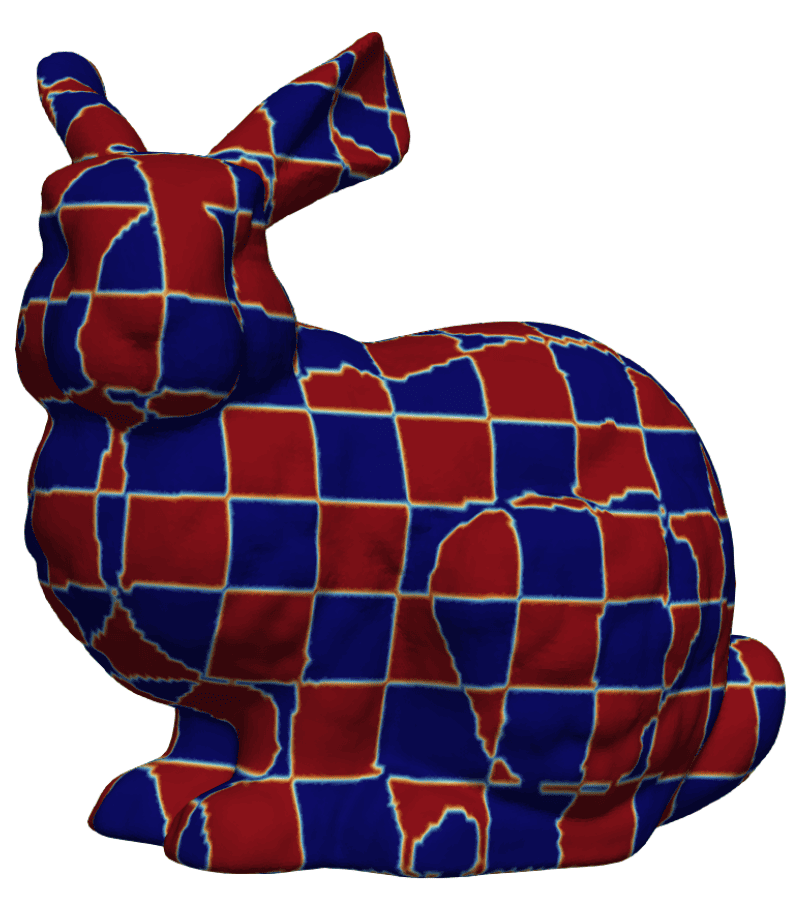}
                {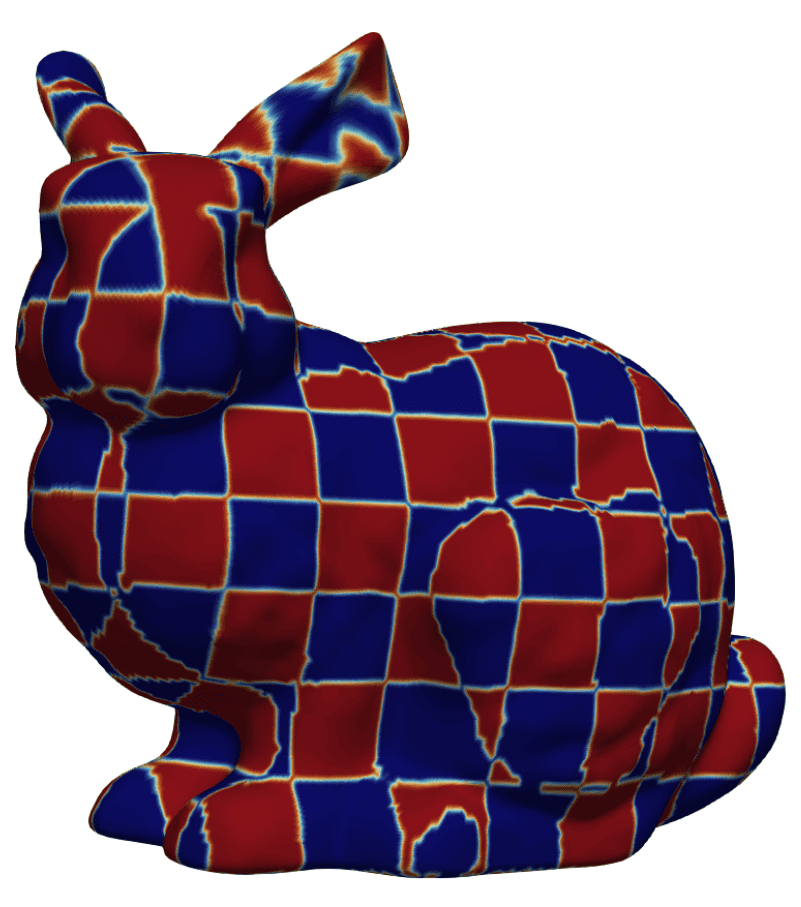}}
            {\simpairlabels{$273.4$ \si{\kilo\joule}}{$269.7$ \si{\kilo\joule}}}%
            \hspace{\simpairsep}%
            \simpairblock
            {\simpairimgs
                {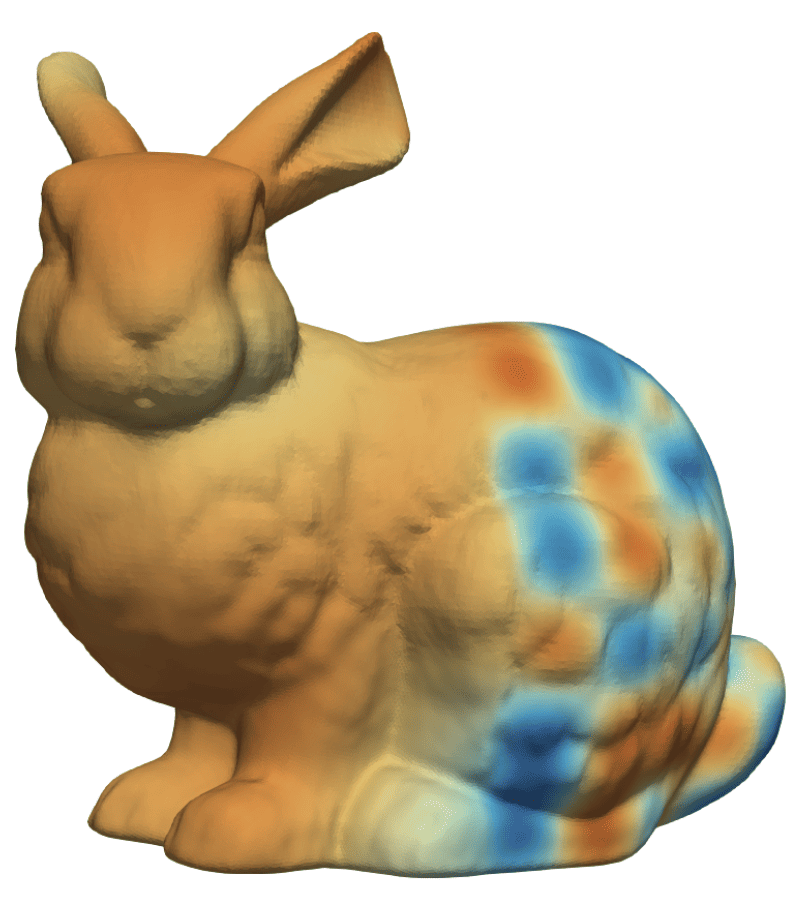}
                {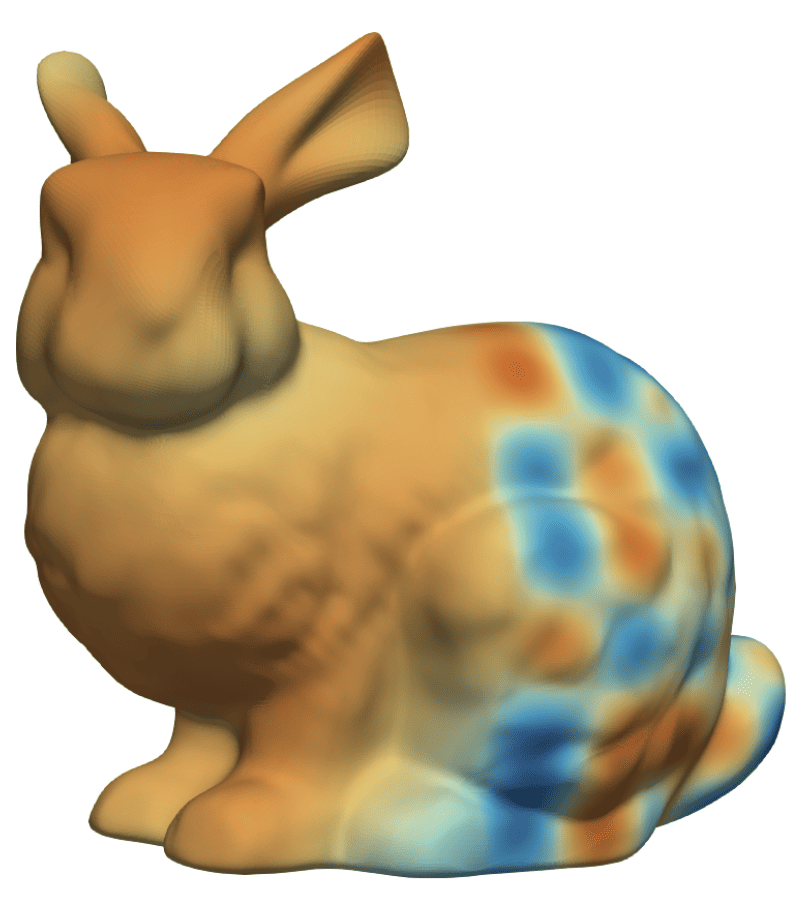}}
            {\simpairlabels{$313.7$ \si{\kilo\joule}}{$309.5$ \si{\kilo\joule}}}%
            \hspace{\simpairsep}%
            \simpairblock
            {\simpairimgs
                {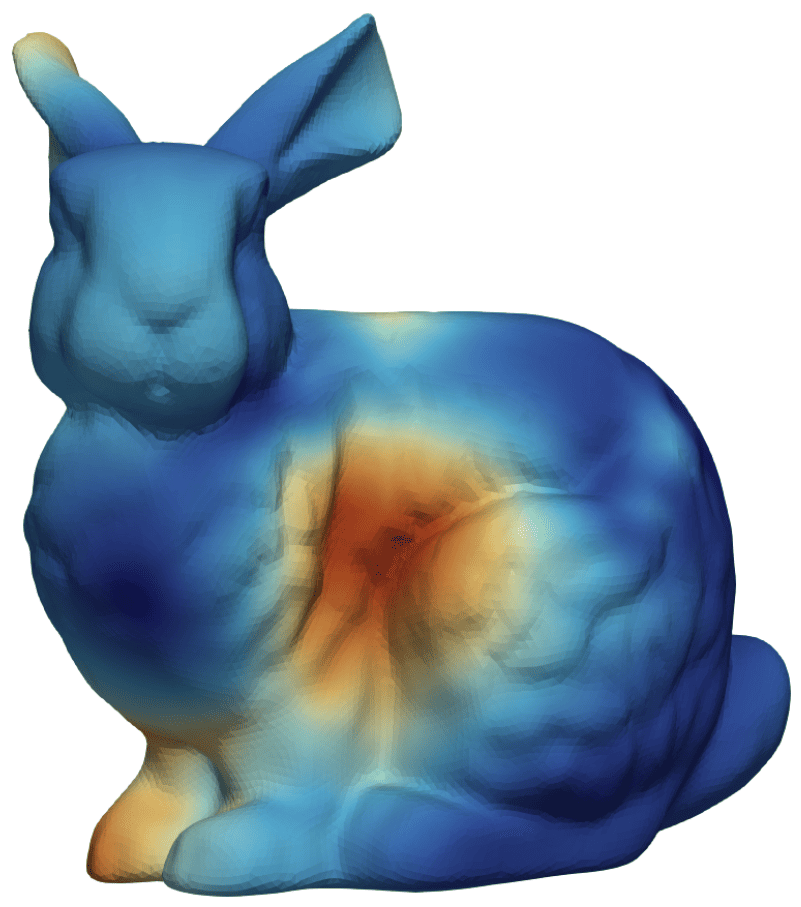}
                {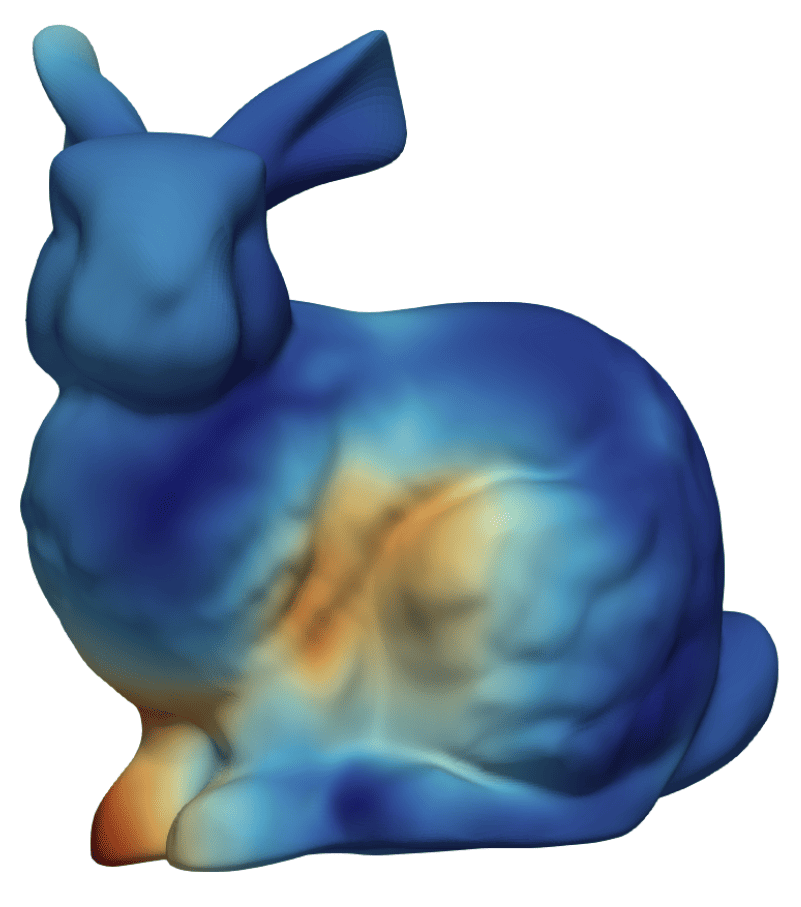}}
            {\simpairlabels{$25.9$ \si{\hertz}}{$23.9$ \si{\hertz}}}%
        }                                                                      \\ \simrowgap
        \simscenerow{Spot}{%
            \simpairblock
            {\simpairimgs
                {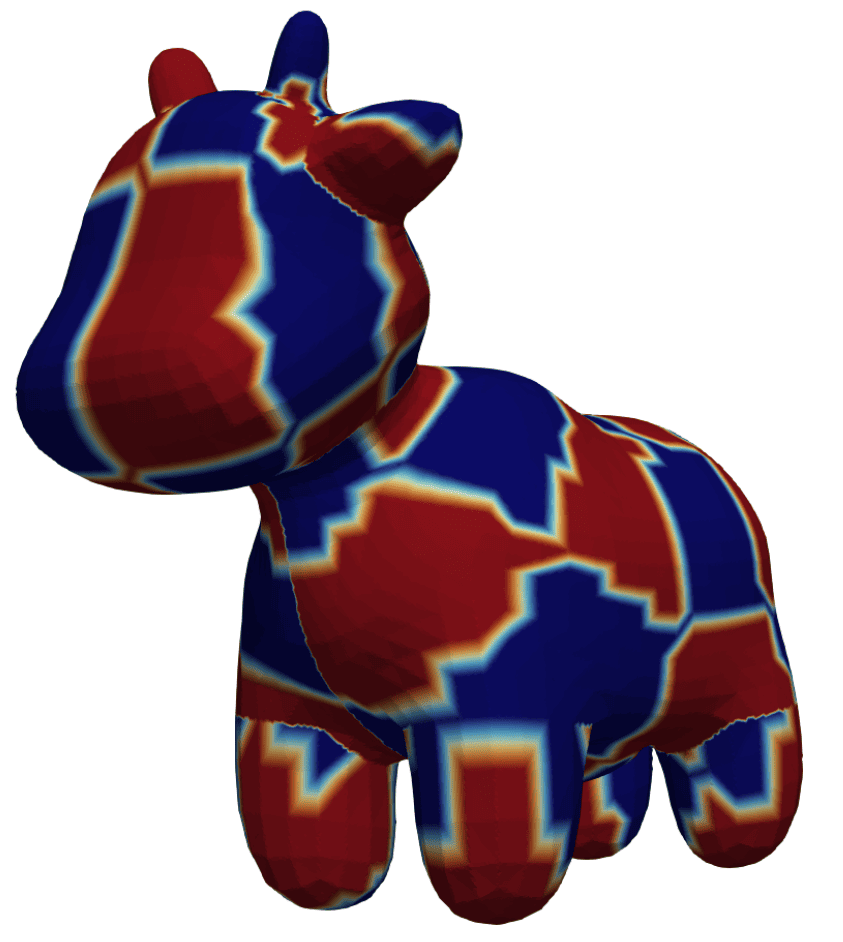}
                {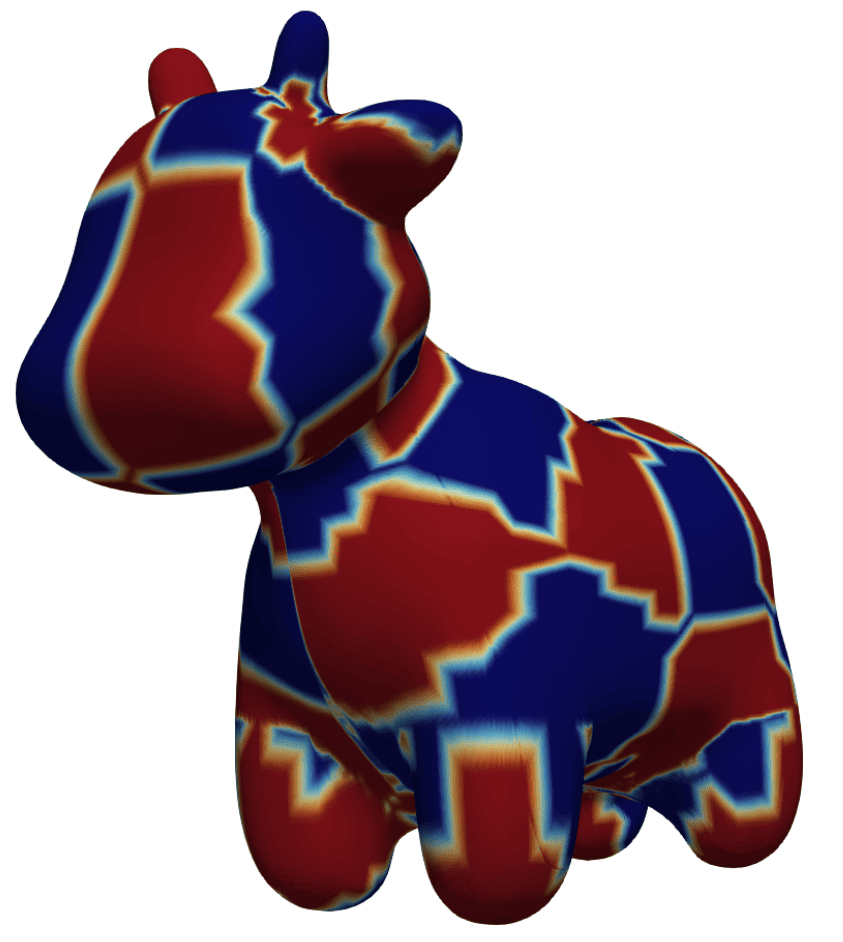}}
            {\simpairlabels{$227.4$ \si{\kilo\joule}}{$222.9$ \si{\kilo\joule}}}%
            \hspace{\simpairsep}%
            \simpairblock
            {\simpairimgs
                {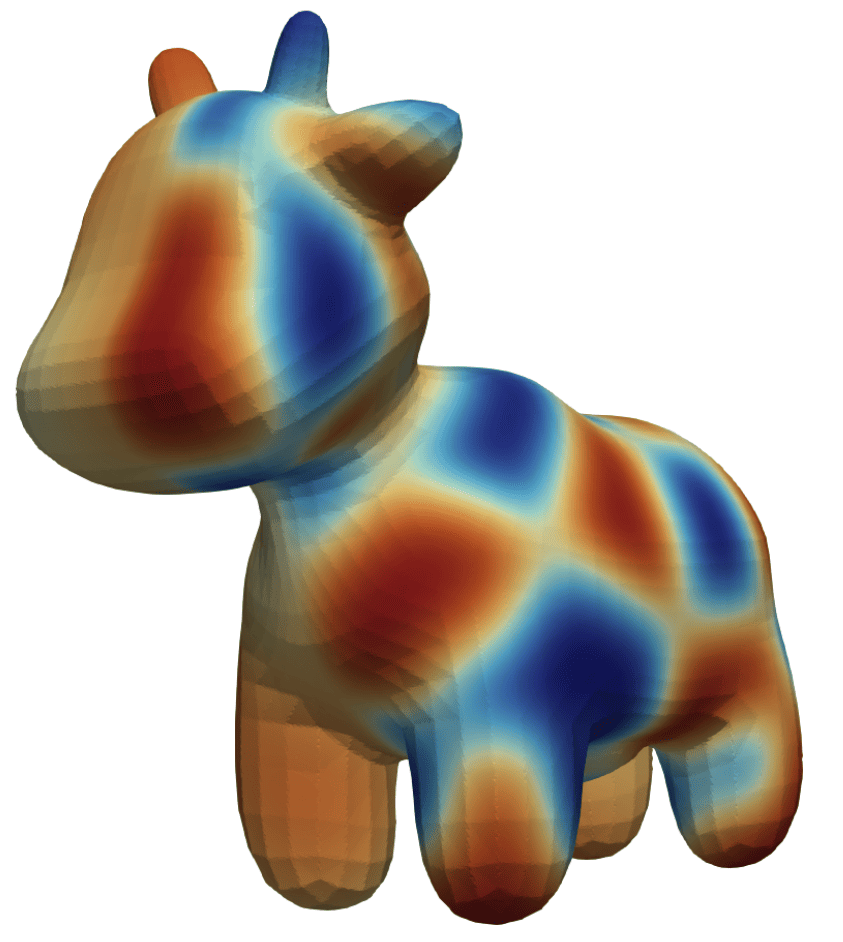}
                {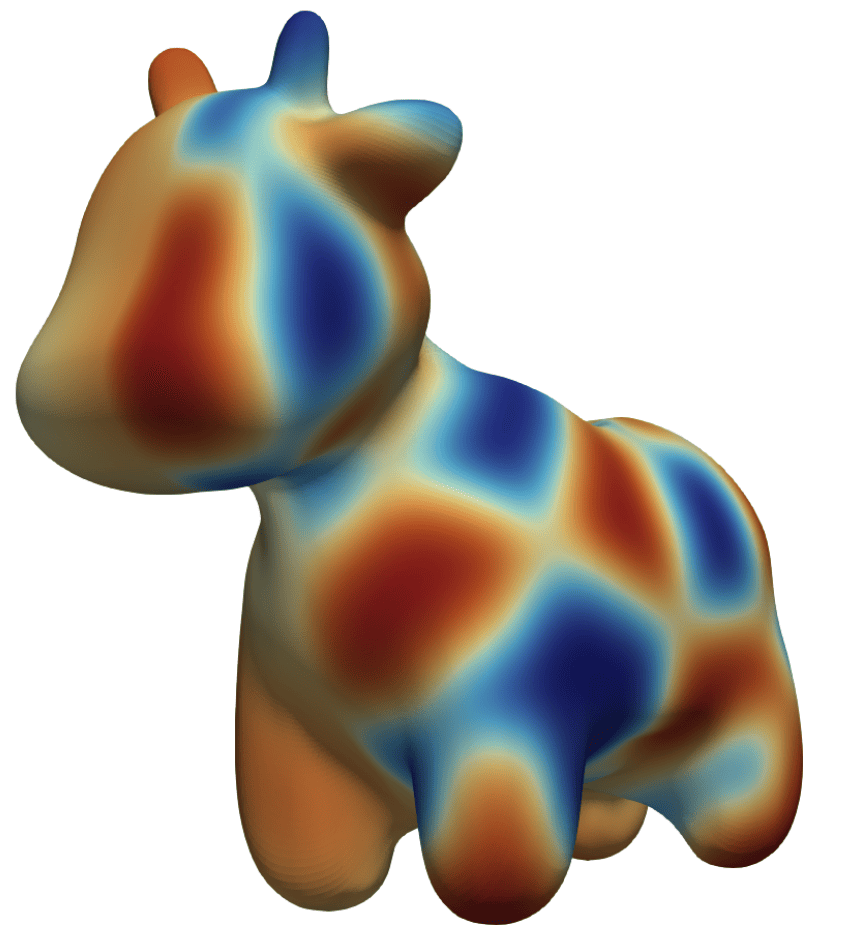}}
            {\simpairlabels{$251.8$ \si{\kilo\joule}}{$246.8$ \si{\kilo\joule}}}%
            \hspace{\simpairsep}%
            \simpairblock
            {\simpairimgs
                {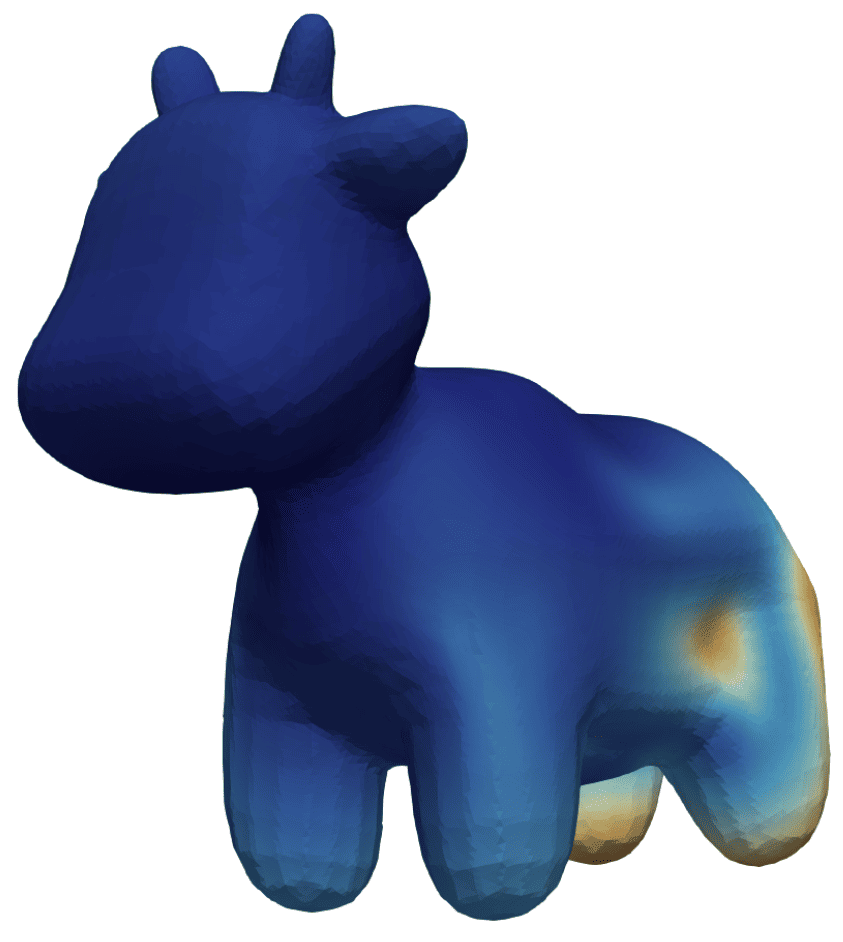}
                {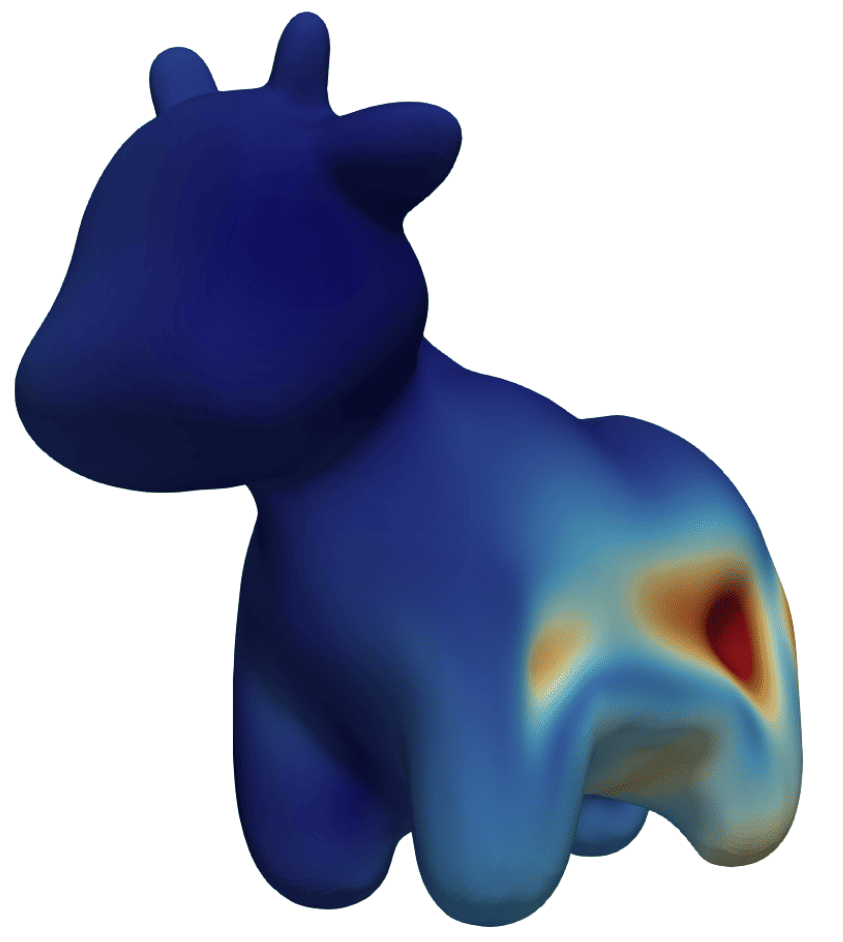}}
            {\simpairlabels{$29.3$ \si{\hertz}}{$37.8$ \si{\hertz}}}%
        }                                                                      \\ \simrowgap
        \simscenerow{Armadillo}{%
            \simpairblock
            {\simpairimgs
                {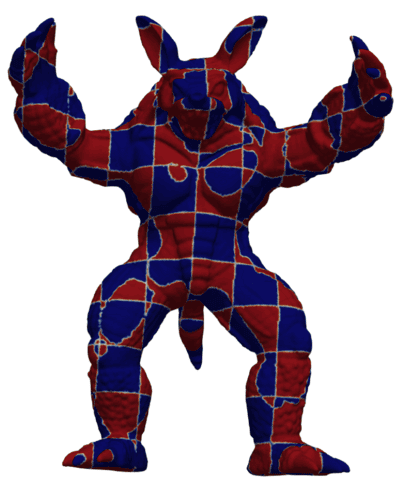}
                {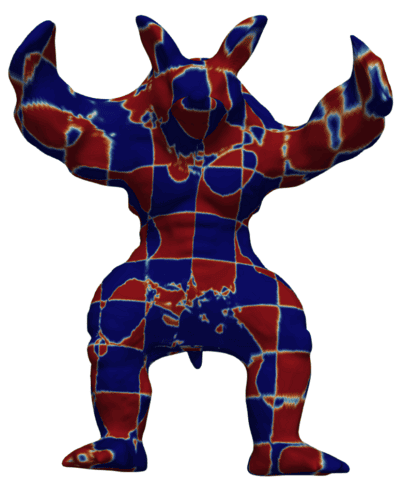}}
            {\simpairlabels{$261.6$ \si{\kilo\joule}}{$247.9$ \si{\kilo\joule}}}%
            \hspace{\simpairsep}%
            \simpairblock
            {\simpairimgs
                {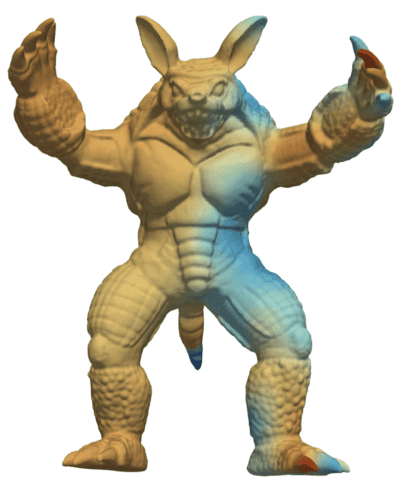}
                {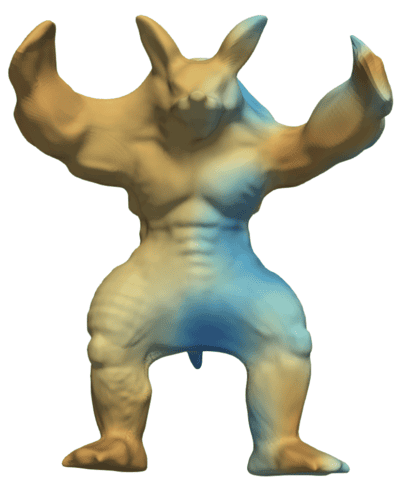}}
            {\simpairlabels{$288.8$ \si{\kilo\joule}}{$274.0$ \si{\kilo\joule}}}%
            \hspace{\simpairsep}%
            \simpairblock
            {\simpairimgs
                {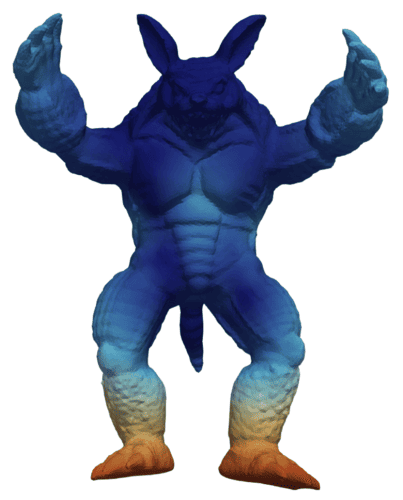}
                {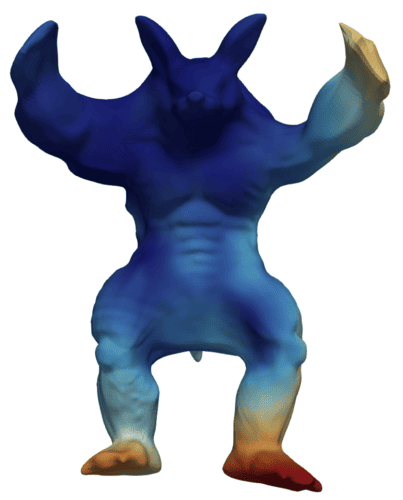}}
            {\simpairlabels{$15.1$ \si{\hertz}}{$17.0$ \si{\hertz}}}%
        }                                                                      \\
    \end{tabular}
    \vspace{6pt}

    \caption{
        Evaluation of the end-to-end pipeline against the ground truth.
        Each row is one of the four textureless meshes, ordered as in \cref{tab:e2e_summary}.
        (Left to right) Initial temperature distributions (ground truth vs. thermal inpainting); final-time heat simulation results (ground truth FEM vs. reconstructed IGA); and modal analyses (ground truth vs. reconstructed IGA).
        Subcaptions report the total thermal energy (heat columns) and the fundamental elastic frequency (modal columns).
    }
    \label{fig:e2e}
\end{figure}

As seen in \cref{tab:e2e_summary} and illustrated in \cref{fig:e2e}, our method agrees closely with the ground truth across all scenes, despite the complete decoupling of geometric reconstruction, initial temperature field, material field, and simulation.
Simulation discrepancies are correlated with reconstruction quality: the well-reconstructed Suzanne and Bunny agree to within $1.41\%$ in the final-time temperature field and $0.77\%$ in $\lambda_1$, whereas the Armadillo---whose thin appendages are the most difficult to recover---shows the largest errors ($6.89\%$ field, $19.85\%$ in $\lambda_1$).
On the other hand, the errors are numerically stable throughout the simulation; for the Armadillo, the initial-state energy error ($5.26\%$) remains essentially constant at the final state ($5.1\%$).
We validated that these errors are significantly higher than those attributable to different simulation frameworks (Supplementary,  \cref{suppmat:solvervalidation}).

\begin{table}[t]
    \centering
    \caption{
        End-to-end validation on the four textureless meshes.
        Isogeometric simulation on the \methodname{}-reconstructed spline versus finite-element simulation on the ground-truth mesh.
        All entries are relative percentage errors, except for the modal assurance criterion (MAC $\in [0,1]$; higher is better).
        The field-discrepancy columns ${FL}^2$ are floor-corrected (Supplementary, \cref{sup_mat:endtoendheat}).
        Errors are low across all scenes and correlate with reconstruction complexity (see geometric reconstruction metrics in \cref{tab:results:reconstruction}).
    }
    \label{tab:e2e_summary}
    \begin{tabular}{lccccccc}
        \toprule
                  & \multicolumn{5}{c}{Heat} & \multicolumn{2}{c}{Modal}                                                                       \\
        \cmidrule(lr){2-6}\cmidrule(lr){7-8}
        Mesh      & Init.\ energy            & Gen.\ rate                & ${FL}^2_{t_0}$ & ${FL}^2_{t_N}$ & $\lambda_1$ & Median freq. & MAC  \\
        \midrule
        Suzanne   & 0.07                     & 2.06                      & 1.97           & 1.41           & 0.72        & 2.47         & 0.53 \\
        Bunny     & 1.35                     & 1.21                      & 6.95           & 1.07           & 0.77        & 5.35         & 0.43 \\
        Spot      & 1.99                     & 1.78                      & 7.02           & 3.35           & 5.93        & 7.94         & 0.34 \\
        Armadillo & 5.26                     & 3.93                      & 14.36          & 6.89           & 19.85       & 8.72         & 0.32 \\
        \bottomrule
    \end{tabular}
\end{table}

\begin{figure}[!tp]
    \centering
    \setlength{\tabcolsep}{0pt}
    \begin{tabular}{@{}l@{}}
        \simheader{\textbf{Heat}}{\textbf{Modal}}                              \\ \simheadgap
        \simscenerow{Building 1}{%
            \simblock
            {\simimgs
                {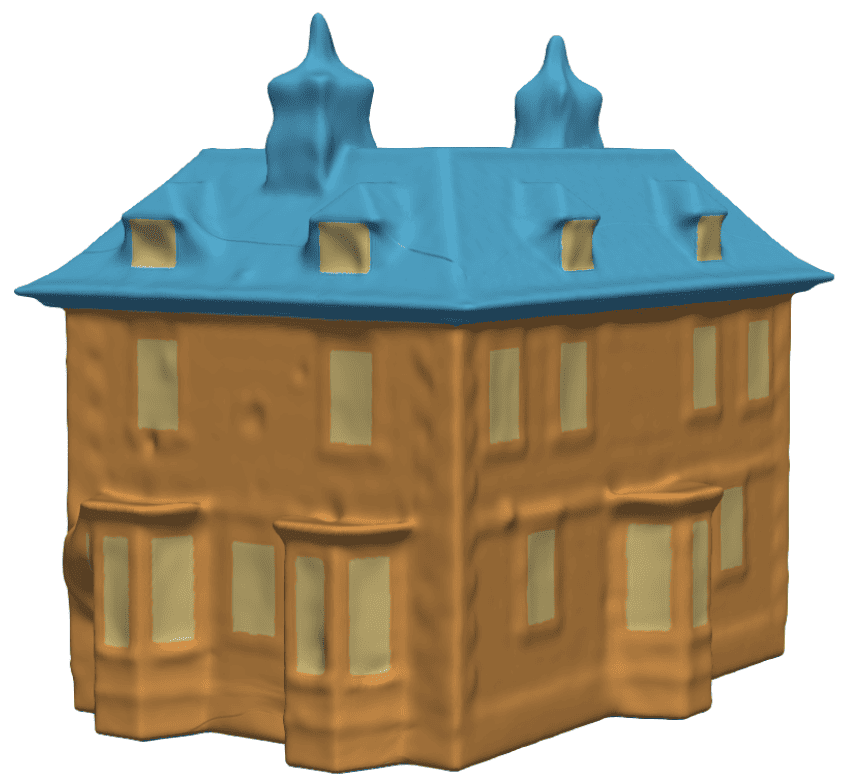}
                {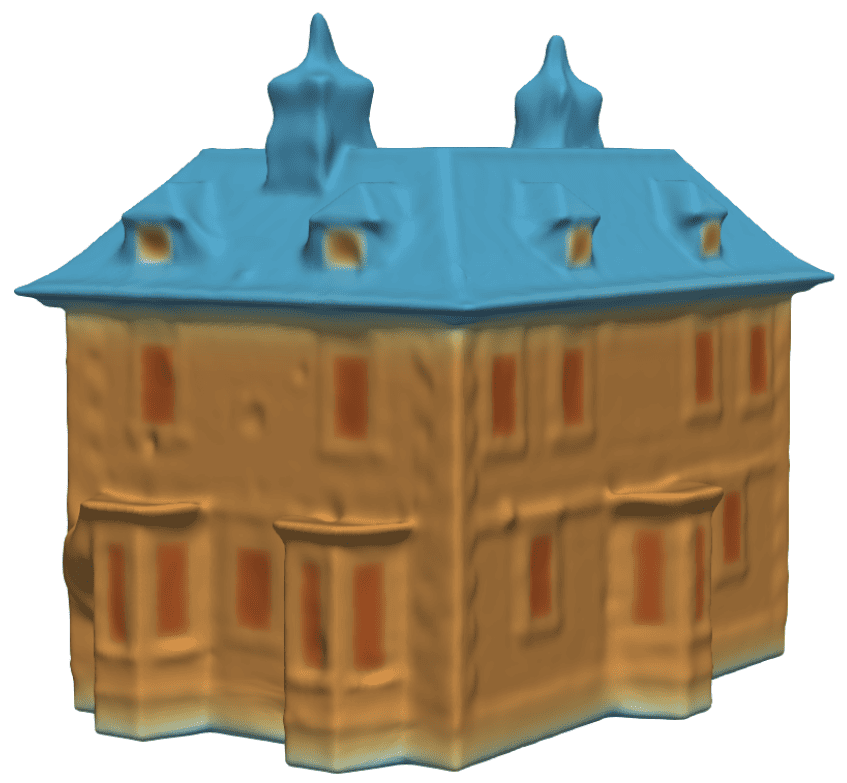}
                {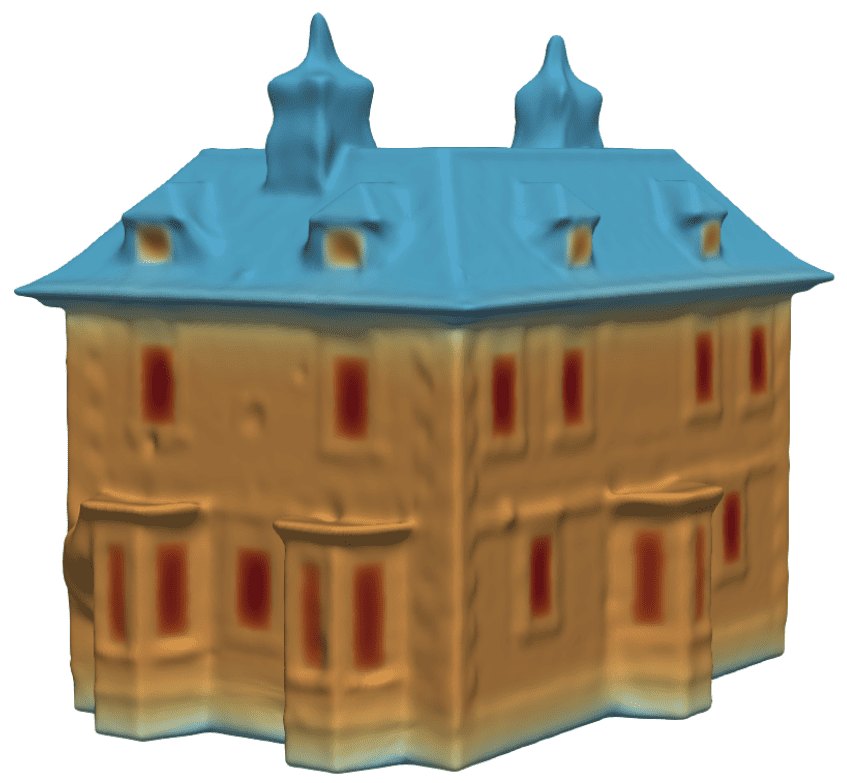}}
            {\simbar{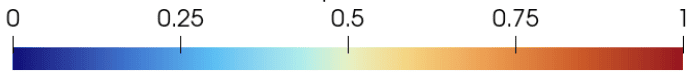}}%
            \hspace{\simgroupsep}%
            \simblock
            {\simimgs
                {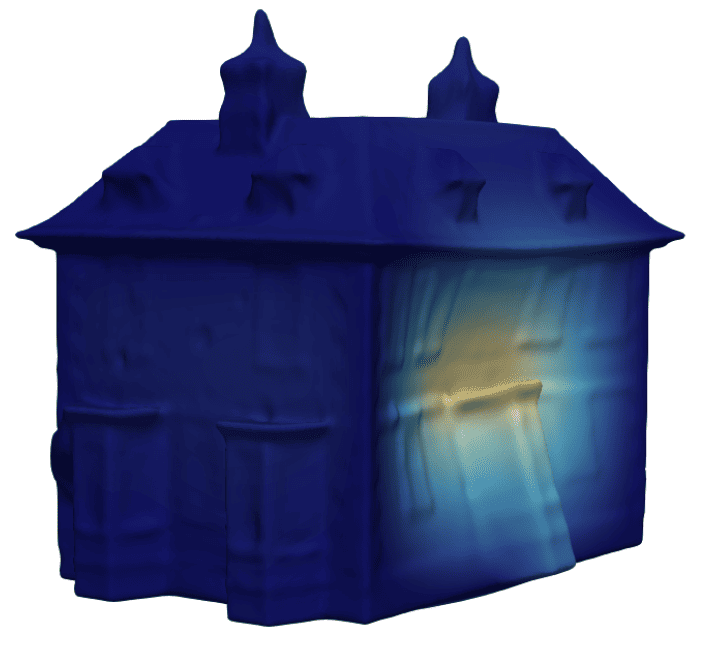}
                {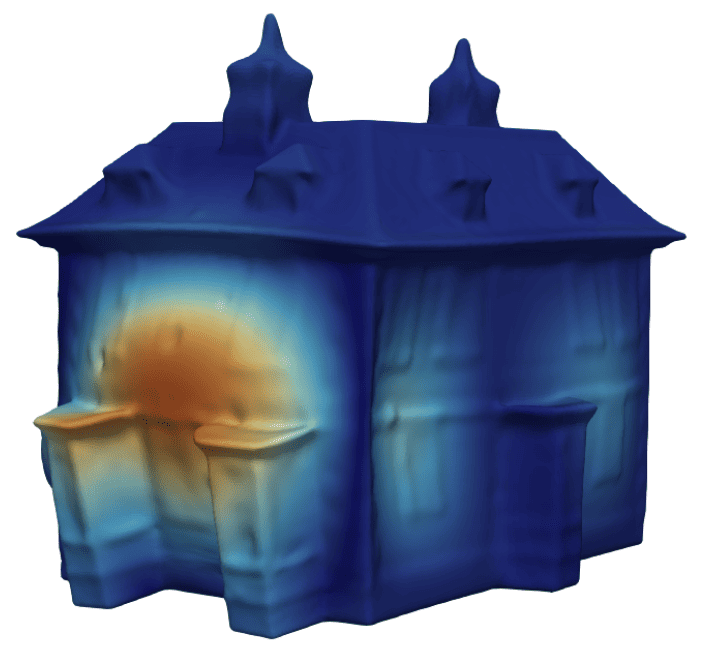}
                {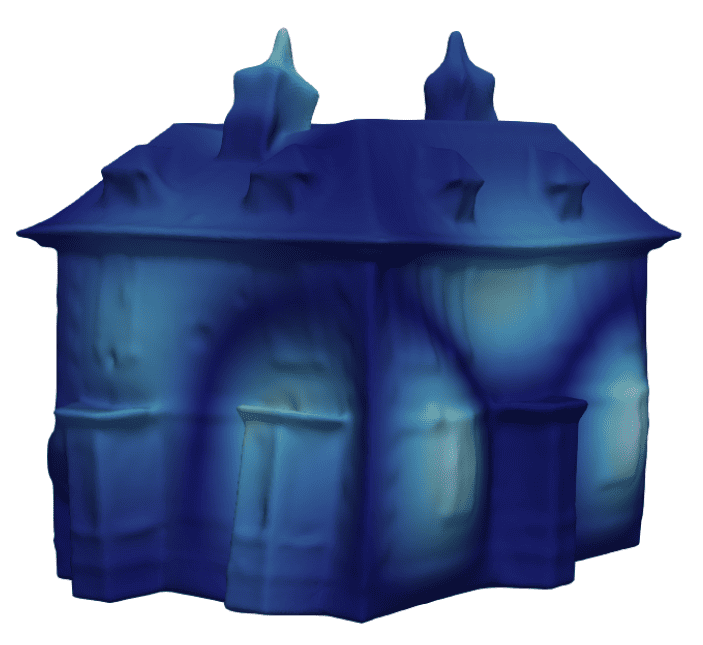}}
            {\simfreqs{$f_{6}=7.9$}{$f_{12}=16.7$}{$f_{19}=24.5$}}%
        }                                                                      \\ \simrowgap
        \simscenerow{Building 2}{%
            \simblock
            {\simimgs
                {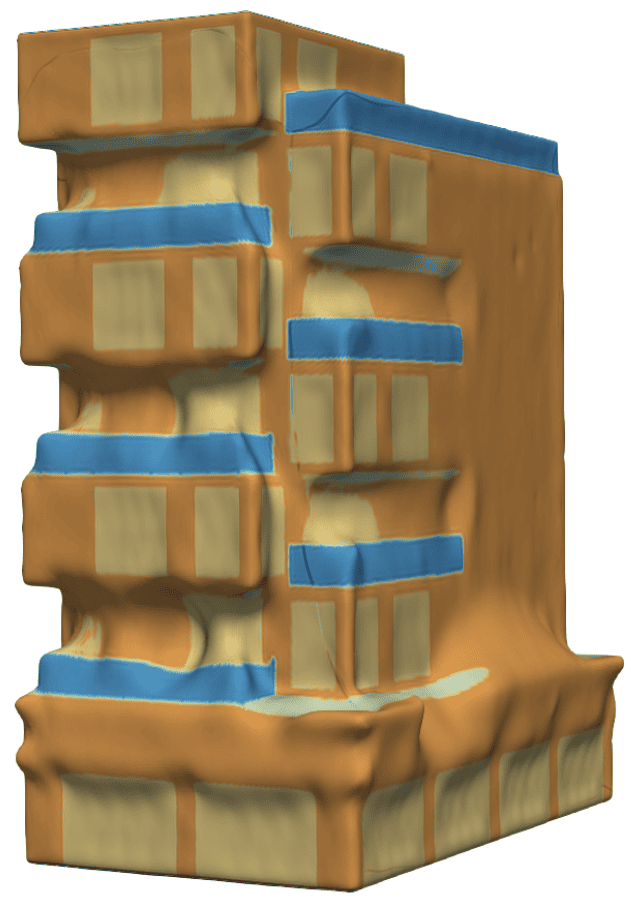}
                {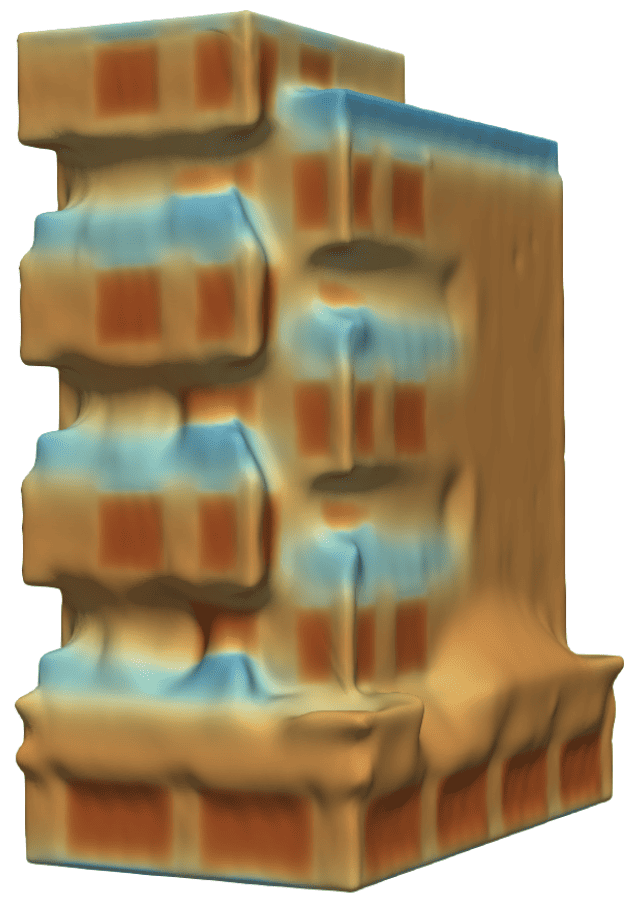}
                {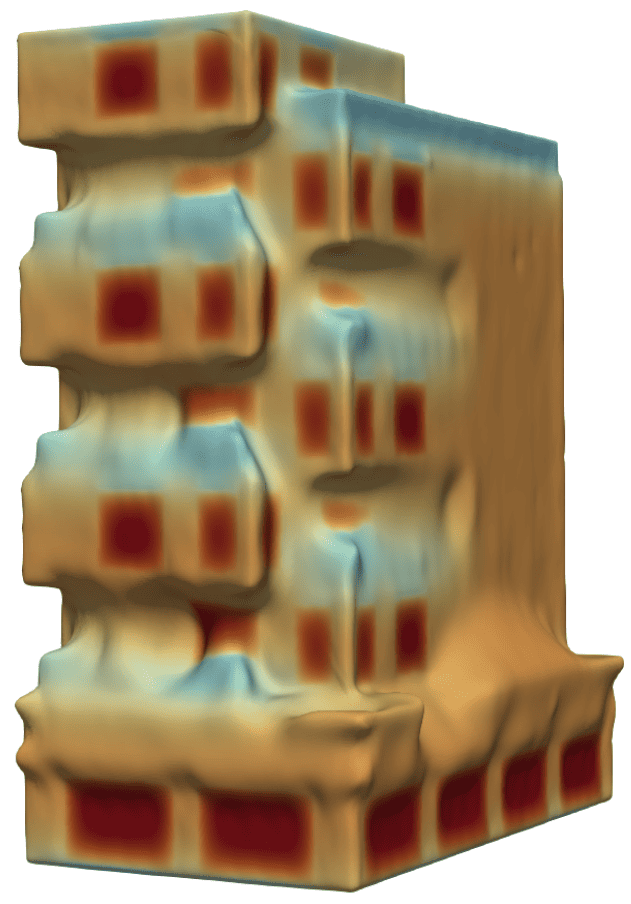}}
            {\simbar{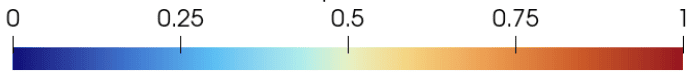}}%
            \hspace{\simgroupsep}%
            \simblock
            {\simimgs
                {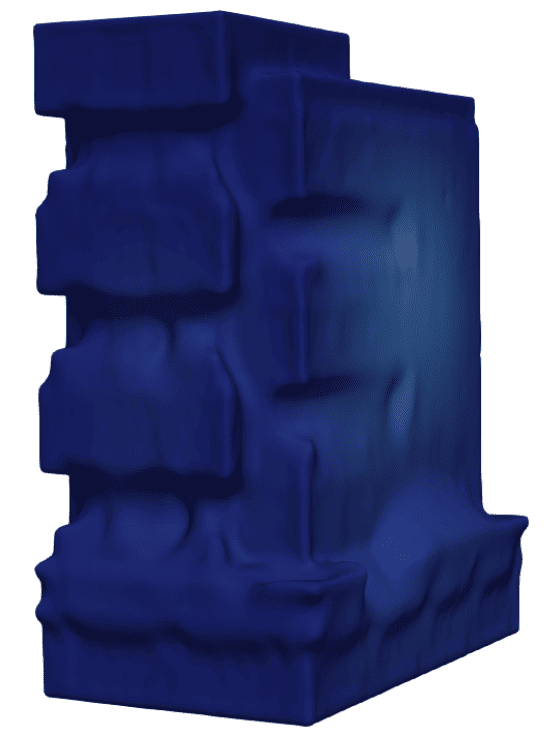}
                {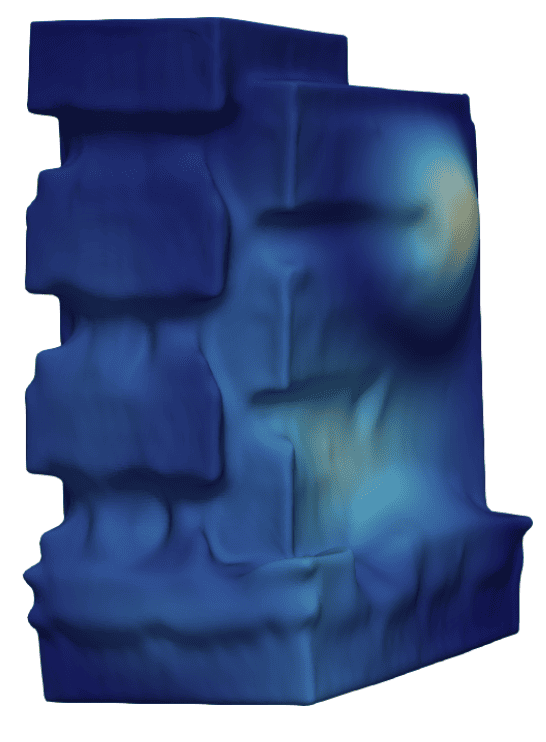}
                {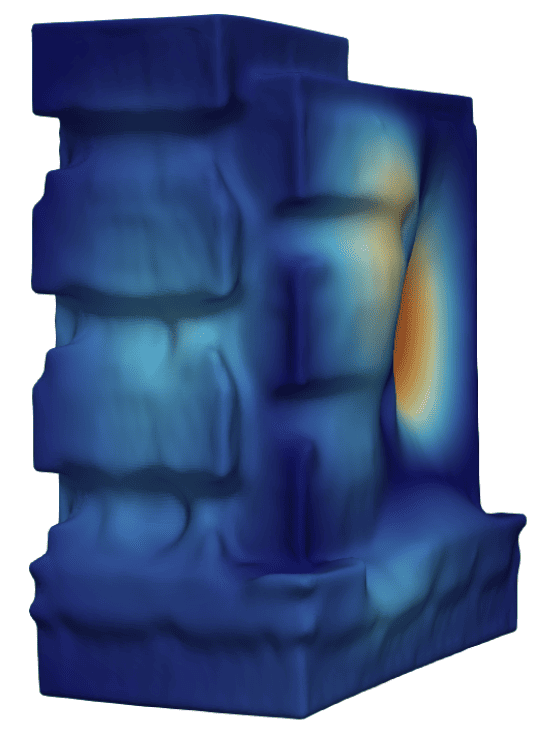}}
            {\simfreqs{$f_{6}=3.6$}{$f_{12}=9.3$}{$f_{17}=11.4$}}%
        }                                                                      \\ \simrowgap
        \simscenerow{Building 3}{%
            \simblock
            {\simimgs
                {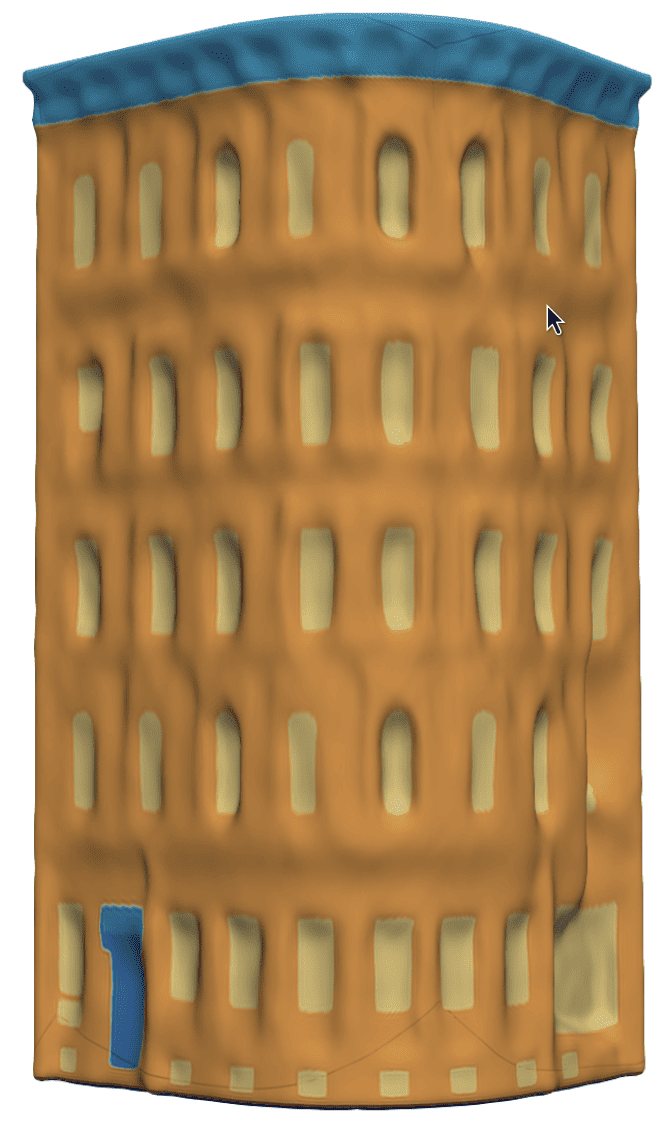}
                {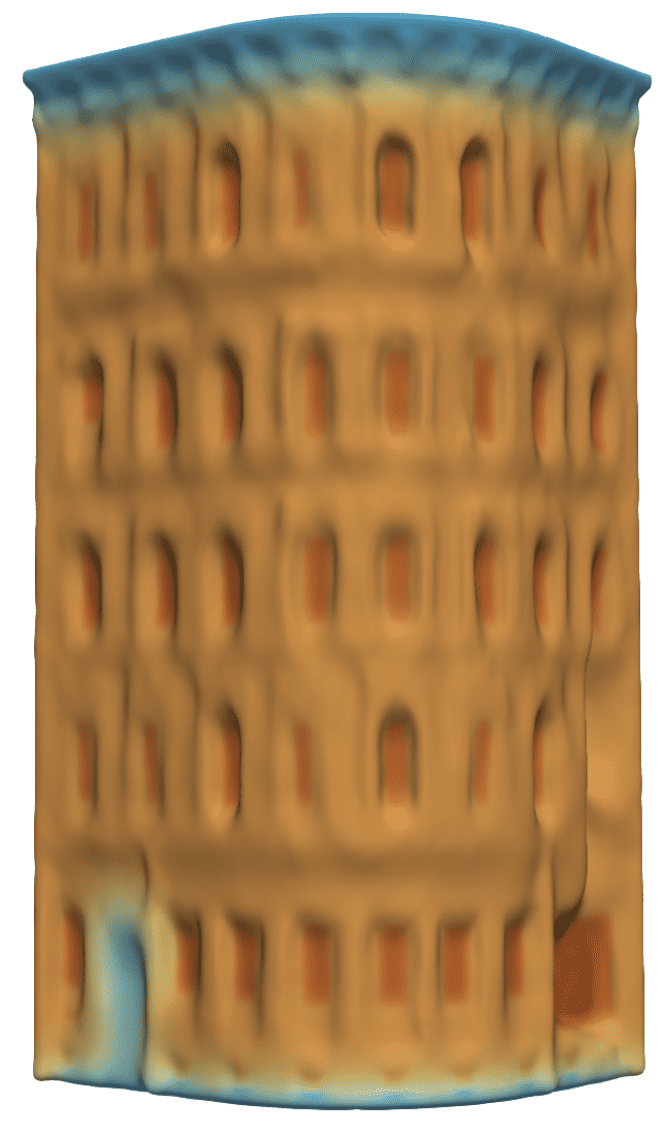}
                {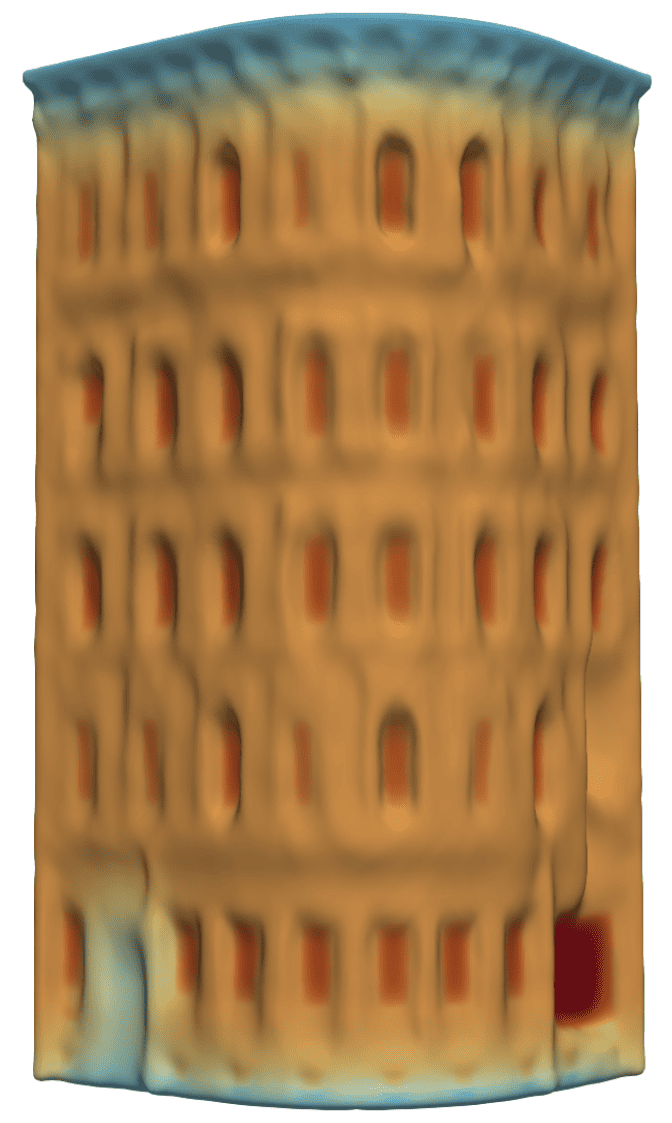}}
            {\simbar{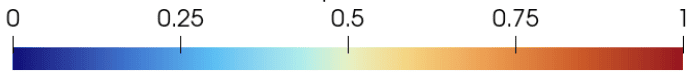}}%
            \hspace{\simgroupsep}%
            \simblock
            {\simimgs
                {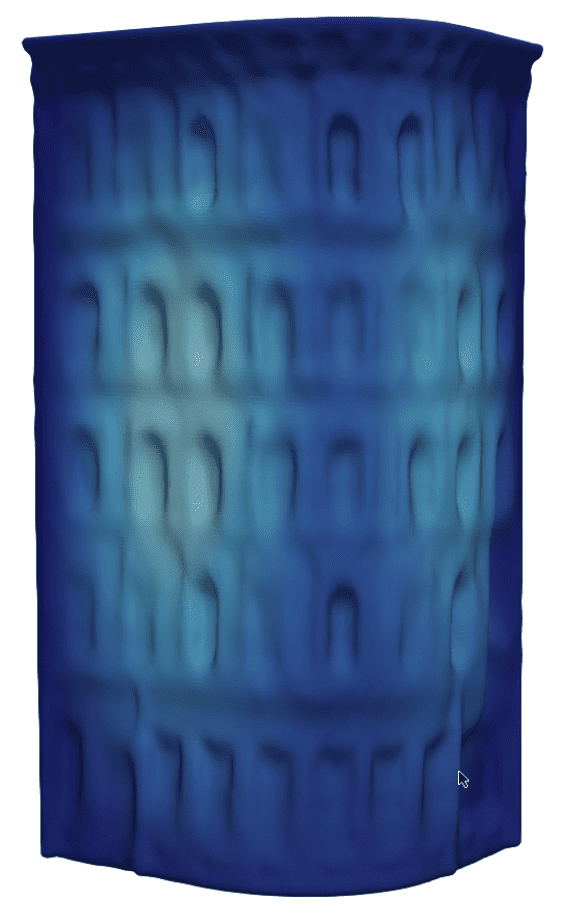}
                {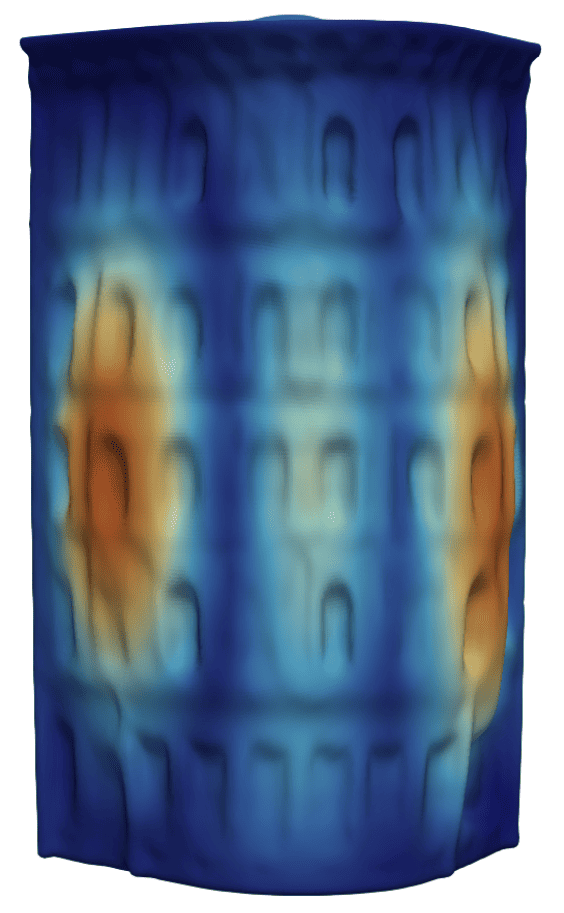}
                {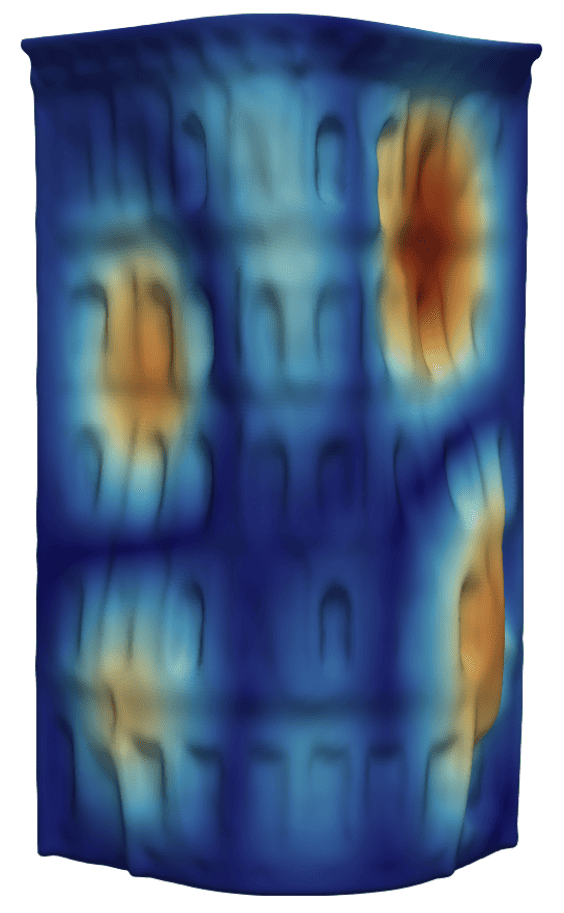}}
            {\simfreqs{$f_{6}=8.9$}{$f_{13}=20.8$}{$f_{17}=26.5$}}%
        }                                                                      \\ \simrowgap
        \simscenerow{Building 4}{%
            \simblock
            {\simimgs
                {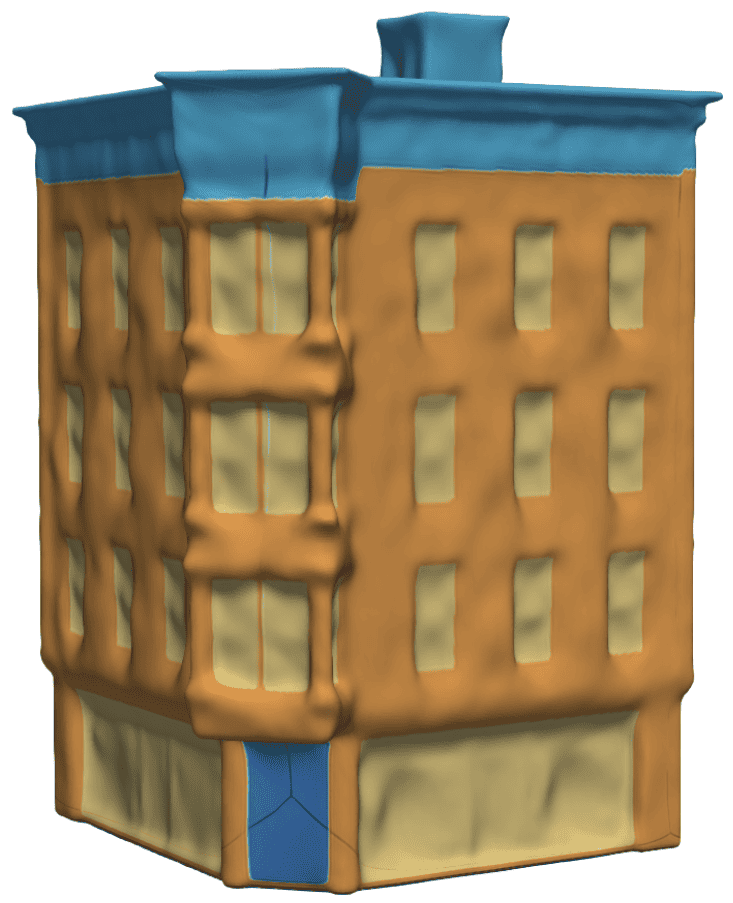}
                {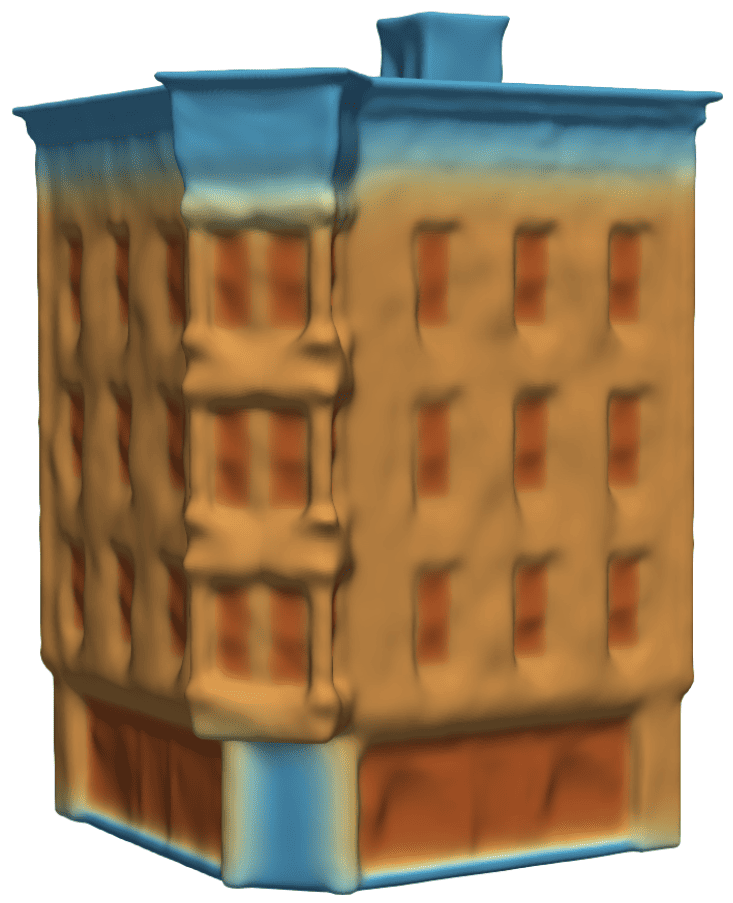}
                {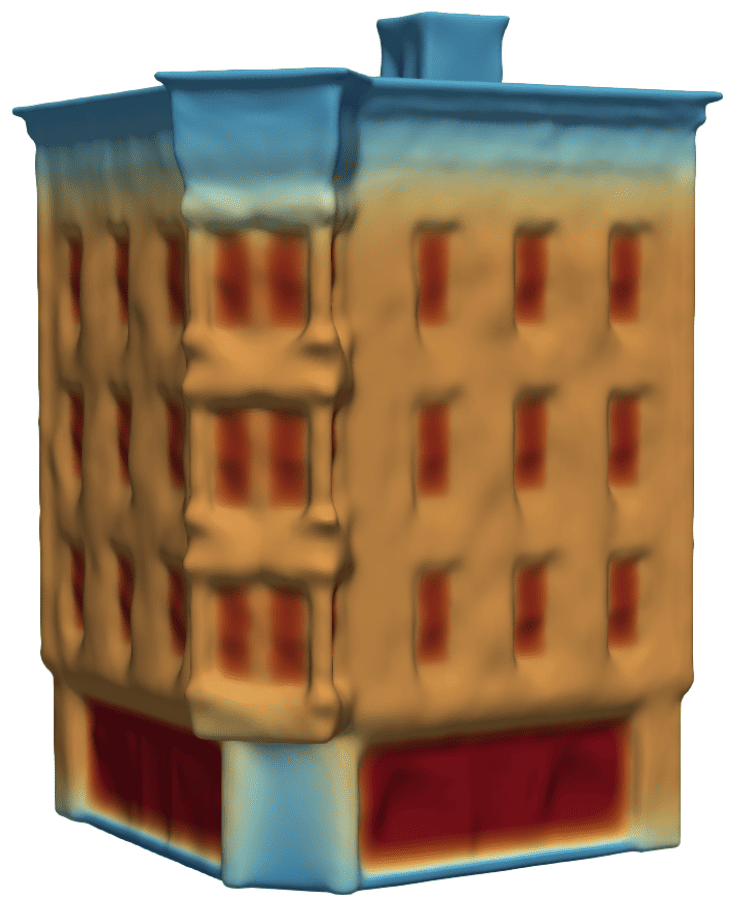}}
            {\simbar{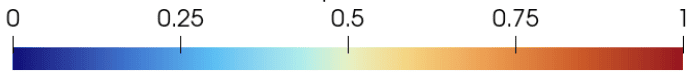}}%
            \hspace{\simgroupsep}%
            \simblock
            {\simimgs
                {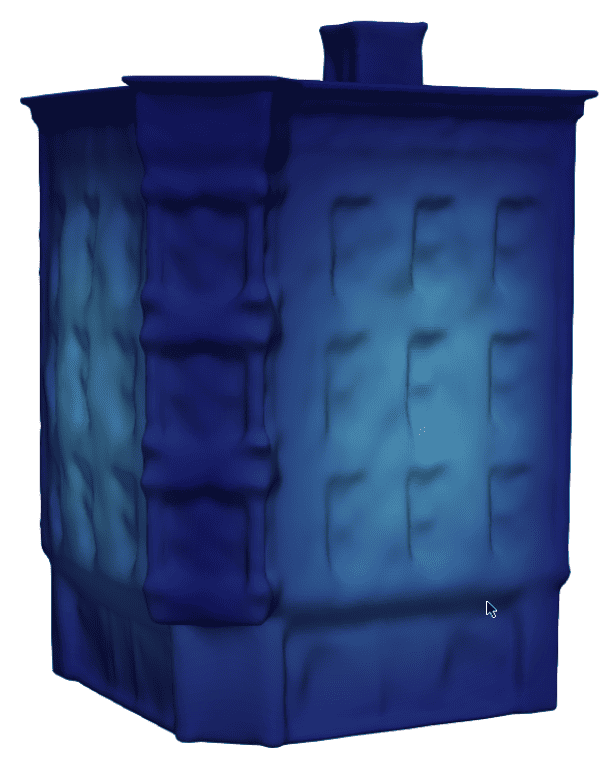}
                {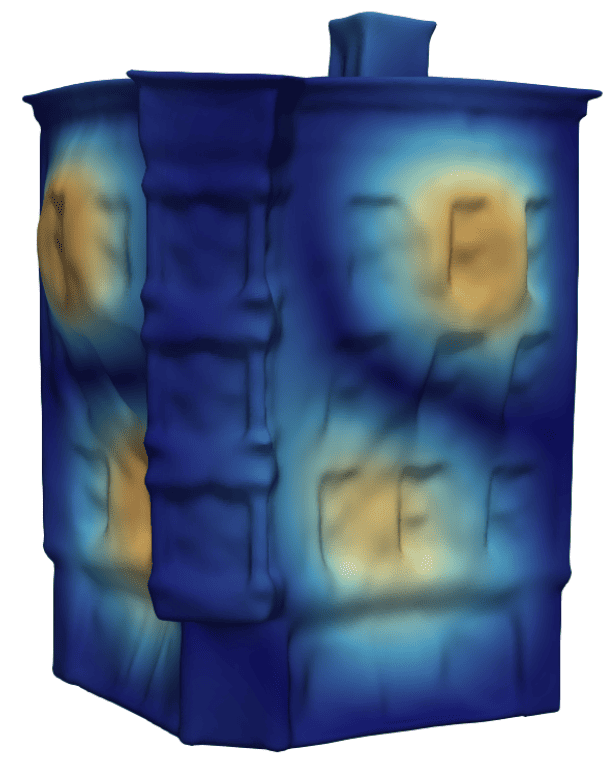}
                {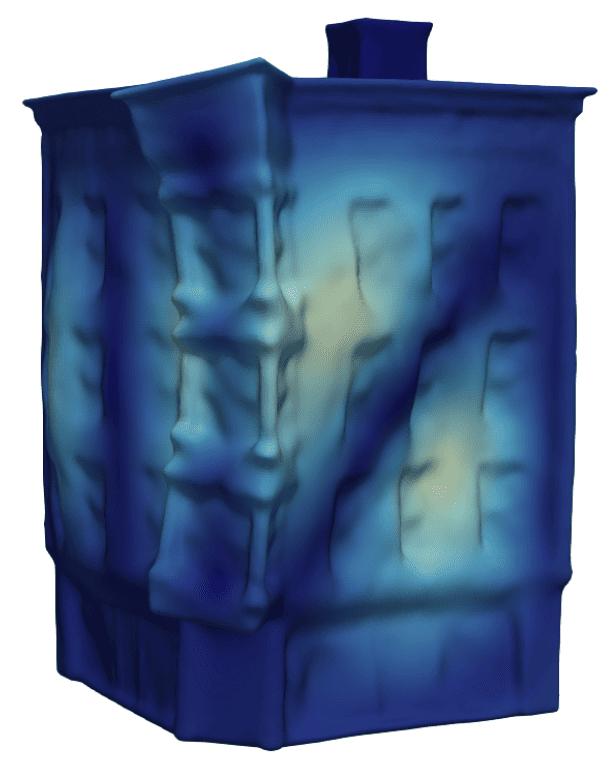}}
            {\simfreqs{$f_{6}=7.8$}{$f_{16}=21.0$}{$f_{17}=21.4$}}%
        }                                                                      \\
    \end{tabular}

    \caption{
        Thermal and modal simulation on the four synthetic building scenes.
        Each row is a single \methodname{} reconstruction, simulated in both modalities.
        \textbf{Heat}: 
        Predefined initial thermal state followed by the simulated heat flow progression at half time and final time, over the normalized temperature scale below.
        \textbf{Modal}: 
        We show the fundamental frequency ($f_6$) and two higher-order modes to characterize the structural response; frequencies are reported in \si{\hertz}.
    }
    \label{fig:sim_synthetic}
\end{figure}

To further evaluate the framework under more realistic conditions, we applied our pipeline to the synthetic building dataset.
While this dataset maintains known poses and no sensor noise, it introduces simultaneous geometry and texture reconstruction.
The reconstruction difficulty increases, from the simpler Buildings 3 and 4, to the more challenging Building 1 with a lot of fine detail and Building 2 which is non-convex, relying on the weaker shading gradients to recess the areas near the balconies.
To increase simulation realism and complexity, the simulations feature a heat source in regions classified as windows.
The resulting thermal evolution is shown in the heat columns of \cref{fig:sim_synthetic}.
The simulations demonstrate physically coherent thermal diffusion; for example, window regions---initially cooler---exhibit a gradual temperature increase as heat radiates from these sources across the manifold.

\begin{figure}[!t]
    \centering
    \setlength{\tabcolsep}{0pt}
    \begin{tabular}{@{}l@{}}
        \simheader{\textbf{Heat}}{\textbf{Modal}}                              \\ \simheadgap
        \simscenerow{Building A}{%
            \simblock
            {\simimgs
                {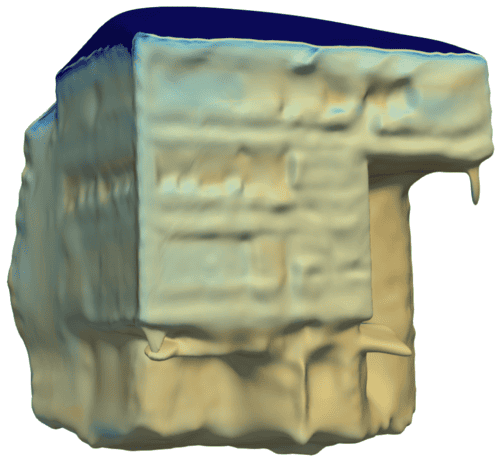}
                {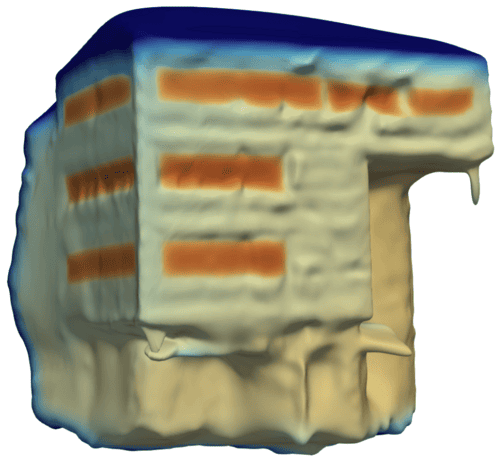}
                {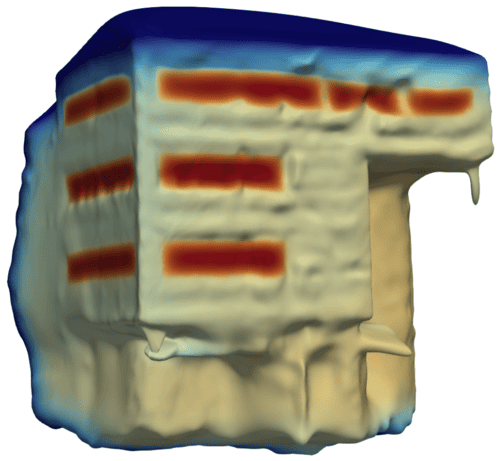}}
            {\simbar{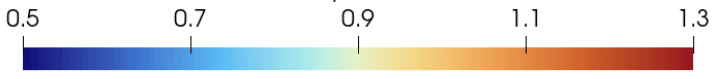}}%
            \hspace{\simgroupsep}%
            \simblock
            {\simimgs
                {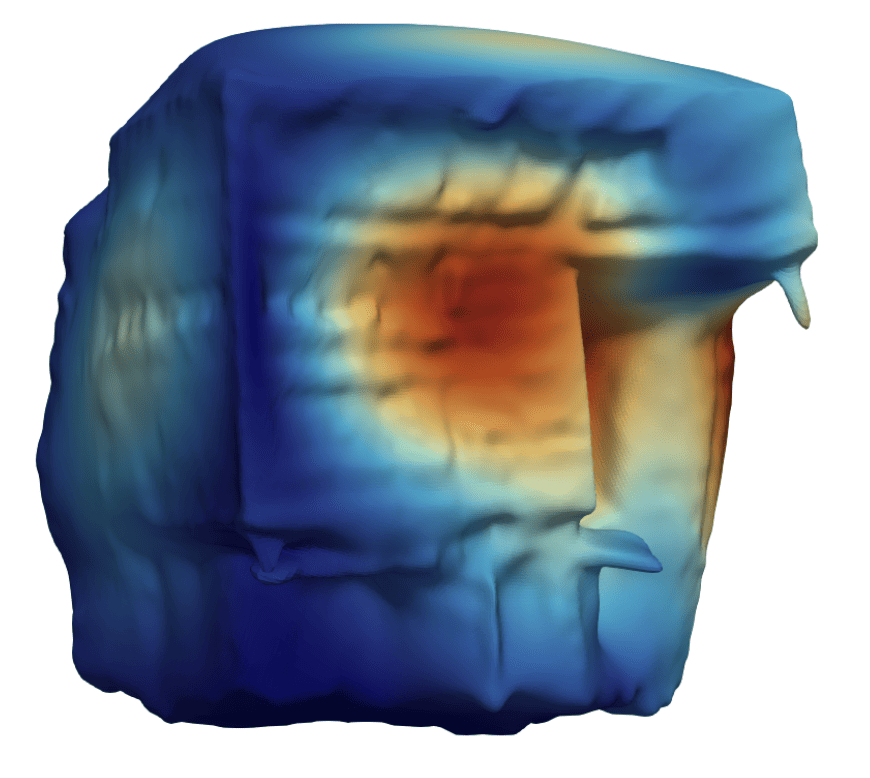}
                {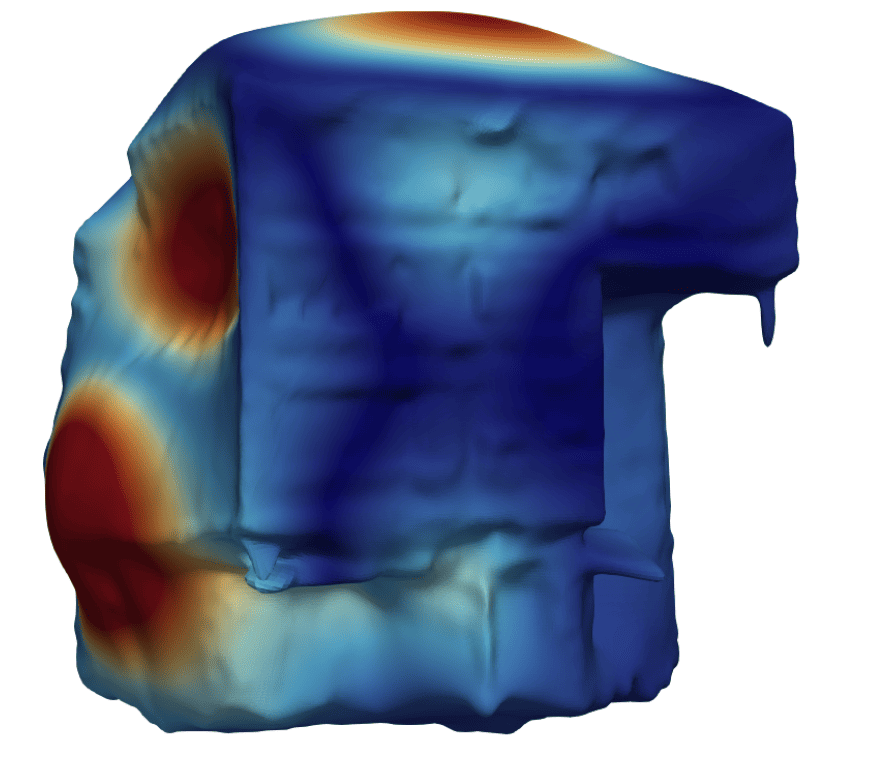}
                {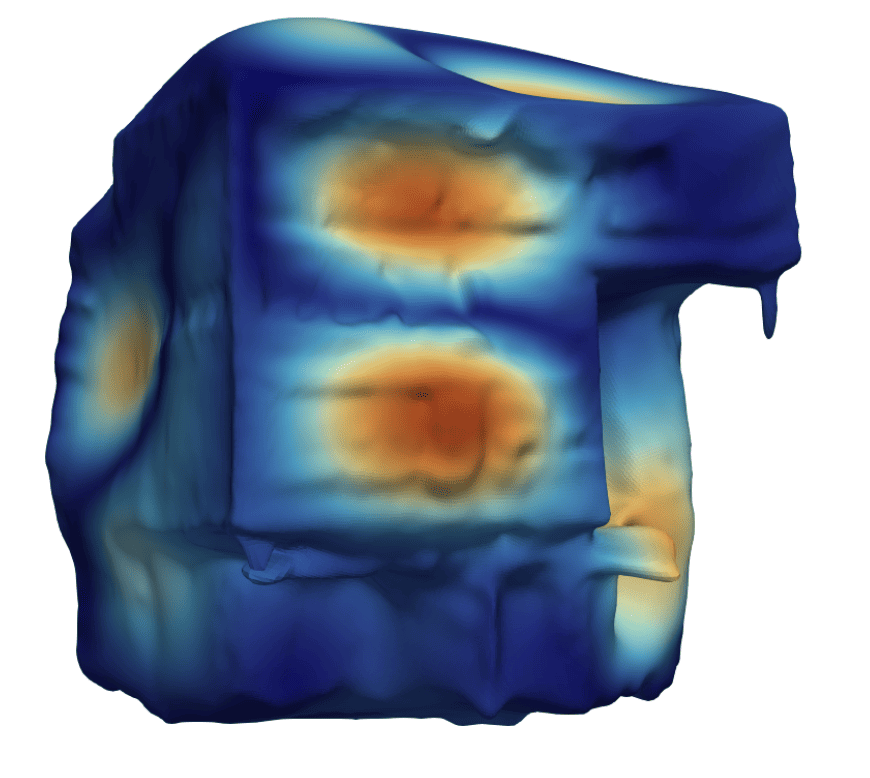}}
            {\simfreqs{$f_{6}=14.3$}{$f_{11}=26.5$}{$f_{19}=39.0$}}%
        }                                                                      \\ \simrowgap
        \simscenerow{Car}{%
            \simblock
            {\simimgs
                {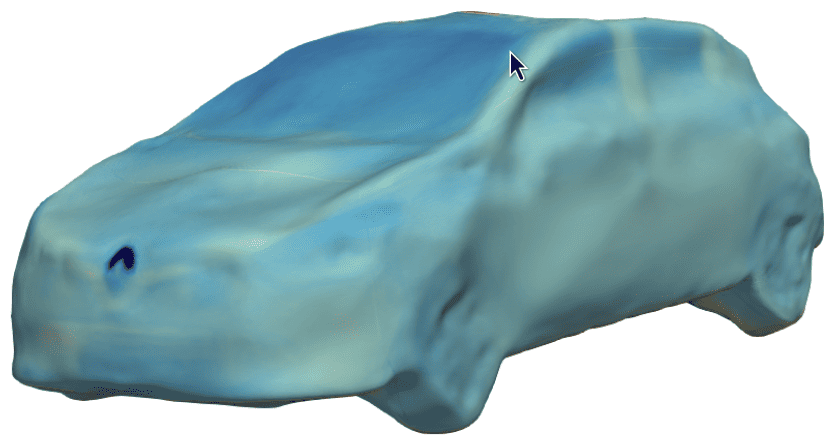}
                {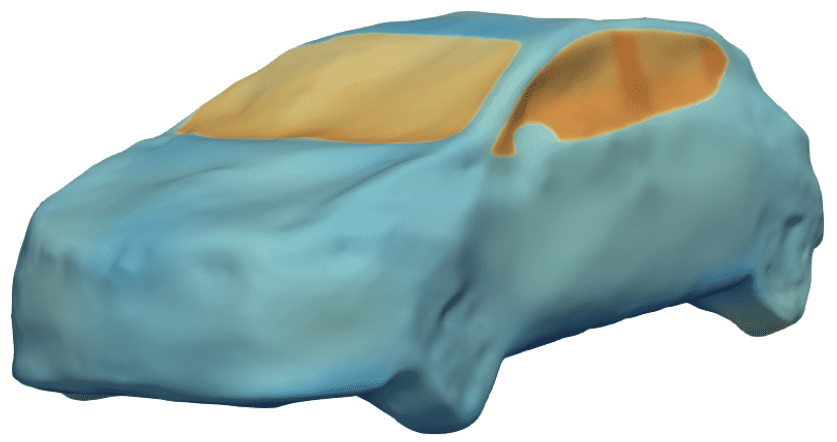}
                {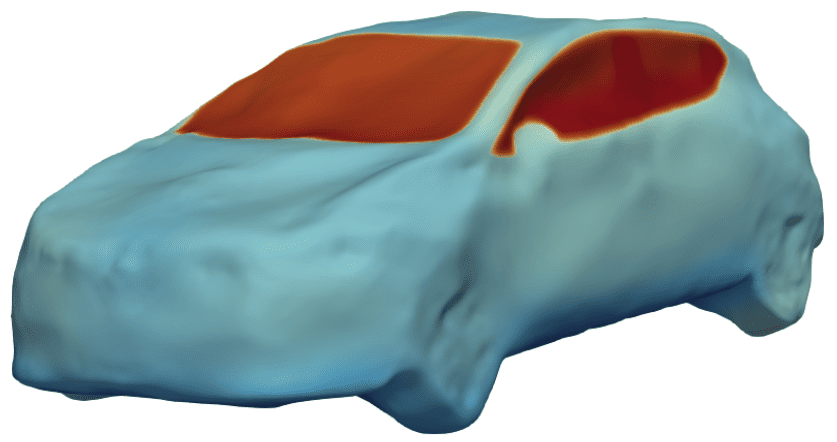}}
            {\simbar{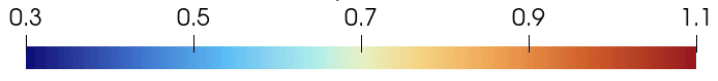}}%
            \hspace{\simgroupsep}%
            \simblock
            {\simimgs
                {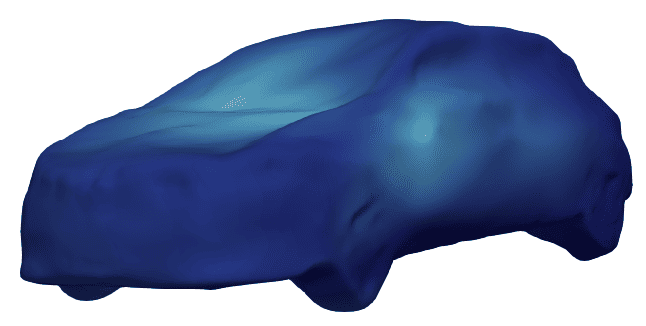}
                {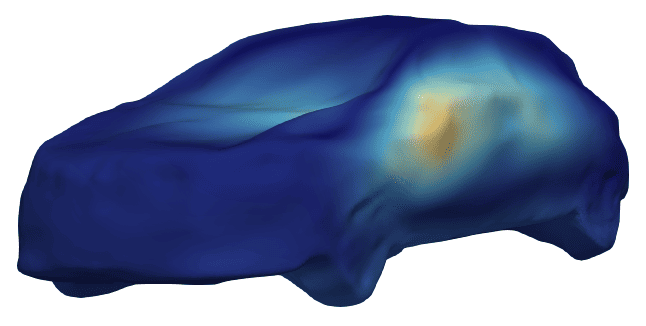}
                {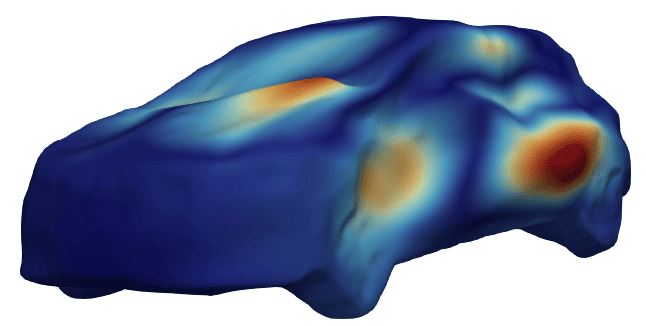}}
            {\simfreqs{$f_{6}=15.4$}{$f_{8}=20.9$}{$f_{18}=43.2$}}%
        }                                                                      \\ \simrowgap
        \simscenerow{Woodshed}{%
            \simblock
            {\simimgs
                {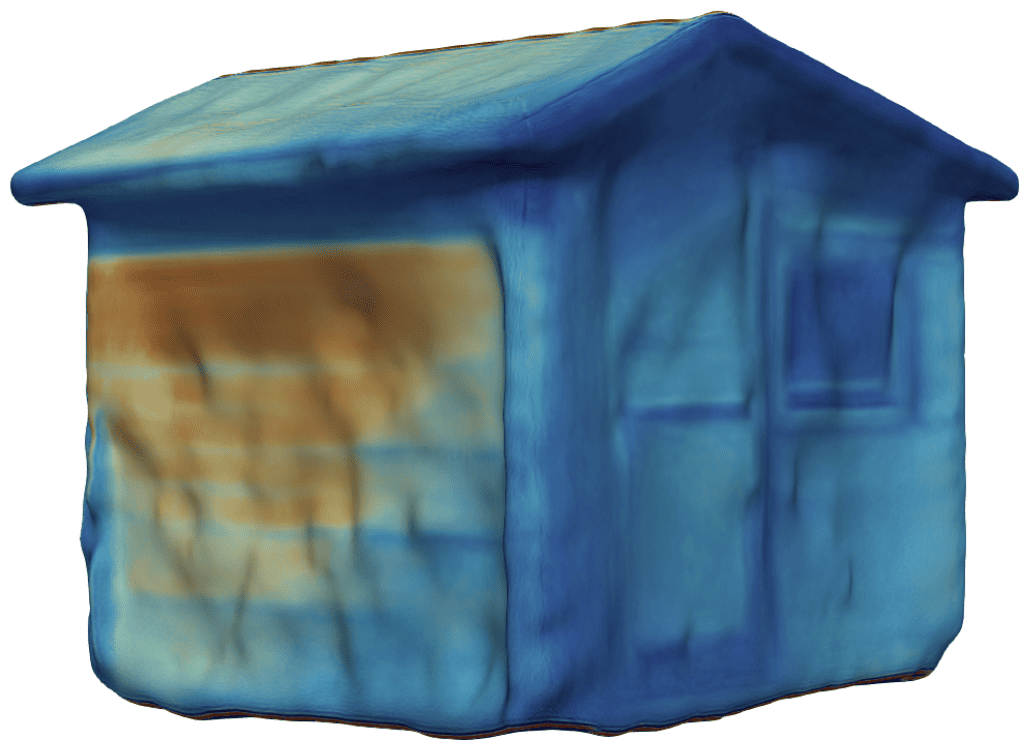}
                {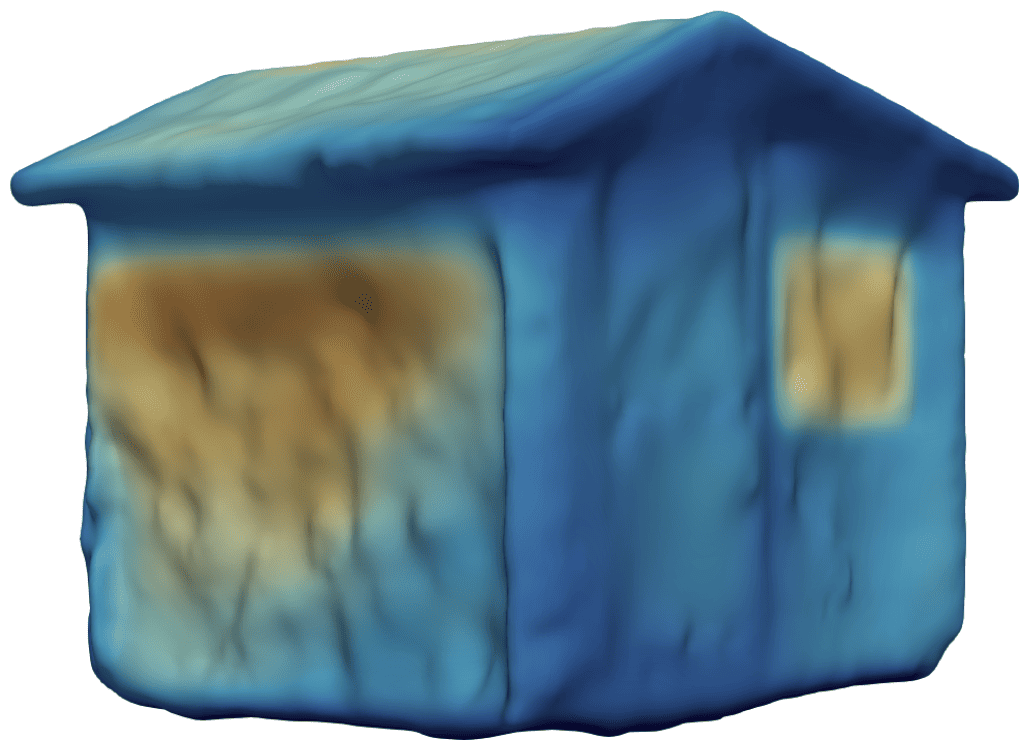}
                {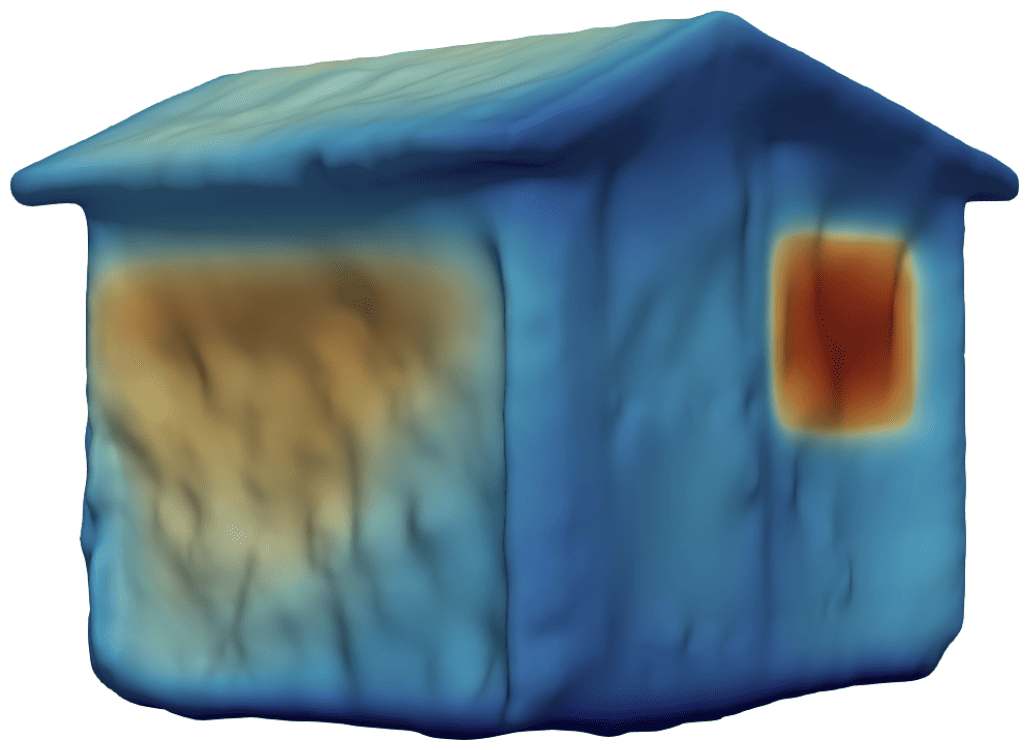}}
            {\simbar{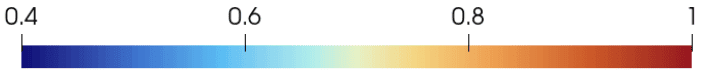}}%
            \hspace{\simgroupsep}%
            \simblock
            {\simimgs
                {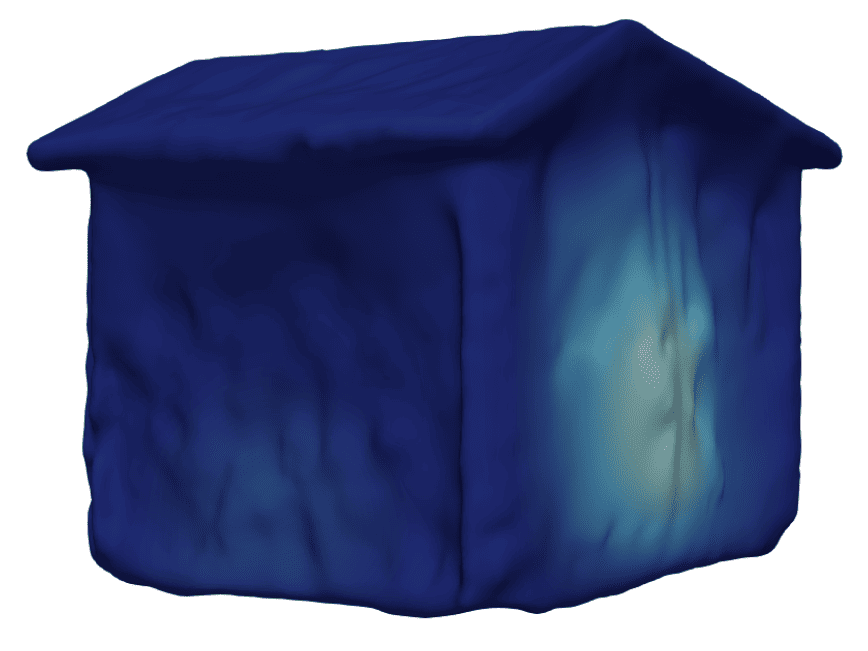}
                {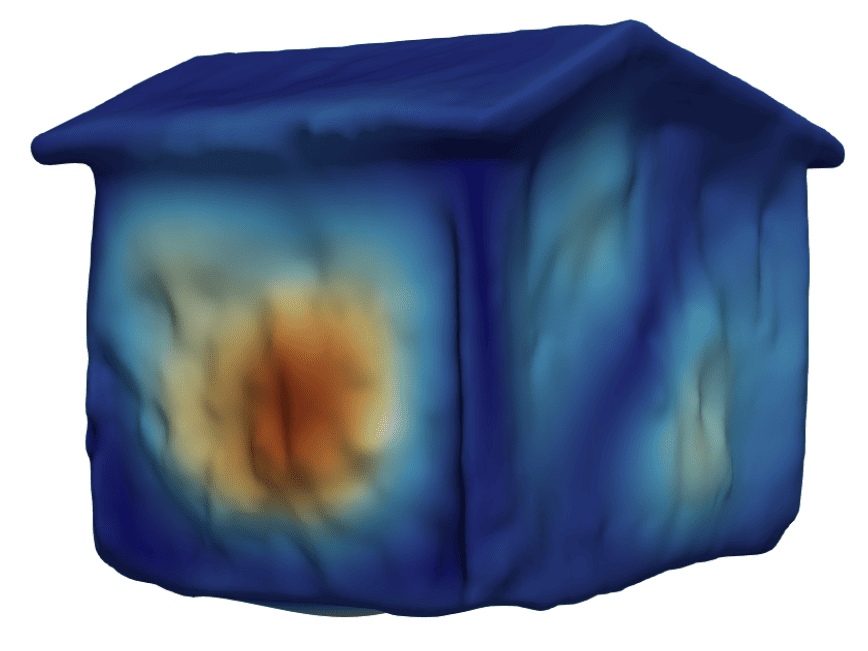}
                {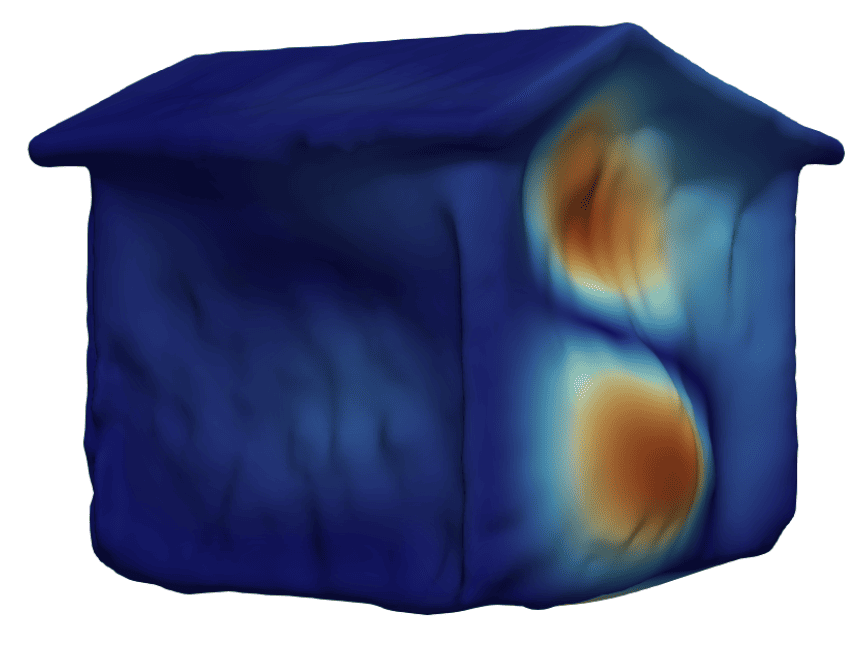}}
            {\simfreqs{$f_{6}=49.0$}{$f_{12}=101.0$}{$f_{15}=110.2$}}%
        }                                                                      \\ \simrowgap
        \simscenerow{Lion}{%
            \simblock
            {\simimgs
                {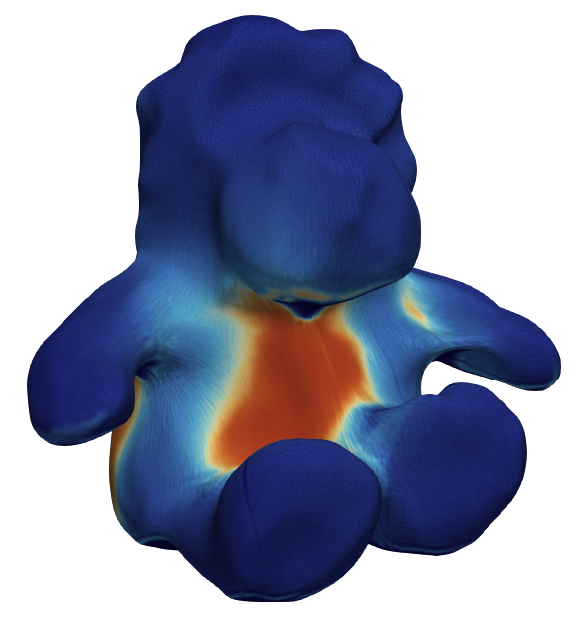}
                {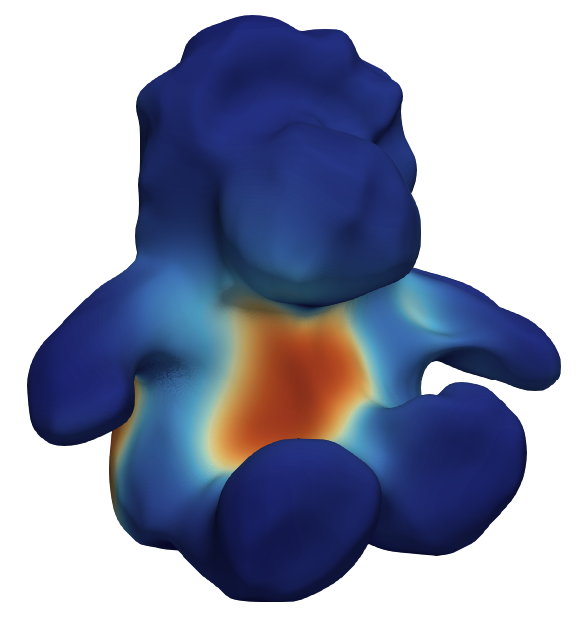}
                {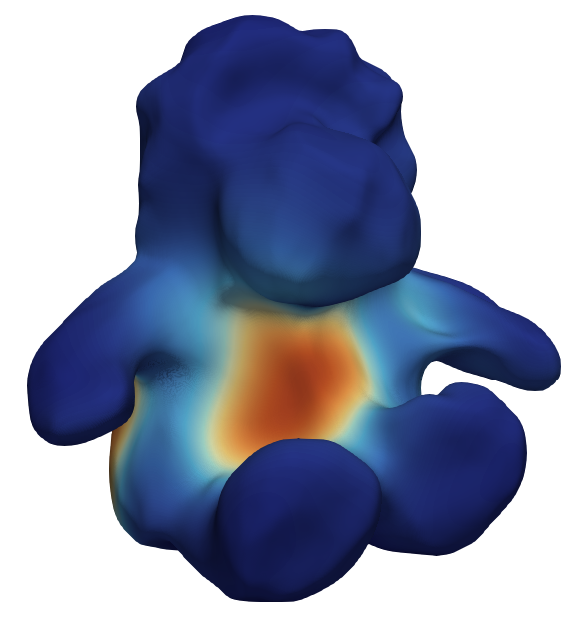}}
            {\simbar{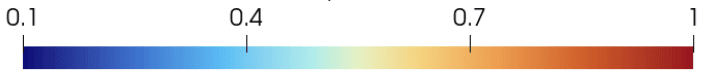}}%
            \hspace{\simgroupsep}%
            \simblock
            {\simimgs
                {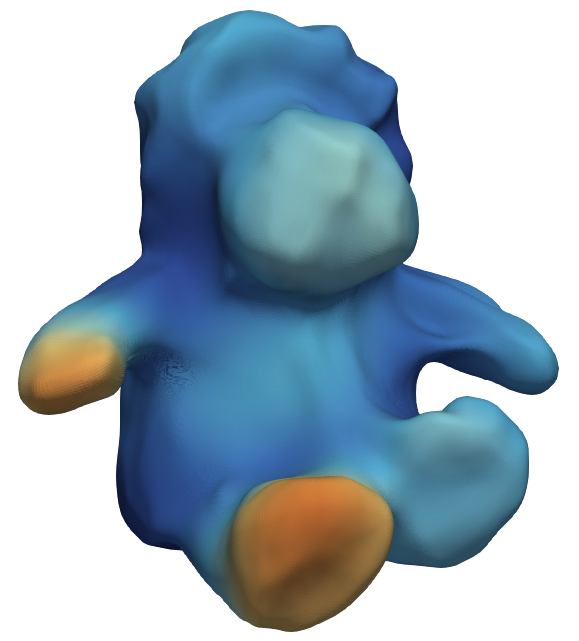}
                {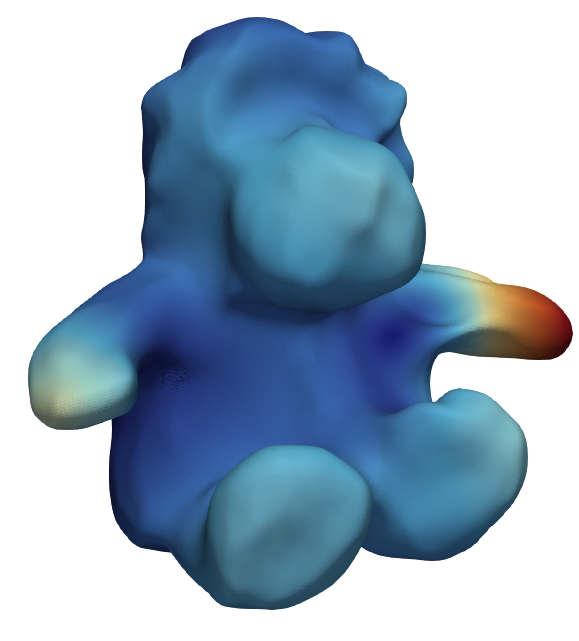}
                {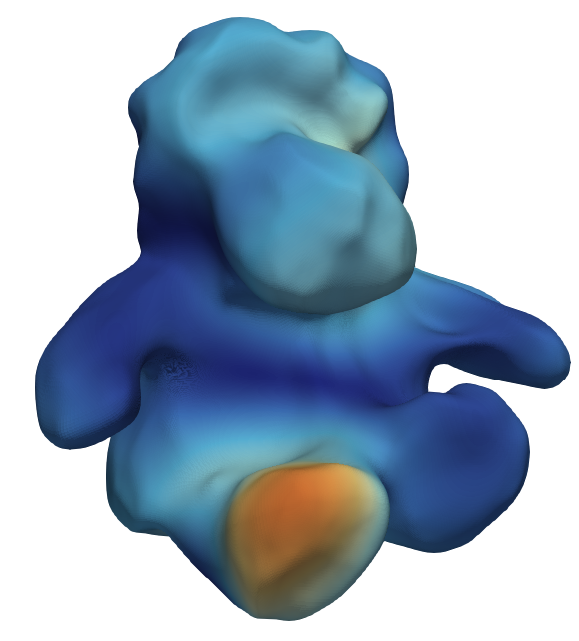}}
            {\simfreqs{$f_{6}=60.5$}{$f_{9}=87.2$}{$f_{14}=130.4$}}%
        }                                                                      \\
    \end{tabular}

    \caption{
        Thermal and modal simulation on the four real-world scenes, following the layout of \cref{fig:sim_synthetic}.
        \textbf{Heat}: initial thermal states captured via infrared thermography, followed by the simulated heat flow progression at half time and final time.
        \textbf{Modal}: the fundamental frequency ($f_6$) and two higher-order modes, reported in \si{\hertz}; on real-world data these additionally depend on the estimated metric scale (\cref{implementation:modal}).
    }
    \label{fig:sim_real}
\end{figure}

We further validate the pipeline on real-world datasets, shown in the heat columns of \cref{fig:sim_real}, with corresponding material properties listed in \cref{tab:scene_thermal_materials}.
Simulation settings were kept consistent across these scenes, incorporating heat generation in window regions.
The one exception is the plush toy model, which lacks windows, where we run single-material heat diffusion without heat sources.
For all real-world experiments, initial temperature fields were acquired via thermal imaging; in the Building~A scene, unseen regions were assigned a low normalized temperature ($0.1$), explaining the pronounced temperature gradient observed on the roof.
Across all scenes, the resulting simulations exhibit physically coherent thermal diffusion, closely matching the qualitative behavior expected from the underlying material and simulation assumptions.

\subsubsection{Modal Analysis Using Reissner-Mindlin Shell Theory}
\label{results:modalAnalysis}

To demonstrate the multi-physics capability of \methodname{} for simulation, we extend our evaluation to mechanical responses via modal analysis.
We employ the Reissner-Mindlin (RM) shell theory model, the standard for thin-to-moderately thick shells.

While conventional FEA is often susceptible to shear, bending, and membrane locking as thickness decreases, these phenomena are more tractable within the IGA framework~\cite{bensonIsogeometricShellAnalysis2010, guarinoIBCMShells2024}.
Specifically, our formulation leverages a global coordinate displacement and rotation field, utilizing IGA's higher-order continuity to maintain locking-free interior degrees of freedom~\cite{bensonIsogeometricShellAnalysis2010}.
Although the absence of strict $G^1$ continuity at patch boundaries may introduce localized spurious membrane stiffness, such penalty-coupled formulations are observed to converge under $h$-refinement~\cite{breitenbergerNonlinearShells2015, guarinoKLInteriorPenalty2024}, consistent with the convergence theory for Nitsche/penalty-type weak coupling of shells~\cite{benzakenNitscheKLShells2021}.

As for the thermal simulations, we first conduct experiments on textureless meshes, validating the results against ground-truth FEM simulation (\cref{fig:e2e}).
In \cref{tab:e2e_summary}, we report the \emph{median relative eigenfrequency error} over the $14$ elastic modes---rather than the maximum, since near-degenerate modes reorder between the two solvers---together with the \emph{mean best modal assurance criterion} (MAC), which quantifies mode-shape correlation ($1$ denotes an identical deformation pattern).
The recovered spectra match the ground truth to a median of $2.5$--$8.7\%$, with the per-mode error growing monotonically with reconstruction difficulty (Suzanne $2.5\%$, Bunny $5.4\%$, Spot $7.9\%$, Armadillo $8.7\%$)---mirroring the geometry error rather than any solver effect.
The end-to-end discrepancies reflect the quality of the geometry recovery, growing consistently with reconstruction error across every metric.

Correspondingly, the mode-shape agreement decreases monotonically with reconstruction error, from a MAC of $0.53$ (Suzanne) to $0.32$ (Armadillo).
The moderate MAC values largely reflect the mixing of near-degenerate, symmetry-related modes rather than lost shape information: small geometric perturbations rotate the basis within each nearly degenerate subspace, so an individual mode of one solver is recovered as a combination of the other's---bounding the attainable best-match MAC well below one even when the subspaces themselves coincide.
\methodname{} thus enables simulations close to the ground truth---with a trade-off between reconstruction quality and simulation fidelity.

We proceed to evaluate our method on the synthetic building dataset.
Synthetic datasets utilize known camera poses, allowing the reconstruction to recover the correct absolute geometric scale and yielding frequencies that align with the expected physical magnitudes for these structures~\cite{hansSeismicAnalysis2005}.
Since no ground clamping is prescribed, the global cantilever sway modes are relegated to the rigid-body spectrum.
Consequently, the lowest elastic eigenmodes---mostly in the $8$--$25$ \si{\hertz} range---correspond to the fundamental out-of-plane bending modes of the individual structural panels, such as walls.
As seen in the modal columns of \cref{fig:sim_synthetic}, our method recovers the resonance frequencies of the reconstructed objects without additional post-processing or specialized treatment.

Finally, we qualitatively evaluate our pipeline on real-world data.
In real-world scenarios, vision-only reconstructions suffer from scale ambiguity; they lack ground-truth extrinsic camera parameters and require a scaling factor to correctly recover the frequencies.
We estimate the scale factors to recover physically plausible results (see \cref{implementation:modal}).
The modal columns of \cref{fig:sim_real} present the results of the pipeline applied to the real-world dataset; we estimate plausible frequencies for all scenes.
These results demonstrate that geometry and fields reconstructed directly from images by our pipeline are of sufficient fidelity and regularity to support numerically stable, physically plausible simulation on real-world data---not just synthetic benchmarks---underscoring the practical viability of \methodname{} for downstream physics-based analysis.

\subsection{Reconstruction Quality and Novel View Synthesis}
\label{sec:results:nvs}

While state-of-the-art reconstruction methods prioritize visual fidelity, \methodname{} is designed for compatibility with CAD tooling and simulation frameworks.
Given that simulation accuracy is correlated with quality of the reconstructed geometry (as seen in the previous section), we benchmark our method against leading 3D scene reconstruction techniques.
Our results demonstrate that \methodname{} achieves comparable or superior performance to current state-of-the-art methods in both shape optimization and novel-view synthesis (NVS).

We compare shape optimization against Large Steps~\cite{nicoletLargeStepsInverse2021a}, and NVS capabilities against Nerfacto~\cite{tancikNerfstudio2023}, NeuS~\cite{wangNeuSLearningNeural2021}, and Gaussian Splatting~\cite{kerbl3DGaussianSplatting2023}.
Geometric accuracy is quantified via the Chamfer distance---the mean nearest-neighbor distance between the reconstructed and ground-truth surfaces (lower is better).
Rendering quality is assessed with three standard image metrics: the pixel-level Peak Signal-to-Noise Ratio (PSNR)~\cite{horeImageQualityMetrics2010} and the Structural Similarity index (SSIM)~\cite{wangImageQualityAssessment2004}, for which higher is better, and the deep-feature Learned Perceptual Image Patch Similarity (LPIPS)~\cite{zhangUnreasonableEffectivenessDeep2018}, for which lower is better.
For all NVS tasks, metrics are averaged across all test views.

\subsubsection{Shape Reconstruction}
\label{sec:subsec:shape_reconstruction}

Shape reconstruction quality is evaluated using the Chamfer distance on synthetic datasets with ground-truth 3D models.
The reconstructed spline surface is approximated by the rendering mesh, and the Chamfer distance between the reconstructed and ground truth meshes is computed by uniformly sampling $1{,}000{,}000$ points across the surface of each mesh.
Our implementation of Large Steps is adapted to use the same cubed-sphere topology and regular meshes for each of the 6 faces, with the same refinement schedule.

As shown in \cref{tab:results:reconstruction}, \methodname{} achieves a significant reduction in Chamfer distance compared to Large Steps across the majority of evaluated datasets.
This observation even includes the challenging Armadillo model, where it outperforms Large Steps by close to an order of magnitude ($0.035$ vs. $0.337$).
This gap stems from the degradation of the combinatorial-Laplacian preconditioner in Large Steps when encountering stretched triangles.
While Large Steps maintains an advantage on the Spot model ($0.0065$ vs. $0.019$ for \methodname{}), our method demonstrates superior overall geometric precision.
This performance gain is primarily attributed to our spline-based representation, which provides inherent structural regularization, bypassing the need for gradient preconditioning required by previous works~\cite{nicoletLargeStepsInverse2021a, worchelDifferentiableRenderingParametric2023}.
This increased stability further enables the use of the more robust L-BFGS optimizer over Adam---a benefit not available for methods requiring gradient preconditioning.

Overall, our findings indicate that \methodname{} outperforms current state-of-the-art shape optimization methods by providing superior geometric accuracy and increased optimization stability, even in the presence of complex geometries.

\begin{table}[t]
    \centering
    \caption{
        Quantitative comparison of reconstruction accuracy via Chamfer distance.
        \methodname{} achieves lower Chamfer distances than Large Steps on all objects except Spot, where Large Steps performs better.
        Our proposed strategy successfully reconciles high-fidelity reconstruction with surface smoothness and watertightness, providing an accurate reconstruction compatible with downstream simulation.
    }
    \label{tab:results:reconstruction}
    \begin{tabular}{lcccccccc}
        \toprule
                    & \multicolumn{4}{c}{BuildNet3D} & \multicolumn{4}{c}{Textureless meshes}                                                                                                        \\
        \cmidrule(lr){2-5}\cmidrule(lr){6-9}
        Method      & 1                              & 2                                      & 3              & 4              & Suzanne        & Bunny          & Spot            & Armadillo      \\\midrule
        Ours        & \textbf{0.167}                 & \textbf{0.322}                         & \textbf{0.094} & \textbf{0.096} & \textbf{0.001} & \textbf{0.012} & 0.019           & \textbf{0.035} \\
        Large Steps & 0.434                          & 0.460                                  & 0.115          & 0.140          & 0.016          & 0.088          & \textbf{0.0065} & 0.337          \\
        \bottomrule
    \end{tabular}
\end{table}

\begin{table}[t]
    \centering
    \caption{Quantitative evaluation of NVS performance of \methodname{} against state-of-the-art methods.
        Best and second-best results are indicated in \textbf{bold} and \underline{underlined}, respectively.
        For the Lion and Car datasets, GS uses full images with background to ensure reconstruction stability (failing to reconstruct otherwise).
        \methodname{} consistently achieves competitive or leading results across both BuildNet3D and real-world datasets.
        The slight decrease in performance from the scenes with known poses (BuildNet3D) to the real-world scenes
        suggests that further performance gains may be driven by advancements in pose estimation rather than reconstruction techniques.
    }
    \label{tab:results:nvs}
    \begin{tabular}{lllccccc}
        \toprule
         & Datasets                                  & Metric               & Ours               & Large Steps        & Nerfacto           & NeuS              & GS                \\ \midrule

        \multirow{12}{*}{\rotatebox{90}{BuildNet3D}}
         & \multirow{3}{*}{Building 1}               & PSNR $\uparrow$      & \textbf{26.498}    & \underline{22.964} & 19.703             & 19.871            & 20.321            \\
         &                                           & SSIM $\uparrow$      & \textbf{0.923}     & \underline{0.825}  & 0.648              & 0.672             & 0.658             \\
         &                                           & LPIPS  $\downarrow$  & \textbf{0.068}     & 0.146              & 0.175              & \underline{0.090} & 0.123             \\
        \cmidrule(lr){2-8}
         & \multirow{3}{*}{Building 2}               & PSNR  $\uparrow$     & \textbf{25.897}    & \underline{23.059} & 17.837             & 18.040            & 16.978            \\
         &                                           & SSIM $\uparrow$      & \textbf{0.917}     & \underline{0.900}  & 0.710              & 0.716             & 0.653             \\
         &                                           & LPIPS   $\downarrow$ & \textbf{0.079}     & \underline{0.097}  & 0.266              & 0.124             & 0.364             \\
        \cmidrule(lr){2-8}
         & \multirow{3}{*}{Building 3}               & PSNR $\uparrow$      & \underline{29.739} & \textbf{29.814}    & 23.031             & 23.229            & 23.956            \\
         &                                           & SSIM  $\uparrow$     & \textbf{0.941}     & \underline{0.937}  & 0.758              & 0.749             & 0.759             \\
         &                                           & LPIPS  $\downarrow$  & \underline{0.053}  & \textbf{0.049}     & 0.122              & \underline{0.053} & 0.102             \\
        \cmidrule(lr){2-8}
         & \multirow{3}{*}{Building 4}               & PSNR $\uparrow$      & \textbf{26.078}    & \underline{23.046} & 19.720             & 19.821            & 20.415            \\
         &                                           & SSIM   $\uparrow$    & \textbf{0.920}     & \underline{0.899}  & 0.617              & 0.635             & 0.674             \\
         &                                           & LPIPS  $\downarrow$  & \textbf{0.071}     & \underline{0.088}  & 0.219              & 0.101             & 0.181             \\
        \midrule
        \multirow{12}{*}{\rotatebox{90}{Real-world}}
         & \multirow{3}{*}{Building A}
         & PSNR $\uparrow$                           & 20.163               & 17.537             & 19.256             & \underline{20.943} & \textbf{24.782}                       \\
         &                                           & SSIM  $\uparrow$     & \underline{0.771}  & 0.733              & 0.430              & 0.689             & \textbf{0.800}    \\
         &                                           & LPIPS  $\downarrow$  & \textbf{0.179}     & \underline{0.233}  & 0.543              & 0.295             & \underline{0.233} \\
        \cmidrule(lr){2-8}
         & \multirow{3}{*}{Car}
         & PSNR  $\uparrow$                          & 18.897               & 18.972             & \underline{19.591} & \textbf{20.996}    & 19.255                                \\
         &                                           & SSIM  $\uparrow$     & \underline{0.797}  & 0.796              & 0.662              & \textbf{0.820}    & 0.540             \\
         &                                           & LPIPS   $\downarrow$ & \textbf{0.159}     & \underline{0.166}  & 0.302              & 0.169             & 0.334             \\
        \cmidrule(lr){2-8}
         & \multirow{3}{*}{Woodshed}
         & PSNR   $\uparrow$                         & 22.335               & 22.021             & 25.722             & \underline{26.965} & \textbf{29.235}                       \\
         &                                           & SSIM $\uparrow$      & 0.881              & 0.861              & 0.730              & \underline{0.888} & \textbf{0.920}    \\
         &                                           & LPIPS  $\downarrow$  & \textbf{0.096}     & 0.118              & 0.150              & 0.108             & \underline{0.098} \\
        \cmidrule(lr){2-8}
         & \multirow{3}{*}{Lion}                     & PSNR $\uparrow$      & \underline{27.327} & 27.042             & \textbf{28.368}    & 11.889            & 23.454            \\
         &                                           & SSIM  $\uparrow$     & \textbf{0.959}     & \underline{0.955}  & 0.809              & 0.507             & 0.792             \\
         &                                           & LPIPS $\downarrow$   & \textbf{0.033}     & \underline{0.035}  & 0.096              & 0.543             & 0.181             \\
        \bottomrule
    \end{tabular}
\end{table}

\subsubsection{Novel View Synthesis}\label{sec:subsec:novel_view_synthesis}

\newif\ifnvsmasked \nvsmaskedfalse
\newif\ifnvsraw    \nvsrawtrue
\newcommand{\nvsfigref}{\ifnvsmasked\cref{fig:result:nvs}\else\cref{fig:result:nvs_raw}\fi}

\newlength{\nvsrowlab}\setlength{\nvsrowlab}{0.030\textwidth}%
\newlength{\nvscolgap}\setlength{\nvscolgap}{0.008\textwidth}%
\newlength{\nvsrefgap}\setlength{\nvsrefgap}{0.022\textwidth}%
\newlength{\nvsrowsep}\setlength{\nvsrowsep}{2pt}%
\newlength{\nvscellw}%
\setlength{\nvscellw}{\dimexpr(\textwidth-\nvsrowlab-\nvsrefgap-\nvscolgap*5)/6\relax}%
\newcommand{\nvsimage}[1]{\includegraphics[width=\linewidth]{#1}}
\newcommand{\nvsrow}[7]{%
    \begin{subfigure}[c]{\nvsrowlab}\centering\rotatebox{90}{\small #1}\end{subfigure}\hspace{\nvscolgap}%
    \begin{subfigure}[c]{\nvscellw}\centering\nvsimage{#2}\end{subfigure}\hspace{\nvsrefgap}%
    \begin{subfigure}[c]{\nvscellw}\centering\nvsimage{#3}\end{subfigure}\hspace{\nvscolgap}%
    \begin{subfigure}[c]{\nvscellw}\centering\nvsimage{#4}\end{subfigure}\hspace{\nvscolgap}%
    \begin{subfigure}[c]{\nvscellw}\centering\nvsimage{#5}\end{subfigure}\hspace{\nvscolgap}%
    \begin{subfigure}[c]{\nvscellw}\centering\nvsimage{#6}\end{subfigure}\hspace{\nvscolgap}%
    \begin{subfigure}[c]{\nvscellw}\centering\nvsimage{#7}\end{subfigure}%
}
\newcommand{\nvsrawrow}[2]{%
    \nvsrow{#1}{images/nvs_raw/#2_reference.png}{images/nvs_raw/#2_ours.png}%
    {images/nvs_raw/#2_large_steps.png}{images/nvs_raw/#2_neus.png}%
    {images/nvs_raw/#2_nerfacto.png}{images/nvs_raw/#2_splatfacto.png}%
}
\newcommand{\nvsheader}{%
    \makebox[\nvsrowlab]{}\hspace{\nvscolgap}%
    \makebox[\nvscellw][c]{\small Reference}\hspace{\nvsrefgap}%
    \makebox[\nvscellw][c]{\small Ours}\hspace{\nvscolgap}%
    \makebox[\nvscellw][c]{\small Large Steps}\hspace{\nvscolgap}%
    \makebox[\nvscellw][c]{\small NeuS}\hspace{\nvscolgap}%
    \makebox[\nvscellw][c]{\small Nerfacto}\hspace{\nvscolgap}%
    \makebox[\nvscellw][c]{\small 3DGS}%
}

\ifnvsmasked
\begin{figure}[!htbp]
    \centering
    \nvsheader

    \vspace{0.12cm}
    \nvsrow{Building 1}{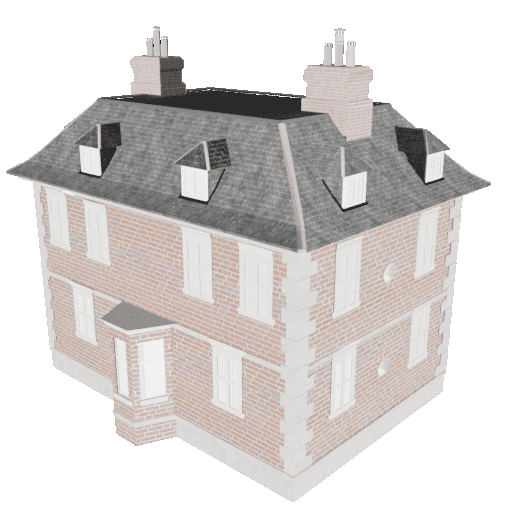}{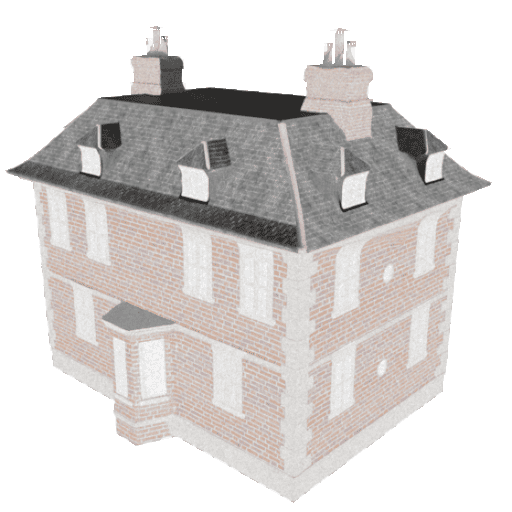}{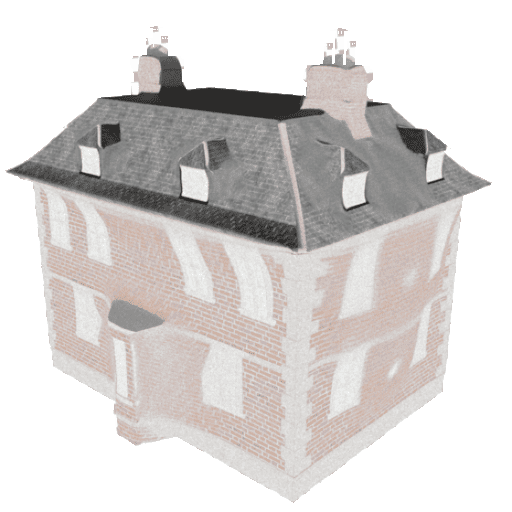}{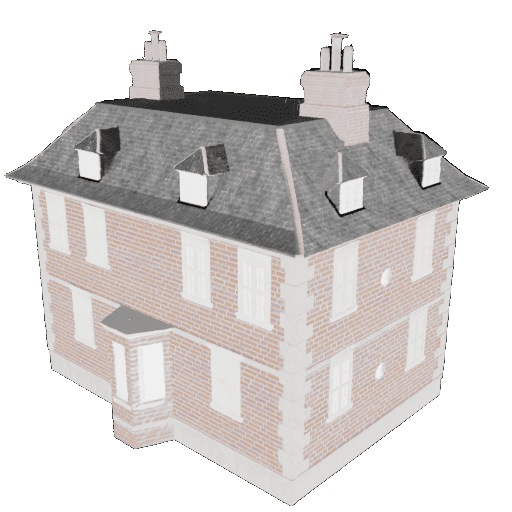}{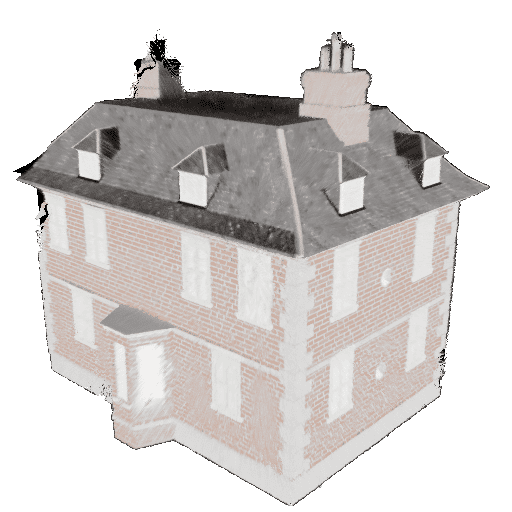}{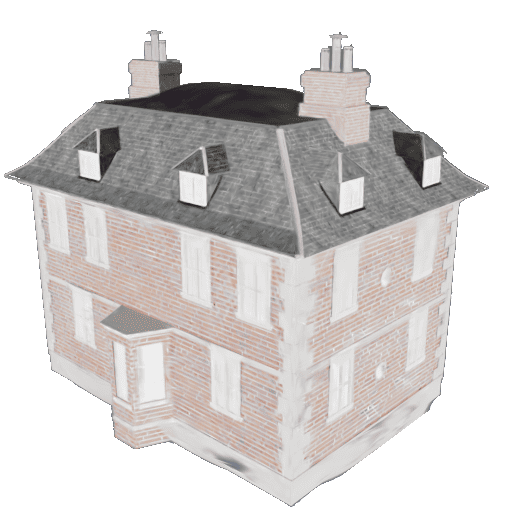}

    \vspace{\nvsrowsep}
    \nvsrow{Building 2}{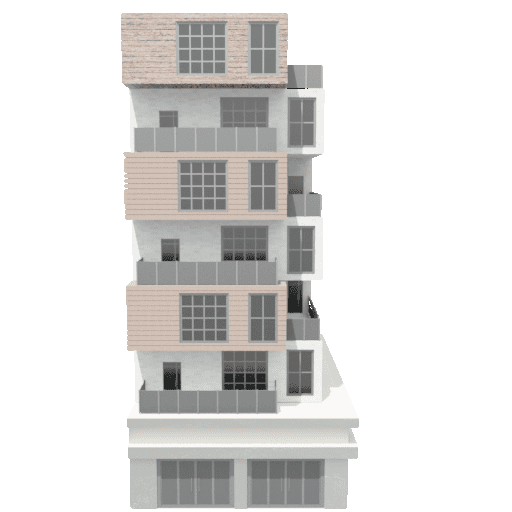}{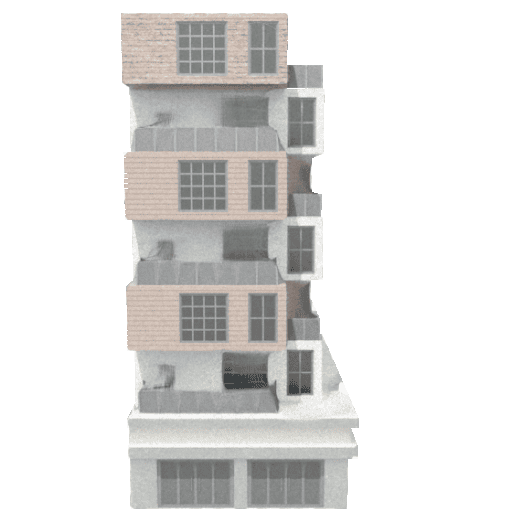}{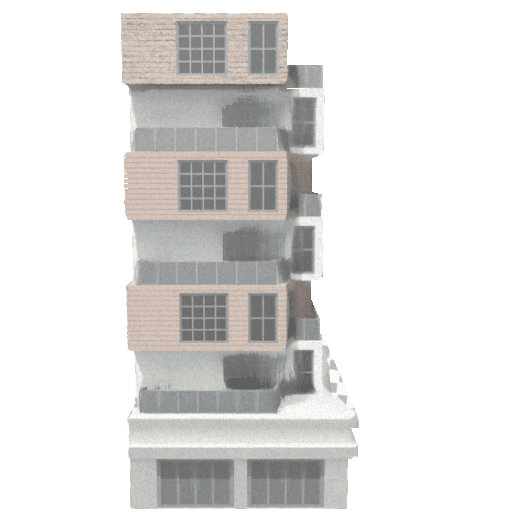}{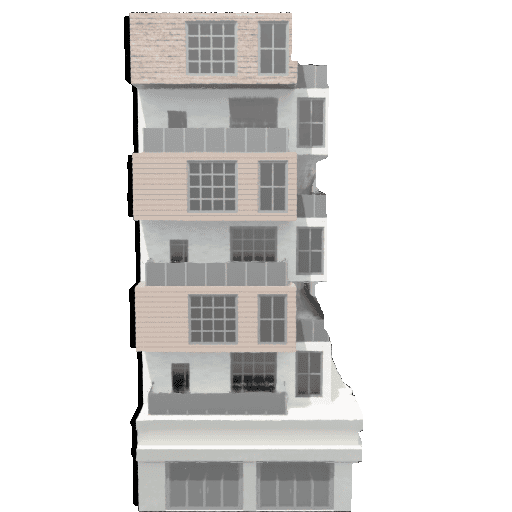}{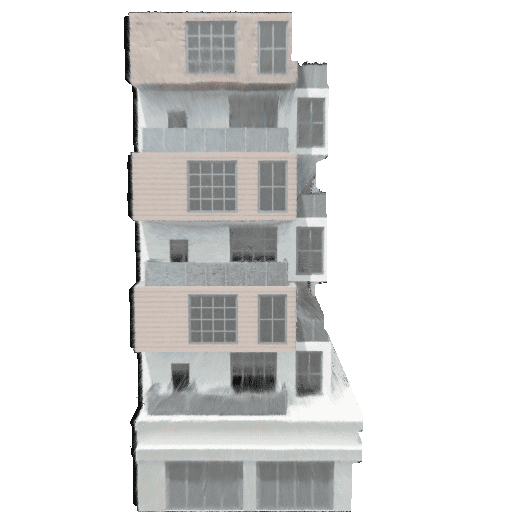}{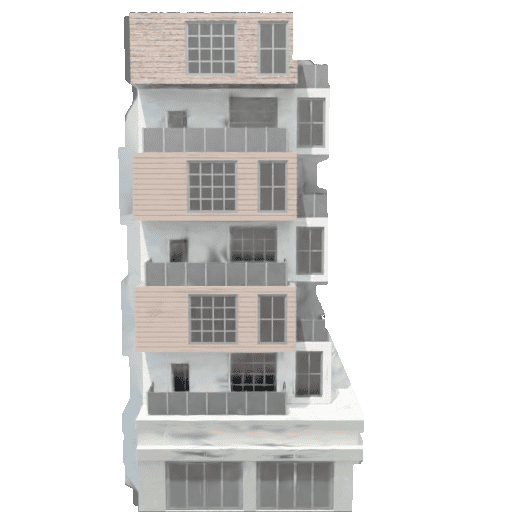}

    \vspace{\nvsrowsep}
    \nvsrow{Building 3}{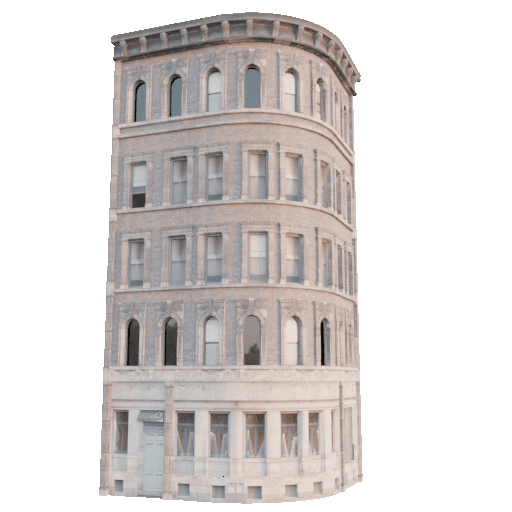}{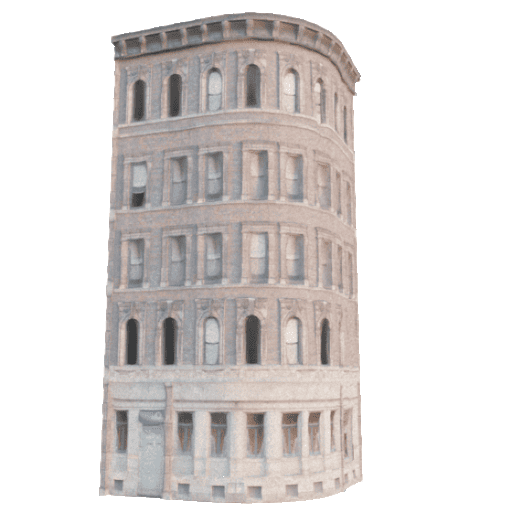}{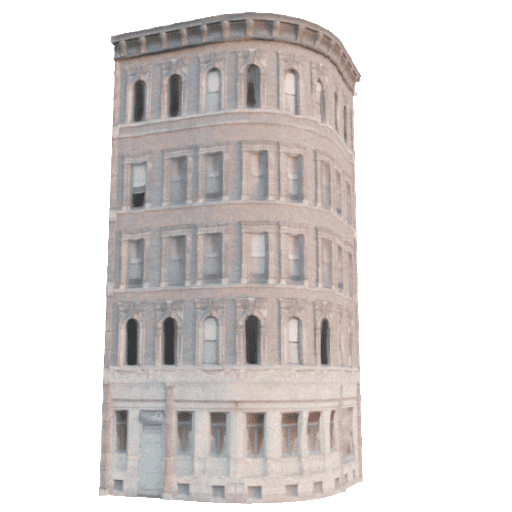}{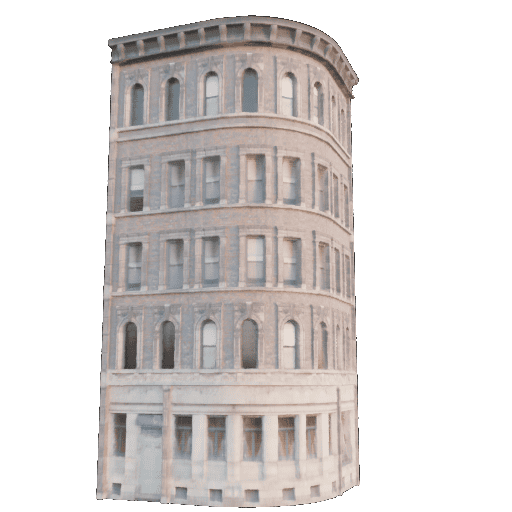}{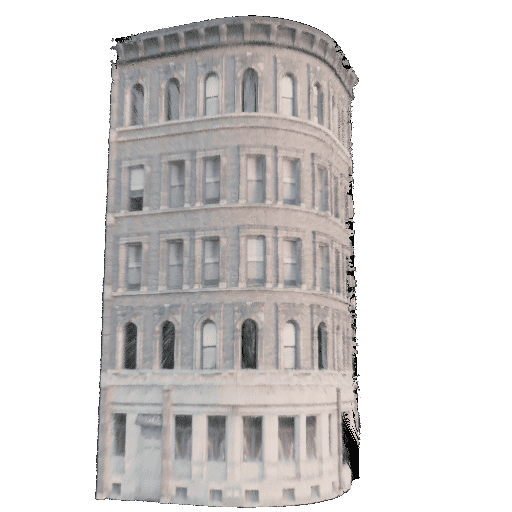}{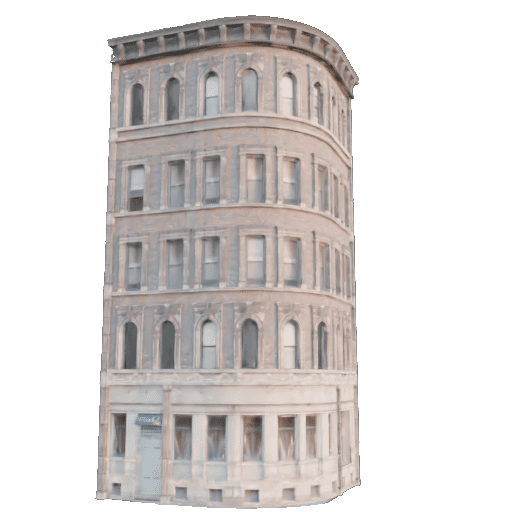}

    \vspace{\nvsrowsep}
    \nvsrow{Building 4}{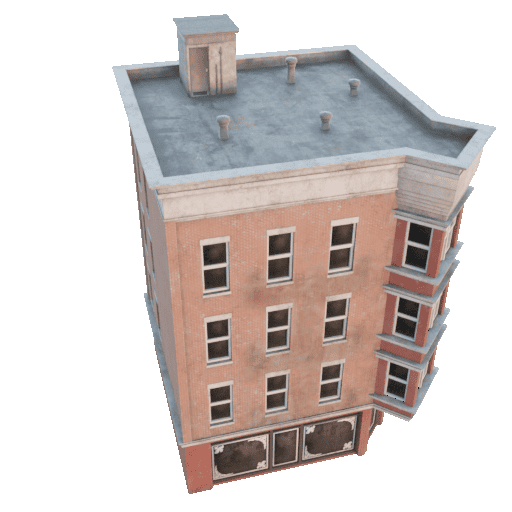}{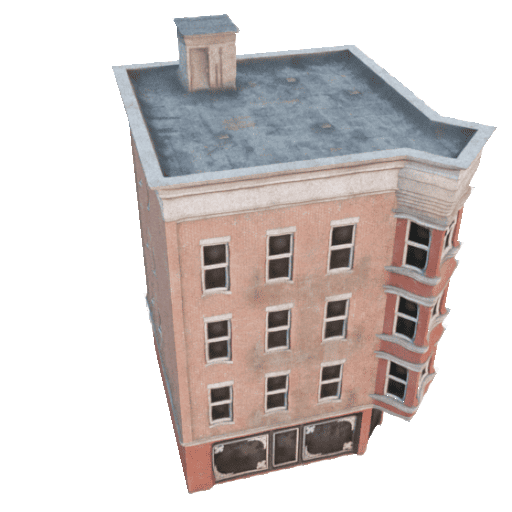}{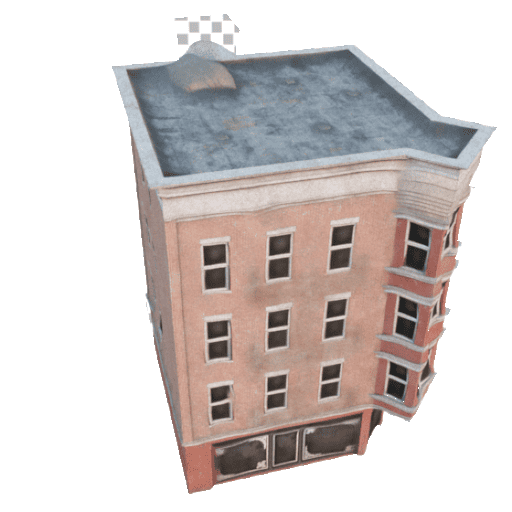}{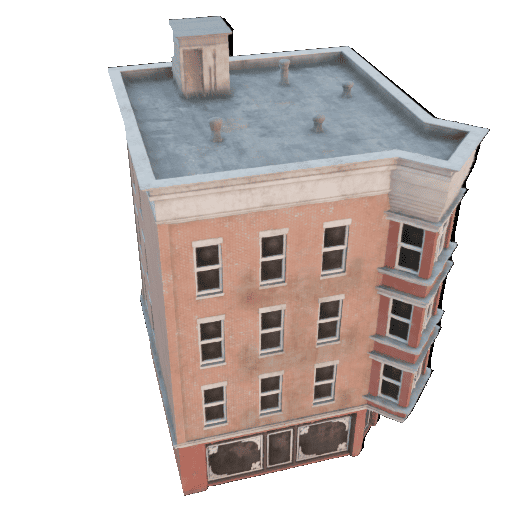}{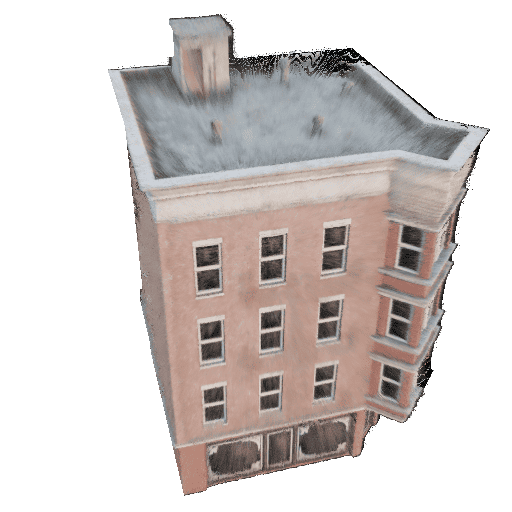}{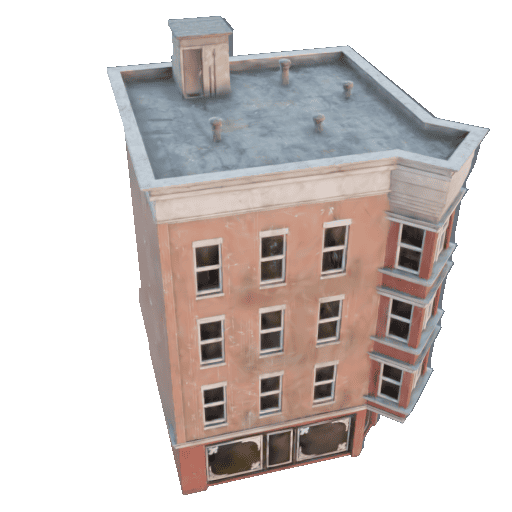}

    \vspace{\nvsrowsep}
    \nvsrow{Car}{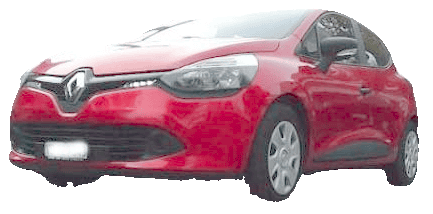}{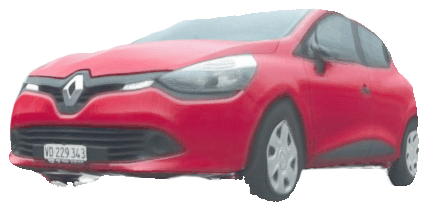}{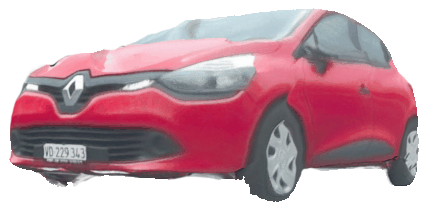}{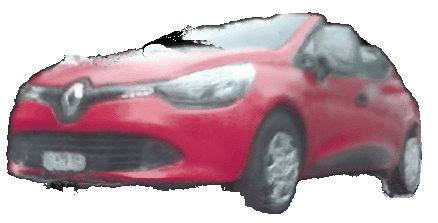}{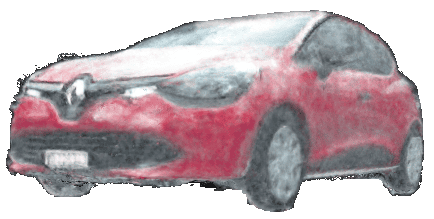}{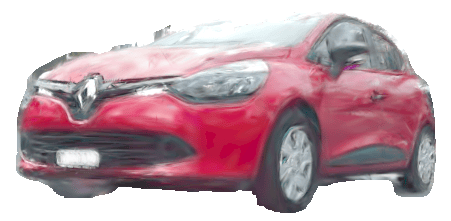}

    \vspace{\nvsrowsep}
    \nvsrow{Woodshed}{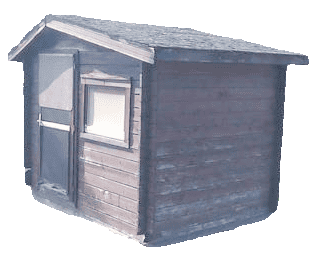}{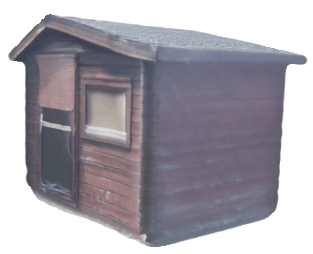}{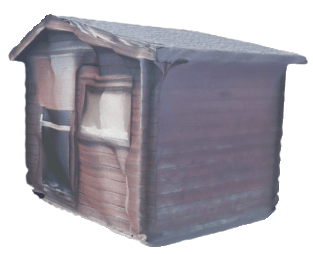}{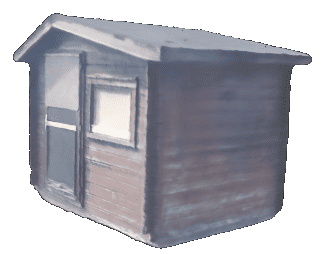}{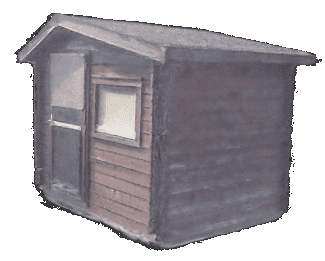}{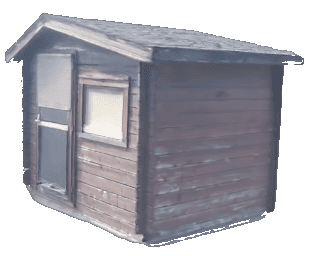}

    \vspace{\nvsrowsep}
    \nvsrow{Lion}{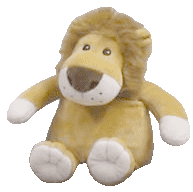}{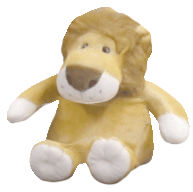}{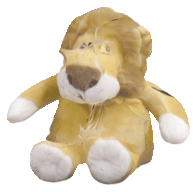}{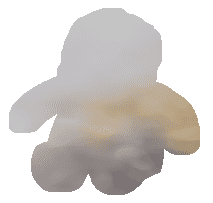}{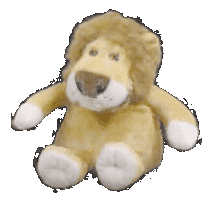}{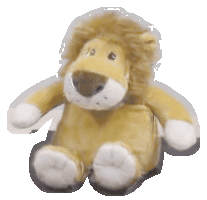}

    \vspace{\nvsrowsep}
    \nvsrow{Building A}{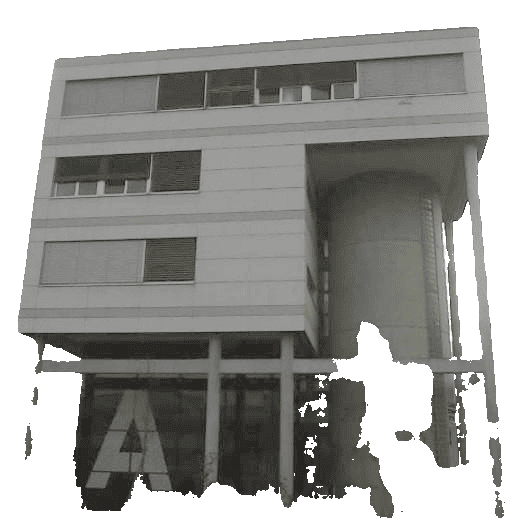}{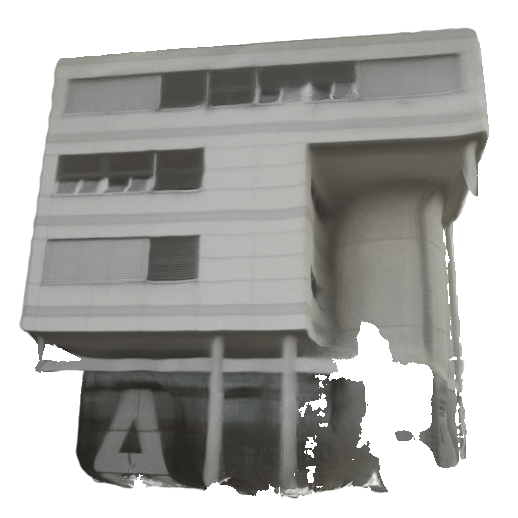}{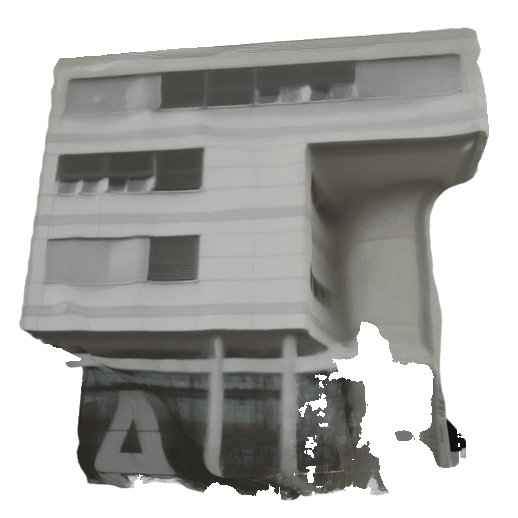}{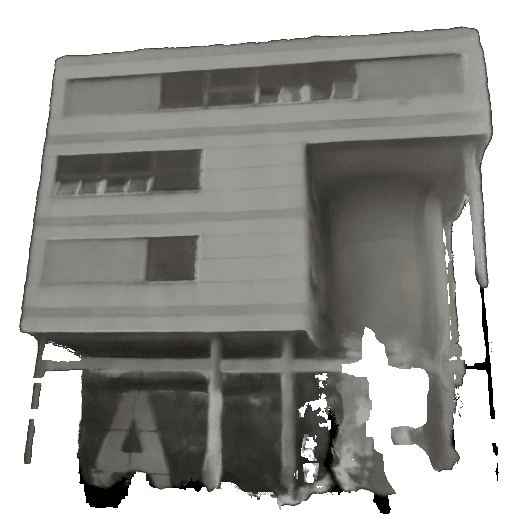}{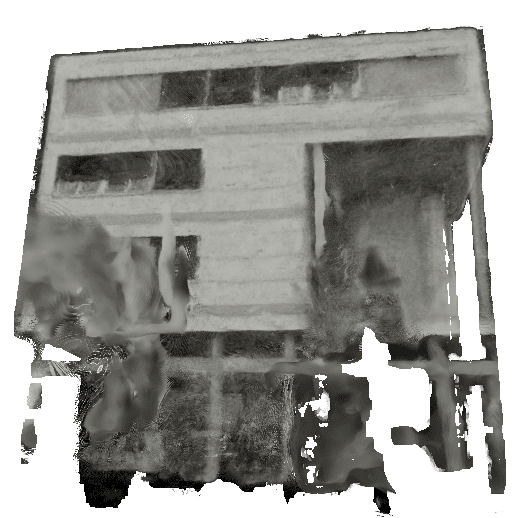}{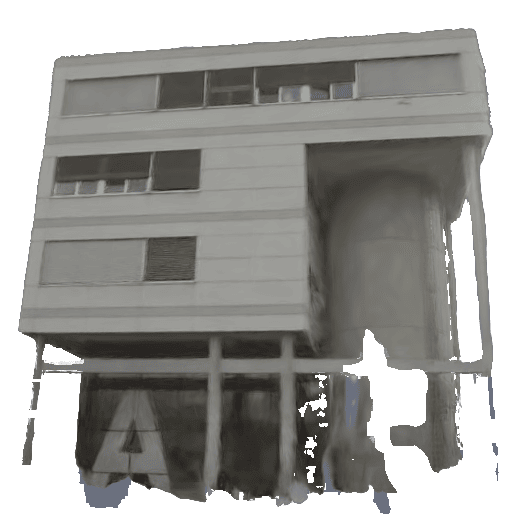}

    \vspace{3pt}
    \caption{
        Visual assessment of reconstructed geometries.
        Every method is cut to the segmentation mask the quantitative evaluation is computed against, so all six columns share one silhouette and missing geometry shows as a gap in it.
        \methodname{} demonstrates superior completion while maintaining watertightness and topological continuity even when occlusions are present (e.g., Building A).
    }
    \label{fig:result:nvs}
\end{figure}
\fi

\ifnvsraw
\begin{figure}[!htbp]
    \centering
    \nvsheader

    \vspace{0.12cm}
    \nvsrawrow{Building 1}{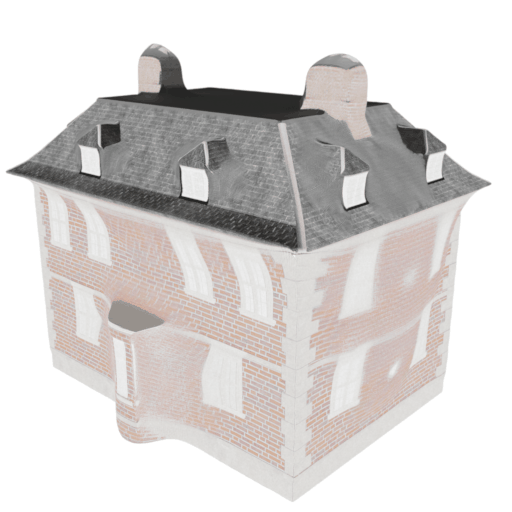}

    \vspace{\nvsrowsep}
    \nvsrawrow{Building 2}{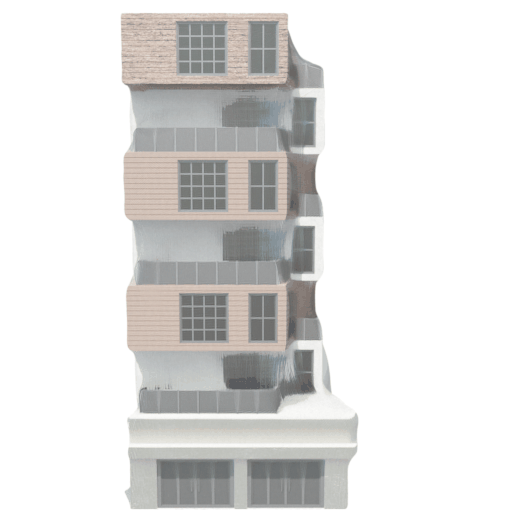}

    \vspace{\nvsrowsep}
    \nvsrawrow{Building 3}{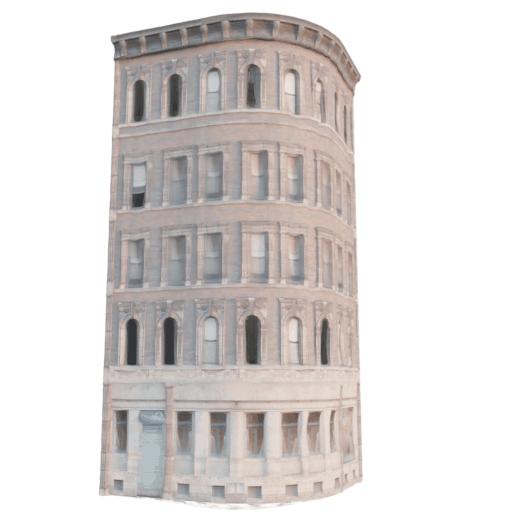}

    \vspace{\nvsrowsep}
    \nvsrawrow{Building 4}{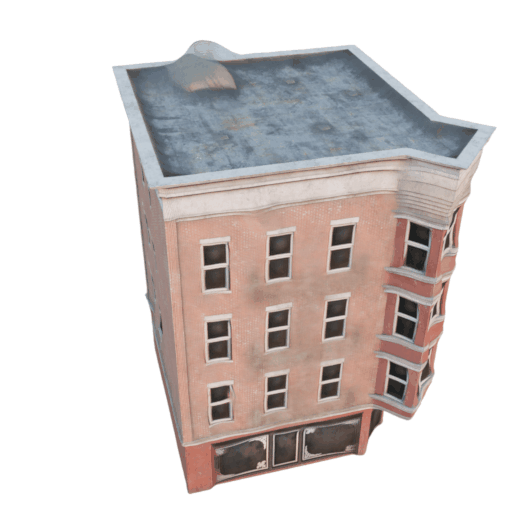}

    \vspace{\nvsrowsep}
    \nvsrawrow{Car}{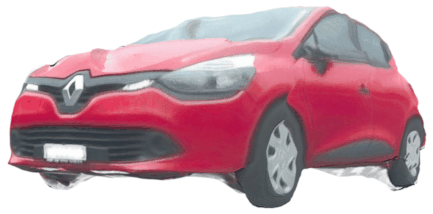}

    \vspace{\nvsrowsep}
    \nvsrawrow{Woodshed}{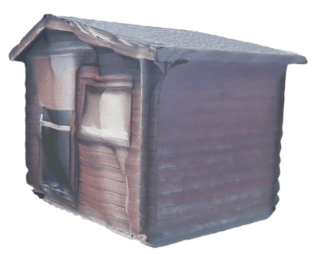}

    \vspace{\nvsrowsep}
    \nvsrawrow{Lion}{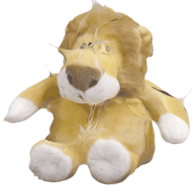}

    \vspace{\nvsrowsep}
    \nvsrawrow{Building A}{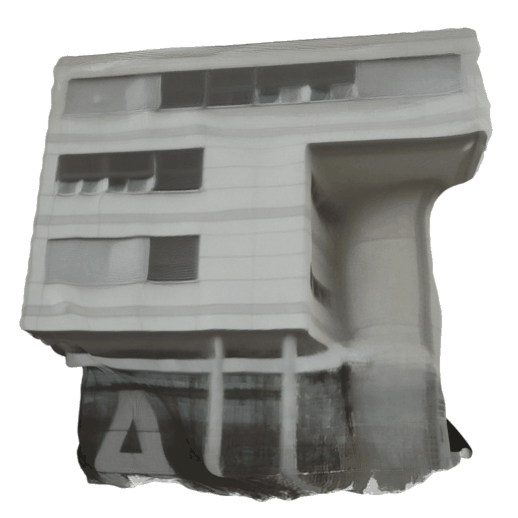}

    \vspace{3pt}
    \caption{
        Visual assessment of reconstructed geometries---renderings are shown as each method produces them, without background removal.
        \methodname{} demonstrates superior completion while maintaining watertightness and topological continuity even when occlusions are present (e.g., Building A), whereas the volumetric baselines deposit floating density around the object and NeuS fails to converge on Lion.
    }
    \label{fig:result:nvs_raw}
\end{figure}
\fi

Beyond geometric accuracy, we evaluate the visual fidelity of \methodname{} through novel-view synthesis (NVS), generating realistic renderings from unobserved camera poses.
Quantitative comparisons between our approach and established baselines are presented in \cref{tab:results:nvs}, while qualitative evaluations of rendered images are provided in \nvsfigref{}.
\ifnvsmasked\ifnvsraw
\cref{fig:result:nvs_raw} repeats that comparison without background removal, showing what each method renders before the evaluation mask is applied.
\fi\fi

In NVS, \methodname{} demonstrates performance comparable or superior to Large Steps and dedicated NVS methods.
In scenarios with known poses, such as the BuildNet3D dataset, \methodname{} outperforms all baseline models, with the exception of the Building 3 scene, where it closely trails Large Steps.
In real-world scenes utilizing VGGT-estimated poses, \methodname{} ranks among the top performers, often securing first or second place.
Furthermore, \methodname{} achieves the best LPIPS scores on most tested scenes, including real-world environments, indicating strong perceptual fidelity.
The improvements brought on are visible in \nvsfigref{} where one can see that \methodname{} reconstructs a more detailed representation with sharper features on the synthetic and real-world scenes.
Notably, in the Building A scene, our representation remains watertight: whereas NeuS, Nerfacto, and 3DGS replicate the holes left in the training images by occlusion removal, \methodname{} ensures geometric completeness while capturing more granular detail than Large Steps---specifically in the column, which is truncated in the Large Steps reconstruction.

Given that our method is based on shape optimization, a notable limitation is the potential loss of topological features, such as holes, present in the original scene.
While these limitations also manifest in Large Steps, alternative frameworks---such as Gaussian Splatting or Neural Radiance Fields (NeRF)---avoid these specific issues at the expense of integration with downstream Finite Element Analysis (FEA) workflows.

\section{Discussion}
\label{sec_discussion}

We presented \methodname{}, a framework to reconstruct spline-based boundary representations from images, such that the reconstructed model can serve simultaneously as a simulation-ready computational domain and a vision-based reconstruction target.
For the two best-reconstructed shapes in our experiments (Suzanne and Bunny), IGA-based heat and modal simulations run directly on the recovered geometry agree with FEM-based simulations on the ground-truth mesh to within 1.07--1.41\% in field discrepancy ${FL}^2$ (floor-corrected, at final time), 0.72--0.77\% in the geometry-intrinsic relaxation rate $\lambda_1$, and 2.47--5.35\% in the median eigenfrequency.
Errors grow smoothly rather than catastrophically as reconstruction difficulty increases: for the Armadillo, whose thin limbs are the hardest of the four shapes to recover, the field and relaxation-rate errors rise to 6.89\% and 19.85\%, and the eigenfrequency error to 8.72\%, yet these errors remain stable over the simulated time horizon (an initial-state thermal-energy error of 5.26\% is essentially unchanged at 5.1\% by the final step).
This pattern---simulation error tracking reconstruction error rather than diverging independently---indicates that the geometric recovery step, not the simulation formulation, is the binding constraint on end-to-end fidelity.

This outcome follows from a specific computational choice: representing geometry, appearance, and physical fields in the same basis that both the differentiable renderer and the isogeometric solver consume directly.
Because a point on a cubic B-spline patch depends on up to sixteen control points rather than the three vertices of a mesh triangle, gradients are distributed over a much larger neighborhood than in mesh-based inverse rendering; this proved sufficient to stabilize a quasi-Newton optimizer (L-BFGS) without the graph-Laplacian preconditioning that mesh-based shape optimization otherwise requires.
The same basis then propagates forward: material and thermal fields inpainted onto the spline are consumed unmodified by the IGA stiffness and mass assembly, while the same basis-function evaluations, control points, and tensor primitives used during optimization are reused directly by the IGA solver, with no intermediate mesh export or format conversion.

To our knowledge, \methodname{} is the first framework to demonstrate shape optimization of a multi-patch B-spline surface from images, and the first to run simulation directly on the resulting reconstruction.
By enabling the direct generation of simulation-ready 3D models from sparse image data, \methodname{} reduces the computational and manual overhead currently associated with manual CAD modeling and mesh generation.
While we evaluate the framework for heat-transfer and modal analysis, we expect the underlying principle to extend beyond these two applications, to any setting where a data-driven reconstruction must ultimately feed a PDE-constrained downstream computation---for instance, thermodynamic analyses such as aerodynamic cooling and insulation.

A natural assumption is that imposing these constraints trades visual fidelity for simulation compatibility.
We do not observe this trade-off: \methodname{} achieves a lower Chamfer distance than Large Steps, a leading mesh-based shape-optimization method, on seven of the eight objects with known ground-truth topology---by close to an order of magnitude on the Armadillo (0.035 versus 0.337)---and matches or achieves the best LPIPS score among all tested baselines on seven of the eight test scenes.
Together, these results suggest that the constraints required for simulation---watertightness, patch continuity, curvature regularity---are compatible with, and in some respects reinforce, the constraints required for accurate visual reconstruction, at least for the object classes tested here.

Despite these advantages, certain limitations necessitate further research.
First, the six-patch cubed-sphere parametrization restricts reconstructions to genus-0 solids: through-holes, handles, and multi-body assemblies cannot currently be represented, limiting direct applicability to a subset of real geometries.
Second, the current reliance on B-splines can lead to surface over-smoothing, potentially removing fine-scale geometric features---details that can be critical in applications such as infrastructure analysis or industrial quality control; adaptive representations such as T-splines~\cite{sederbergTSplines2003} or hierarchical B-splines~\cite{gianelliHierarchicalBSplines2012} are a possible research direction to tackle this problem.
Third, on real-world scenes, camera poses and absolute scale are estimated rather than known---poses come from VGGT and carry its errors, and metric scale is currently fixed via a single estimated calibration factor per scene rather than recovered automatically---a practical barrier to a fully autonomous digitize-and-simulate workflow.
Finally, the current per-scene optimization time (4--6 hours on a single workstation GPU) is a practical constraint for time-sensitive use cases, though not a fundamental limitation of the approach.
Addressing pose and scale robustness---e.g., through cross-modal fusion of RGB, thermal, and inertial measurements~\cite{skorokhodovSEAR_2026, camposORBSLAM32021}---and extending the geometric representation to adaptive, trimmed, or higher-genus families are the most direct paths to closing these gaps.

We see the methodological innovation demonstrated here---recovering a compact geometric and physical representation directly from observations, in a form a physics solver can consume without intermediate translation---as a pattern with relevance beyond digital-twin construction, wherever a computational pipeline must go directly from sparse, noisy observations to a representation constrained by downstream physics.
More broadly, when embedded within existing scientific and industrial workflows, frameworks that remove the manual CAD-reconstruction and mesh-repair steps currently separating image capture from simulation can democratize high-impact applications, from predictive maintenance of civil infrastructure~\cite{blaauwendraadStructuralShellAnalysis2014}, to structural health monitoring of aging industrial assets, to analysis of biological structures~\cite{houShellElementsforCellMechanics2018}.

\section{Methods}
\label{methods}

From a set of images, we reconstruct an accurate 3D model of an object using a spline-based boundary representation, directly compatible with traditional simulation pipelines such as IGA and FEA; in contrast to previous state-of-the-art reconstruction methods that produce models that require time-consuming post-processing to ensure compatibility.
Our representation guarantees properties necessary for simulation without compromising reconstruction accuracy; unlike mesh-based approaches, a B-spline boundary representation---together with normal and curvature regularization terms---maintains surface smoothness ($C^2$ within patches and near-$G^1$ across patch boundaries) and watertightness, enabling simulations to be carried out directly on the shape representation.
Furthermore, to the best of our knowledge, we are the first to demonstrate shape optimization of a multi-patch B-spline surface from images, as well as simulation on the reconstructed boundary representation model.

The method is comprised of three core stages.
First, we construct a multi-patch spline initial shape (\cref{subsec_multi_patch_spline}), ensuring watertightness across the surface and approximate $G^1$ continuity along the patch edges, while employing a synchronized mesh representation for efficient differentiable rendering.
We minimize the discrepancy between the rendered shape and reference images by repeated application of a two-step optimization procedure that alternates between shape and texture updates (\cref{sec:method:optimization}).
Finally, surface information inpainting (\cref{sec:method:inpainting}) augments the representation with inpainted fields (e.g., a material-class field and an initial thermal state) for use in downstream simulations.
Inpainted information is expressed using the same basis functions as the geometry, ensuring direct compatibility with IGA.
An overview of the full pipeline is provided in \cref{fig:flowchart}.

\subsection{Initialization of the Multi-Patch Spline Surface}\label{subsec_multi_patch_spline}

To ensure robust reconstruction and high-fidelity downstream simulation, \methodname{} requires an initial boundary representation composed of well-conditioned parametrized surfaces.
Applying the standard spherical parametrization to a single B-spline patch is unsuitable for this task; it creates a disproportionately high control point density near the poles and degenerate edges.
Such anisotropic distributions degrade the numerical performance and stability of both optimization and simulation pipelines.
Specifically, in the optimization phase, anisotropy reduces the effectiveness of gradient-based optimization methods, while in the simulation phase, the disproportionately small elements near the poles lead to poorly conditioned linear systems.
Equivalent issues are encountered in mesh-based optimization, where anisotropy prevents the graph Laplacian from providing effective regularization~\cite{botschLaplacians2008}, while small, stretched triangles near the poles deteriorate the conditioning of FEM systems.
Thus, an approximately uniform and isotropic distribution of degrees of freedom is required.

We employ a cubed-sphere representation composed of six patches, each corresponding to one face.
To ensure smoothness, control points are distributed uniformly within each patch, with control points along the edges and corners shared between corresponding patches.
The parametric mapping $S: [0, 1]^2 \to \mathbb{R}^3$ of each patch is given by a tensor-product B-spline basis, obtained from normalized open uniform knot vectors with $s$ knot spans and polynomial degree $p$.
While this formulation inherently provides $C^{p-1}$ continuity within patches, it only maintains $C^0$ continuity across boundaries; to bridge this gap, we introduce a regularization term that promotes normal continuity ($G^1$) across patch boundaries in \cref{subsubsec_normal_continuity}.
As polynomial B-splines cannot represent conics exactly, the initial shape is not a perfect sphere; it is obtained by projecting the cube control points onto a sphere followed by enforcing normal continuity across patches through optimization.
Detailed properties of the B-spline basis are provided in Supplementary, \cref{appendix:subsec_b_splines}.

Before shape optimization, the initial cubed-sphere should be visible in each reference view, and its rendered silhouette should mostly overlap with the target.
Estimating the required initial affine transform is framed as an optimization problem with 9 variables (3 translation, 3 rotation, and 3 scale variables), where the objective is to minimize the L1 discrepancy between the mask of the initial shape observed from the camera views and the reference masks obtained through image segmentation.
Before optimization, the position is initialized at the convergence point of the cameras' optical axes, while the initial scale is estimated using uniform sampling of the admissible range followed by a golden section search (GSS) on the most promising bracket.
Full details are provided in Supplementary, \cref{supp_mat:initial_pose}.

\subsection{Optimization}
\label{sec:method:optimization}

During optimization, the shape is expressed using two synchronized representations;
a composite B-spline boundary representation, as described in the previous section, and a tessellated mesh surface, used for computationally efficient inverse rendering.
Directly rendering the spline-based boundary representation is prohibitively expensive, as a non-linear solve is required for each ray, with known near-tangent intersection issues close to silhouettes~\cite{nishitaRayTracingSplines1990}.

Tessellated mesh surfaces are obtained by diagonal triangulation of a regular square grid in the parametric domain of each B-spline patch with $s_t$ subdivisions per parametric direction.
The surface mapping is applied to the parametric vertices, lifting them into $\mathbb{R}^3$ and yielding a mesh.
Vertex normals correspond to surface normals evaluated at parametric vertices, except at the edges, where they are computed as a normalized average of the two or three (in case of a corner) adjacent patches.
Since vertex positions and normals are derived directly from the global spline control points, the rendered surface stays watertight and is shaded smoothly.
Generating texture coordinates that produce a single seamless texture on general meshes is a non-trivial task, typically requiring multiple charts~\cite{levyConformalTextureMaps2002}.
Therefore, the obtained per-patch meshes are directly used for rendering, and each per-patch mesh is assigned a texture and utilizes the $[0, 1]^2$ space for both parametric and UV coordinates.

The optimization loop proceeds as follows.
First, our spline-based representation is converted into a mesh (as described above), yielding mesh vertices and vertex normals, as well as carrying an associated texture.
This collection of textured meshes is rendered, and gradients are computed using differentiable ray tracing
and a projective sampling integrator~\cite{zhangProjective2023}.
The resulting gradients guide shape optimization of the composite B-spline surface by back-propagating errors to the global control points and updating their positions.
Additional gradients stemming from regularization are also applied directly to the spline surface.
To maintain a faithful representation of the spline surface, mesh vertices and normals are updated after each iteration (\cref{sec:method:lbfgs}).
In the second phase, the control points are frozen, while the texture is updated using the same differentiable ray tracing pipeline (\cref{sec:method:texture-sigmoid-projection}).

\subsubsection{Coarse-to-Fine Scheme}\label{subsec_coarse_to_fine_scheme}

In mesh-based shape optimization, large gradient steps often lead to self-intersections and instability, requiring gradient preconditioning schemes~\cite{nicoletLargeStepsInverse2021a}.
Spline-based primitives inherently mitigate these issues: $C^2$ intra-patch continuity is maintained for cubic patches ($p=3$), with each point evaluation influenced by multiple control points, diffusing gradients over a much broader area than for mesh-based representation.

The number of control points and the polynomial degree are important parameters that influence the quality and level of detail of the reconstructed shape.
While an insufficient number of points constrains surface expressiveness and hinders the recovery of fine geometric detail, higher polynomial degrees diffuse gradients more strongly, potentially leading to over-smoothing.
To address this trade-off, we employ an $h$-refinement strategy that incrementally increases the number of knot spans---and thus the surface resolution---throughout the optimization process, while the degree is held fixed.
Specifically, we initialize the process with a cubic ($p=3$) B-spline basis with two knot spans per parametric direction.

$h$-refinement is performed at predetermined iteration numbers to increase the number of control points and attainable detail once a given refinement level stalls.
Since the minimum refinement factor that preserves uniformity of the open knot vector is two, $h$-refinement is implemented by inserting a new knot at the midpoint of each knot span in both the $\xi$ and $\eta$ directions---for details on the knot insertion algorithm, see Supplementary, \cref{sup_mat:subsubsec_knot_insertion}.
With each refinement step, the support of individual basis functions becomes more localized, while the total number of basis functions increases, enabling each patch to capture finer detail.
Since the magnitude of silhouette gradients scales with the discrepancy between the current estimate and the target geometry, refinement at later stages allows the introduction of detail without compromising optimization stability in the early stage.

\subsubsection{L-BFGS Optimizer for Shape Optimization}
\label{sec:method:lbfgs}

While using the Adam optimizer with uniform moment estimates combined with gradient preconditioning effectively stabilizes shape optimization on meshes by diffusing sharp silhouette gradients~\cite{nicoletLargeStepsInverse2021a,worchelDifferentiableRenderingParametric2023}, this approach introduces excessive smoothing to ensure stability when optimizing dense meshes.
Recovering finer geometric details requires reducing the smoothing strength; however, doing so introduces instabilities, as Adam's momentum amplifies strong spurious gradients.

We avoid the preconditioning altogether through the aforementioned coarse-to-fine scheme, and we overcome the instability by using the limited memory Broyden–Fletcher–Goldfarb–Shanno (L-BFGS) method~\cite{nocedalLBFGS1989}.
While a point on a triangular mesh is influenced by only three vertices, a point on a cubic B-spline surface depends on up to 16 control points.
This increased coupling inherently improves optimization stability and motivates the use of a quasi-Newton optimizer that captures these couplings through its curvature estimates.
L-BFGS conditions the descent direction by constructing an approximation of the Hessian based on $m$ previous gradients and optimization states.
We use the standard inexact line search algorithm to ensure monotonic decrease in the objective function; to maintain a strictly positive-definite Hessian approximation, updates are discarded if the estimated curvature falls below a small positive threshold.

\subsubsection{Regularization}\label{subsec_regularization}

We employ two complementary regularization terms to maintain visual smoothness and guard against self-intersections, as well as to improve numerical stability of both the optimization procedure and downstream simulation tasks.
The first enforces normal continuity across patch interfaces, while the second penalizes excessive curvature within individual spline patches.

\paragraph{Boundary Normal Continuity}\label{subsubsec_normal_continuity}
Each B-spline patch---defined by two open uniform knot vectors of degree $p$---exhibits $C^{p-1}$ continuity within the patch.
At patch boundaries, all patches share the same degree $p$ and identical normalized open uniform knot vectors with matched knot spans, ensuring watertightness, $C^0$ continuity across edges, and maximal smoothness within each patch.

However, $C^0$ continuity is insufficient for visually smooth transitions across patches; discontinuities in surface normals along shared edges can lead to visible creases.
To address this, we enforce alignment of surface normals along all shared patch edges.

Let $E$ denote the set of all edges shared between adjacent patches.
Each edge is represented by a one-dimensional parametric curve $\boldsymbol{e}(t) \coloneq(\xi(t), \eta(t))$, defined as:
\begin{equation}
    \boldsymbol{e}(t) =
    \begin{cases}
        (0, t), \quad t \in [0, 1], & \text{if patch's left side}   \\
        (1, t), \quad t \in [0, 1], & \text{if patch's right side}  \\
        (t, 0), \quad t \in [0, 1], & \text{if patch's bottom side} \\
        (t, 1), \quad t \in [0, 1], & \text{if patch's top side}    \\
    \end{cases}
\end{equation}
Given a shared edge between two patches, there is no guarantee that edge orientations of the two patches are compatible: a point evaluated at $t=0$ along one patch might correspond to a point at $t=1$ on its neighbor.
In this case, the orientation of the edge of the neighboring patch must be flipped through re-parametrization.
To detect orientation mismatches between two neighboring patches $S_1$ and $S_2$, we compute tangent vectors along the shared edge:
\begin{equation}
    \begin{aligned}
        \boldsymbol{d}_1 = S_1(\boldsymbol{e}_1(1)) - S_1(\boldsymbol{e}_1(0)), \\
        \boldsymbol{d}_2 = S_2(\boldsymbol{e}_2(1)) - S_2(\boldsymbol{e}_2(0)),
    \end{aligned}
\end{equation}
with ${e}_1$ and ${e}_2$ being the edge parametric curves representing the shared edge on $S_1$ and $S_2$, respectively.
Since the control points are shared, the two vectors $\boldsymbol{d}_1$ and $\boldsymbol{d}_2$ are collinear: the sign of the dot product indicates whether the parameterizations point along the same direction (if positive) or in opposite directions (if negative).
We re-parametrize the edge along the second patch such that
\begin{equation}
    f(t) =
    \begin{cases}
        t,     & \boldsymbol{d}_1 \cdot \boldsymbol{d}_2 > 0, \\
        1 - t, & \boldsymbol{d}_1 \cdot \boldsymbol{d}_2 < 0,
    \end{cases}
\end{equation}
Finally, the normal continuity loss is expressed as:
\begin{equation}
    \mathcal{L}_{\text{N}} = \frac{1}{|E|} \sum_{e \in E} \int_{0}^1 \norm{\boldsymbol{n}_1(\boldsymbol{e}_1(t)) - \boldsymbol{n}_2(\boldsymbol{e}_2(f(t))) }^2 \, dt.
\end{equation}
The loss is normalized globally by the number of edges $|E|$ and locally by integration over a unit interval.

Since the integrand is too complex to be evaluated analytically---due to the complexity of B-spline basis functions---we use numerical quadrature (see Supplementary, \cref{appendix:numerical_quadrature}).
We empirically found that using the composite midpoint rule with the number of integration samples matching the number of triangles along the edge of the tessellated patch yields sufficient accuracy.

\paragraph{Spline Curvature Regularization}\label{subsubsec_curvature_regularization}

Shape optimization is susceptible to geometric artifacts, such as folds and cusps, which result in inaccurate reconstructions and adversely affect downstream applications---e.g., conditioning of linear systems in IGA simulation.
To mitigate these distortions, we introduce a regularization term that penalizes excessive curvature within each spline patch---guarding against the formation of cusps and improving the conditioning of the numerical systems used in subsequent simulation tasks.
This term is formulated to minimize the Willmore energy~\cite{willmoreRiemannianGeometry1993}---targeting the difference between the two principal curvatures---to promote geometric smoothness:
\begin{equation}
    \mathcal{L}_{\text{W}} = \frac{1}{4} \int_S (k_1 - k_2)^2 dA,
\end{equation}
where $k_1$ and $k_2$ denote the principal curvatures at a given point on the surface.
The sphere is practically the smoothest shape we can expect to obtain, and the Willmore energy vanishes identically.
Further details are provided in Supplementary, \cref{subsubsec_diff_geometry}.

Since analytic integration of this expression for a cubic B-spline patch is infeasible, numerical quadrature is used to evaluate the double integral.
The integral is decomposed into a sum of integrals over individual knot spans, each computed using the tensor-product Gauss--Legendre quadrature~\cite{stoer_1980_intro} along both parametric directions $\xi$ and $\eta$---see Supplementary, \cref{subsec_tensor_quad} for tensor-product Gauss--Legendre quadrature and \cref{sup_amt:willmore} for its application to the Willmore integral.

\subsubsection{Shape Optimization Objective}\label{subsec_objective_function}

The inherent smoothing properties of the spline basis stabilize the simulation while also dampening spatial variation, reducing detail reconstruction.
Using only L1 and L2 loss---the common choice for shape optimization \cite{nicoletLargeStepsInverse2021a, laineNvdiffrast2020}---results in reduced detail in the reconstructed shape.
We use the L1 loss together with the Multi-Scale SSIM (MS-SSIM)~\cite{wangMultiscaleSSIM2003}, a generalization of SSIM loss that incorporates image details at different resolution scales.
Using a combination of the L1 and MS-SSIM loss leads to a significant improvement in reconstruction quality---compared to the L1 or L2 loss, as well as L1 loss with SSIM---preserving fine geometric detail without introducing artifacts.

Finally, given a set of $N$ reference images with resolution $H \times W$ $\{\mathbf{r}_i\}_{i=1}^N, \mathbf{r}_i \in \mathbb{R}^{H \times W \times 3}$, the complete loss is defined as
\begin{equation}
    \label{eq:loss}
    \begin{split}
        \mathcal{L}(\mathbf{x}, \mathbf{t}) ={} & \beta \mathcal{L}_{\text{N}}(\mathbf{x}) + \gamma \mathcal{L}_{\text{W}}(\mathbf{x})
        + \frac{1}{N} \sum_{i=1}^{N} \left[ \alpha \mathcal{L}_{\text{L1}}(R_i(\mathbf{x}, \mathbf{t}), \mathbf{r}_i) \right.                          \\
                                                & + \left. (1-\alpha) \mathcal{L}_{\text{MS-SSIM}}(R_i(\mathbf{x}, \mathbf{t}), \mathbf{r}_i) \right],
    \end{split}
\end{equation}
where $\alpha \in [0, 1]$ controls the balance between the L1 and MS-SSIM terms, while $\beta$ and $\gamma$ weight the normal continuity and curvature regularization terms.
$\mathbf{x} \in \mathbb{R}^{n_\text{c} \times3}$ denotes all $n_\text{c}$ global control points while $\mathbf{t} \in \mathbb{R}^{6\times r \times r \times 3}$ denotes the tensor representing the 6 patch textures of resolution $r \times r$.
$R_i: \mathbb{R}^{n_\text{c} \times3} \times \mathbb{R}^{6\times r \times r \times 3} \to \mathbb{R}^{H \times W \times 3}$ denotes the differentiable rendering function for the camera pose associated with image $i$.

\subsubsection{Texture Optimization}
\label{sec:method:texture-sigmoid-projection}

Shape and texture are updated iteratively through an alternating scheme.
Within each iteration, all views are processed twice: first for shape optimization using global L-BFGS, and subsequently for texture optimization using batched Adam~\cite{kingmaAdam2015}.
This alternation accommodates the distinct optimizers and batch sizes required for each process, ensuring independent updates.
We assume that the surface material is Lambertian---i.e., perfectly diffuse---with surface-varying albedo parametrized by a texture assigned to each patch.
The albedo is defined at each patch point as a triplet of values from the interval $[0, 1]$.

We introduce a sigmoid projection to ensure that albedo values remain within the valid range $[0, 1]$.
Let $\mathbf{z} \in \mathbb{R}^{6 \times r \times r \times 3}$ denote the unconstrained latent variable, with $r$ the texture resolution.
$\mathbf{z}$ is mapped to the constrained albedo texture $\mathbf{t} \in \mathbb{R}^{6 \times r \times r \times 3}$ via
\begin{equation}
    \textbf{t}(\textbf{z}) = \epsilon + \frac{1 - 2\epsilon}{1+\exp(-\sigma \, \mathbf{z})},
\end{equation}
where $\sigma$ controls the steepness of the projection (we use $\sigma=1$) and $\epsilon$ is a small constant (in practice we used $\epsilon=10^{-4}$) introduced to prevent numerical issues that may otherwise lead to values slightly outside the $[0, 1]$ interval and to stabilize the inverse mapping
\begin{equation}
    \mathbf{z}(\mathbf{t}) = \frac{1}{\sigma} \ln \frac{\mathbf{t} - \epsilon}{1-\epsilon - \mathbf{t}}.
\end{equation}
The sigmoid operation is applied element-wise to all entries of the tensor.
To avoid numerical underflow or overflow when evaluating the exponential, the input to the sigmoid is clamped to the interval $[\ln \epsilon / \sigma, -\ln \epsilon / \sigma]$ prior to projection.
Similarly, for the inverse transformation, values are clamped to $[2\epsilon, 1 - 2\epsilon]$.

Texture optimization is performed over the latent variable $\mathbf{z}$ using the Adam optimizer while the control points defining the shape are held fixed.
The texture is then updated and assigned to the per-patch meshes, allowing the shape optimization step to proceed with fixed textures.
The loss for texture optimization is composed of only the L1 and MS-SSIM terms of \cref{eq:loss} balanced by the same $\alpha$.

\subsection{Inpainting Physical Information onto Optimized Shapes}
\label{sec:method:inpainting}

Physics-based models often depend on spatially varying surface quantities that are not estimated during shape optimization.
Instead, those must be inferred and mapped onto the reconstructed surface afterwards; a process we will refer to as inpainting.
While previous work assigns semantic~\cite{valentinMeshSemantic2013,kunduVirtualSemantic2020} or thermal~\cite{yangInfraredFusion2018} values directly to the discretized mesh surface, directly performing such a mapping on splines is challenging due to the non-linear nature of spline basis.
To ensure stability of the mapping procedure and accurate representation of the information by the basis, we frame the mapping procedure as an optimization problem.
It recovers the spatially varying field, from which the physical parameter values (e.g., temperature, conductivity, or density) are subsequently derived for later simulation.
To increase the resolution of the spline representation, we apply $h$-refinement in both parametric directions to the optimized shape prior to the inpainting optimization, raising the per-patch resolution to $256$ knot spans per direction.

Let $\{\mathbf{r}_i^\text{INP}\}_{i=1}^M$ be a set of $M$ reference normalized grayscale posed-images with values in the $[0, 1]$ interval, representing the field to be inpainted onto the spline basis, accompanied by corresponding camera pose information.
The inpainting loss is defined as
\begin{equation}
    \mathcal{L}_\text{INP}(\mathbf{f}) = \frac{1}{M} \sum_{i=1}^{M} \mathcal{L}_\text{L1} \left( R_i^\text{INP}(\mathbf{f}), \mathbf{r}_i^\text{INP} \right),
\end{equation}
where $R_i^\text{INP}: \mathbb{R}^{n_\text{f}} \to \mathbb{R}^{H \times W}$ is the differentiable inpainting rendering function (with $n_\text{f}$ the number of basis functions in the global spline basis) and $\mathbf{f} \in \mathbb{R}^{n_\text{f}}$ are the basis function coefficients representing the inpainted field.
To ensure the inpainted field is constrained to the $[0, 1]$ interval, we employ the sigmoid parametrization introduced in \cref{sec:method:texture-sigmoid-projection}.

Unlike the albedo-based rendering used to optimize textures, the patches are rendered as area light sources, such that the loss minimizes the difference between the rendered brightness and the target field information.
Since we assume no scattering phenomena on the inpainted field, the area light source is assigned a perfectly black material.
To enable efficient differentiable rendering, a texture is generated by evaluating the spline field at parametric coordinates corresponding to pixel midpoints and is then attached to each patch's mesh.
This approach is similar to the one used for shape optimization, allowing gradients to flow through the texture to the appropriate basis function coefficients.

\section{Implementation}

\subsection{Datasets}\label{implementation:datasets}
\subsubsection{Synthetic Textureless Meshes}
\label{sec:implementation:texturelessmeshes}

Suzanne, the Stanford Bunny, Spot, and Armadillo meshes are obtained from the ``common-3d-test-models'' repository.\footnote{\url{https://github.com/alecjacobson/common-3d-test-models}}
The meshes are rendered with the environmental map \textit{envmap2} from Mitsuba~\cite{Mitsuba3} and from camera poses obtained using Fibonacci camera sampling around the center of each object.
We use a Lambertian BSDF with albedo $0.5$ (maximum value for RGB is 1.0) for rendering.
We sample $N = 12$ views for \textit{suzanne} and \textit{bunny}, $N = 24$ views for \textit{spot}, and $N = 32$ views for \textit{armadillo}.
Reference images are rendered with a resolution of $512 \times 512$ pixels and $4096$ samples per pixel, while $64$ samples per pixel are used during optimization.

To ensure watertightness, detached interior bodies---such as Suzanne's eyeballs---are removed and sealed, while plain openings---such as the three holes on the bottom of the Stanford bunny---are also sealed.
Given that the GT meshes are not of sufficiently high quality for direct FEM simulation, they are remeshed using \texttt{gmsh}~\cite{geuzaineGmsh2009} with a target edge length of $0.012$ \si{\meter} for the heat validation and $0.025$ \si{\meter} for the modal validation.

\subsubsection{BuildNet3D}
\label{par:synthetic_buildings_desc}

We select the four realistic buildings from the BuildNet3D~\cite{xuExploitingSemanticScene2025b} dataset and randomly split the rendered and semantic mask images into training and test sets containing 180 and 20 images, respectively.
Building masks are used to isolate the building in the image, while a light checkerboard (dark square albedo $0.5$ instead of $0.0$) environmental map is used as a background for both training and test viewpoints.
Such a background was chosen to maximize silhouette contrast and provide uniform illumination while reducing the rendering variance of a pure black/white checkerboard due to emitter sampling.
A purely white background is insufficient to prevent the growth of white blobs that are detrimental to reconstruction quality.
All images have a resolution of $512 \times 512$ pixels.

\subsubsection{Real Scenes}
\label{sec:implementation:real_scenes_desc}

The four real-world scenes include both RGB images and associated thermal maps.
The background of each image is removed by segmenting the object using the Segment Anything 3 (SAM3) model~\cite{carionSAM3Segment2025} and a prompt describing the object.
To create the reference images for each scene, we use as environment map the same light checkerboard as for BuildNet3D scenes.
Images have resolution of $480 \times 640$.
Camera poses are then obtained using the Visual Geometry Grounded Transformer (VGGT)~\cite{wangVGGT2025}.
The dense captures are used for the benefit of pose estimation and the NVS baselines; the end-to-end validation uses only $12$--$32$ views.

Some parts of the target object are never observed in the training data, hence, during thermal inpainting, unobserved parts of the reconstructed model are assigned a low ($0.1$) temperature for simulation.
Substituting it with an average temperature would be preferred for practical applications, but we keep the temperature low to specifically highlight these areas.

\subsection{Thermal and Semantic Inpainting}

Thermal information represents a continuous, smoothly varying field captured via infrared imaging.
This information is converted into a normalized grayscale image, where pixel intensities are mapped to a fixed temperature scale via a known affine transform.
Notably, the rendering process does not explicitly simulate physical infrared radiation transport; instead, the field is optimized to match the captured intensity directly.

Semantic information is extracted from RGB images using an off-the-shelf text-promptable segmentation model~\cite{carionSAM3Segment2025}.
We define semantic information---such as the identification of structural elements like windows---as a specification for material property assignment in the downstream simulations.
Thus, in material field inpainting, each pixel is assigned a categorical material label based on the segmented images (see \cref{sec:implementation:real_scenes_desc} for segmentation details).
As for the thermal case, the inpainting procedure treats the labeled images as ground truth and models the material field as a continuous scalar in $[0, 1]$.
Post-optimization, discrete material assignments are obtained by thresholding the continuous surface values to their nearest categorical labels.

\subsection{Experiment Setup}\label{implementation:setup}

Our method is implemented in Python, using the Mitsuba 3.8.0~\cite{Mitsuba3} framework, while the experiments are run on an NVIDIA DGX Spark workstation.
Shape and texture optimization requires between 4 and 6 hours per scene on the Spark's GB10 for the BuildNet3D and real-world datasets.

\subsection{Method parameters}\label{implementation:params}

For all scenes, we set $\alpha = 0.85$, $\beta = 10^{-2}$, and $\gamma = 2 \cdot 10^{-5}$.
These parameters were determined empirically, on the Suzanne scene, adjusting one parameter at a time;
the results obtained in the experiments were robust to variations of $\alpha$ around 0.85, while $\beta$ and $\gamma$ were minimized without compromising reconstruction quality.
We observe that $\mathcal{L}_{\text{N}}$ vanishes towards the end of the optimization while $\mathcal{L}_{\text{W}}$ becomes large, corresponding to the coarse-to-fine scheme recovering more fine detail.
The L-BFGS optimizer with full batches (all images) and history size $m=15$ is utilized.

In case of the textureless meshes, we run the optimization for $250$ iterations for \textit{bunny, suzanne, spot}, and $400$ iterations for \textit{armadillo}.
We increase the number of knot spans at iterations \{25, 50, 80, 140\} with an additional refinement at iteration $250$ for \textit{armadillo}.
For BuildNet3D and real-world scenes, we run the optimization for $250$ iterations and increase the number of knot spans at iterations \{20, 50, 80, 120, 180\}---apart for Building A where we use 300 iterations and increase at \{30, 60, 90, 150, 210\}, and for Lion with 400 iterations and increases at \{50, 100, 150, 200, 250\}.
Each patch is assigned a $512\times512$ texture and is tessellated with a uniform diagonal triangulation in the parametric domain with $256\times256 \times 2$ faces per-patch.

\subsection{Heat Simulation}\label{implementation:heat}

To model heat flow on a manifold surface, we use the Isogeometric Analysis (IGA) framework~\cite{hughesIsogeometricAnalysisCAD2005a}.
Starting from the 3D heat equation with the assumption that the geometry is given by a thin volume that can be approximated with a closed manifold $\mathcal{M} \subset \mathbb{R}^3$, we obtain the following weak formulation:
\begin{equation}
    \begin{aligned}
         & \int_\mathcal{M} d(\boldsymbol{y}) \, C(\boldsymbol{y}) \, \partial_t u(\boldsymbol{y}, t) \,\varphi(\boldsymbol{y}) \, dS +\int_\mathcal{M} d(\boldsymbol{y}) \, \nabla \varphi(\boldsymbol{y}) \cdot \left (\mathbf{K}(\boldsymbol{y}) \, \nabla u(\boldsymbol{y}, t) \right) \, dS \\
         & = \int_\mathcal{M} \, f_M(\boldsymbol{y}, t) \, \varphi(\boldsymbol{y}) \, dS + \int_\mathcal{M} F_N(u, \boldsymbol{y}) \, \varphi(\boldsymbol{y}) \, dS,
    \end{aligned}
\end{equation}
where $u(\boldsymbol{y}, t)$ is the temperature at each $\boldsymbol{y} \in \mathcal{M}$ at time $t \geq0$ [\si{\kelvin}].
Material properties are represented through the spatially varying volumetric heat capacity $C: \mathcal{M} \to \mathbb{R}$ [\si{\joule\per\cubic\meter\per\kelvin}], the thickness $d: \mathcal{M} \to \mathbb{R}$ [\si{\meter}], and the thermal conductivity tensor $\mathbf{K}: \mathcal{M} \to \mathbb{R}^{3 \times 3}$ [\si{\watt\per\meter\per\kelvin}].
Finally, $f_M: \mathcal{M} \times \mathbb{R}^+ \to \mathbb{R}$ is the spatially varying time-dependent surface source term [\si{\watt\per\square\meter}], while $F_N: \mathbb{R} \times \mathcal{M} \to \mathbb{R}$ is the non-linear, spatially varying Neumann flux.

$\varphi(\boldsymbol{y})$ is a test function from the space $V \subset H^1{(\mathcal{M})}$.
Note that watertightness ($C^0$ continuity) of the geometry is sufficient here, as the weak formulation requires only $H^1$ regularity of the solution.
Using the IGA formalism~\cite{dedeIsogeometricAnalysisSecond2015}, the discrete solution space $V_h \subset V$ is constructed from the spline basis functions that define the manifold $\mathcal{M}$.
The semi-discrete formulation is discretized in time using the Crank--Nicolson scheme \cite{crankNicolsonScheme1947} due to its conservation of total heat and unconditional stability.
The full derivation, from the 3D heat equation to the Crank--Nicolson scheme is available in Supplementary, \cref{appendix:heat_equation}.
In our experiments, we chose $F_N = 0$ and $f_M(\boldsymbol{y}) = f_0 \, w(\boldsymbol{y})$ where $w: \mathcal{M} \to [0, 1]$ represents the semantic label (either wall or window).
While this mapping should ideally be discrete, in practice $w$ is represented by smooth spline basis functions, as obtained during the inpainting step (described in \cref{sec:method:inpainting}).
Temperature fields are treated on an affinely normalized scale: the end-to-end validation prescribes temperatures directly in $[0, 1]$~\si{\kelvin}, while real-scene thermal images are normalized during preprocessing; reported thermal energies refer to this scale.

In the absence of the Neumann term, the total heat at time $t$ is strictly determined by the initial condition $u_0(\boldsymbol{y})$ and the total heat generation:
\begin{equation}
    \int_\mathcal{M} d(\boldsymbol{y}) \, C(\boldsymbol{y}) \, u(\boldsymbol{y}, t) \, dS =
    \int_\mathcal{M} d(\boldsymbol{y}) \, C(\boldsymbol{y}) \, u_0(\boldsymbol{y}) \, dS + t\int_\mathcal{M} f_M(\boldsymbol{y}) \, dS.
\end{equation}
As the Crank--Nicolson scheme is used for time evolution, total heat is conserved up to numerical error.
Note that the simulation is not limited to two material types---more could be introduced---and that the source term can vary across the surface in a more complex manner; the only limitation in accuracy is related to the underlying B-spline basis resolution and the chosen time step size.

\begin{table}[t]
    \centering
    \caption{
        Thermal material properties used in the heat simulation.
        The bulk body material is reported per scene; the glass rows apply to the window-mask regions of the corresponding scenes, while the Lion uses uniform material properties.
        Material thickness $d$ refers to the shell-element thickness used in the manifold-PDE reduction.
        Scale $L$ is the target scale for the longest axis of the scene bounding box to resolve scale ambiguity
        inherent to pose estimation and is applied to the 4 real datasets.
        ``All Buildings'' includes BuildNet3D, Building A, and Woodshed.
        The Synthetic meshes rows give the aluminum/steel pair used for the end-to-end ground-truth validation (\cref{tab:e2e_summary}) on the Suzanne, Spot, Bunny, and Armadillo meshes; these are run at native ground-truth scale, so no $L$ is applied.
    }
    \label{tab:scene_thermal_materials}
    \begin{tabular}{llccccc}
        \toprule
        Scene(s)      & Material  & $L$ [\si{\meter}] & $k$ [\si{\watt\per\meter\per\kelvin}] & $\rho$ [\si{\kilogram\per\meter\cubed}] & $C_p$ [\si{\joule\per\kilogram\per\kelvin}] & $d$ [\si{\meter}] \\
        \midrule
        BuildNet3D    & concrete  & ---               & 2.00                                  & 2250                                    & 900                                         & 0.200             \\
        Building A    & concrete  & 15.0              & 2.00                                  & 2250                                    & 900                                         & 0.200             \\
        Woodshed      & wood      & 2.5               & 0.15                                  & 600                                     & 2400                                        & 0.020             \\
        Lion          & wool/felt & 0.3               & 0.04                                  & 100                                     & 1400                                        & 0.020             \\
        Car Body      & steel     & 4.0               & 50.0                                  & 7850                                    & 500                                         & 0.002             \\
        Car Windows   & glass     & ---               & 0.90                                  & 2500                                    & 840                                         & 0.001             \\
        All Buildings & glass     & ---               & 0.90                                  & 2500                                    & 840                                         & 0.010             \\
        \midrule
        \multirow{2}{*}{\makecell[l]{Synthetic                                                                                                                                                            \\meshes}}
                      & steel     & ---               & 50.0                                  & 7850                                    & 500                                         & 0.010             \\
                      & aluminum  & ---               & 237.0                                 & 2700                                    & 900                                         & 0.010             \\
        \bottomrule
    \end{tabular}
\end{table}

The physical time step is derived from the system's natural relaxation time, defined as $\Delta t = \tau \cdot T_{\text{relax}}$, where $T_{\text{relax}} = 1/\lambda_1$ and $\lambda_1$ denote the smallest non-zero generalized eigenvalue of the assembled stiffness--mass pair $(\mathbf{A}, \mathbf{M})$.
To accurately capture sharp transients present in the initial data, we set $\tau = 10^{-5}$.
The end-to-end validation runs (\cref{tab:e2e_summary}) use $1000$ steps, while the building and real-world scenes are run for $500$ steps; \cref{fig:sim_synthetic,fig:sim_real} show the temperature field at the initial time, the midpoint (step $250$), and the final time (step $500$).
Material properties for the test cases in \cref{results:heatSimulation} are detailed in \cref{tab:scene_thermal_materials}.

We validate our IGA implementation against a FEM-based baseline implemented in DOLFINx~\cite{barattaDOLFINx2023} (version 0.9.0)---see Supplementary, \cref{suppmat:solvervalidation}.
The same FEM baseline is used in our end-to-end experiments on textureless meshes.

\subsection{Modal Analysis}\label{implementation:modal}

To compute the dynamics of the models obtained through \methodname{}, we model the reconstructed geometry as a thin surface, allowing the 3D equations of motion to be reasonably reduced to a shell elastodynamics problem.
Shell-based elastodynamics is highly sensitive to geometric quality and to the choice of functional space used to represent the discrete solution~\cite{guarinoIBCMShells2024}.
Unlike the heat equation, where IGA and FEA-based simulation methods produce near-identical results, shell theory requires higher geometric quality regarding both continuity and curvature.

As noted in \cref{results:modalAnalysis}, FEM implementations of Reissner-Mindlin (RM) shell theory can face various locking phenomena that cannot be resolved without specialized finite element types~\cite{batheMITCElements1986}.
On the other hand, IGA largely overcomes these issues due to the inherent smoothness of the geometry and the basis functions.
Our independent finite-element reference therefore employs an assumed-strain (B-bar) projection~\cite{huReissnerMindlinBbar2020} to obtain a locking-free spectrum, whereas the IGA discretization is fully integrated and requires no such correction; both implementations are detailed in Supplementary, \cref{appendix:reissner_mindlin}.
We use a formulation with 6 degrees of freedom per control point---comprising 3 spatial and 3 rotational displacements in global Cartesian coordinates---which necessitates the use of drilling stabilization to ensure only physical rotations are allowed and the RM kinematic assumption is satisfied~\cite{guarinoRMStabilization2025}.

To conduct modal analysis, we assemble two bilinear forms for the stiffness and the mass and solve the generalized eigenvalue problem
\begin{equation}\label{eq:rm_weak_form}
    a(\delta\boldsymbol{u}, \boldsymbol{u}) = \lambda \, m(\delta\boldsymbol{u}, \boldsymbol{u}), \qquad \forall \, \delta\boldsymbol{u} \in V,
\end{equation}
where $V = [H^1(\mathcal{M})]^6$, $a: V \times V \to \mathbb{R}$ is the stiffness bilinear form, and $m: V \times V \to \mathbb{R}$ is the mass bilinear form.
The trial field $\boldsymbol{u} = (\boldsymbol{v}, \boldsymbol{\omega})$ packs three translational displacements $\boldsymbol{v}$~[\si{\meter}] and three rotational displacements $\boldsymbol{\omega}$~[\si{\radian}] per point, both expressed in the global Cartesian frame, and $\delta\boldsymbol{u}$ is the analogous test field.
The eigenvalue $\lambda$~[\si{\per\second\squared}] is the squared natural angular frequency, so that the physical frequency in Hz is $f = \sqrt{\lambda} / (2\pi)$.

The stiffness bilinear form can be decomposed into membrane, bending, shear, and drilling contributions:
\begin{equation}\label{eq:rm_stiffness_decomposition}
    a(\delta\boldsymbol{u}, \boldsymbol{u}) = a_{\text{mem}}(\delta\boldsymbol{u}, \boldsymbol{u}) + a_{\text{bend}}(\delta\boldsymbol{u}, \boldsymbol{u}) + a_{\text{shear}}(\delta\boldsymbol{u}, \boldsymbol{u}) + a_{\text{drill}}(\delta\boldsymbol{u}, \boldsymbol{u}),
\end{equation}
which are given by the following expressions evaluated at each point $\boldsymbol{y} \in \mathcal{M}$:
\begin{subequations}\label{eq:rm_bilinear_forms}
    \begin{align}
        a_{\text{mem}}(\delta\boldsymbol{u}, \boldsymbol{u})   & = \int_\mathcal{M} d(\boldsymbol{y}) \, \mathcal{D}^{\alpha\beta\gamma\delta}(\boldsymbol{y}) \, \delta\varepsilon_{\alpha\beta}(\boldsymbol{y}) \, \varepsilon_{\gamma\delta}(\boldsymbol{y}) \, dS, \label{eq:rm_a_mem}                                                                                                                                          \\
        a_{\text{bend}}(\delta\boldsymbol{u}, \boldsymbol{u})  & = \int_\mathcal{M} \frac{d(\boldsymbol{y})^3}{12} \, \mathcal{D}^{\alpha\beta\gamma\delta}(\boldsymbol{y}) \, \delta\kappa_{\alpha\beta}(\boldsymbol{y}) \, \kappa_{\gamma\delta}(\boldsymbol{y}) \, dS, \label{eq:rm_a_bend}                                                                                                                                      \\
        a_{\text{shear}}(\delta\boldsymbol{u}, \boldsymbol{u}) & = \int_\mathcal{M} \kappa_s \, \mu(\boldsymbol{y}) \, d(\boldsymbol{y}) \, g^{\alpha\beta}(\boldsymbol{y}) \, \delta\gamma_\alpha(\boldsymbol{y}) \, \gamma_\beta(\boldsymbol{y}) \, dS, \label{eq:rm_a_shear}                                                                                                                                                     \\
        a_{\text{drill}}(\delta\boldsymbol{u}, \boldsymbol{u}) & = \int_\mathcal{M} \alpha_{\text{drill}}(\boldsymbol{y}) \left(\delta\boldsymbol{\omega}(\boldsymbol{y}) \cdot \boldsymbol{\hat n}(\boldsymbol{y}) - \delta\omega_{\text{phys}}(\boldsymbol{y})\right) \left(\boldsymbol{\omega}(\boldsymbol{y}) \cdot \boldsymbol{\hat n}(\boldsymbol{y}) - \omega_{\text{phys}}(\boldsymbol{y})\right) dS. \label{eq:rm_a_drill}
    \end{align}
\end{subequations}
Here $d(\boldsymbol{y})$~[\si{\meter}] is the shell thickness, $\mathcal{D}^{\alpha\beta\gamma\delta}(\boldsymbol{y})$~[\si{\pascal}] is the in-plane elastic stiffness tensor (constructed from Young's modulus $E(\boldsymbol{y})$~[\si{\pascal}] and Poisson's ratio $\nu(\boldsymbol{y})$), and $\mu(\boldsymbol{y}) = E(\boldsymbol{y}) / [2(1+\nu(\boldsymbol{y}))]$~[\si{\pascal}] is the shear modulus.
The membrane strain $\varepsilon_{\alpha\beta}(\boldsymbol{y})$ and transverse shear strain $\gamma_\alpha(\boldsymbol{y})$ are dimensionless, the bending curvature change $\kappa_{\alpha\beta}(\boldsymbol{y})$ has units~[\si{\per\meter}], and the shear correction factor $\kappa_s = 5/6$ is dimensionless.
The inverse surface metric $g^{\alpha\beta}(\boldsymbol{y})$ and the unit normal $\boldsymbol{\hat n}(\boldsymbol{y})$ are dimensionless; $\alpha_{\text{drill}}(\boldsymbol{y})$~[\si{\pascal\meter}] is the drilling-stabilization penalty, and $\omega_{\text{phys}}(\boldsymbol{y})$~[\si{\radian}] denotes the physically admissible drilling rotation.

Analogously, the mass bilinear form is given as:
\begin{equation}\label{eq:rm_mass_bilinear}
    \begin{aligned}
        m(\delta\boldsymbol{u}, \boldsymbol{u}) & = \int_\mathcal{M} \rho(\boldsymbol{y}) \, d(\boldsymbol{y}) \, \delta\boldsymbol{v}(\boldsymbol{y}) \cdot \boldsymbol{v}(\boldsymbol{y}) \, dS                                                                                                                                                                                                       \\
                                                & + \int_\mathcal{M} \frac{\rho(\boldsymbol{y}) \, d(\boldsymbol{y})^3}{12} \left(\delta\boldsymbol{\omega}(\boldsymbol{y}) \cdot \boldsymbol{\omega}(\boldsymbol{y}) - (\delta\boldsymbol{\omega}(\boldsymbol{y}) \cdot \boldsymbol{\hat n}(\boldsymbol{y}))(\boldsymbol{\omega}(\boldsymbol{y}) \cdot \boldsymbol{\hat n}(\boldsymbol{y}))\right) dS.
    \end{aligned}
\end{equation}
where $\rho(\boldsymbol{y})$~[\si{\kilogram\per\meter\cubed}] is the mass density, with the first term contributing translational inertia and the second the rotational inertia of the shell director.
For the full derivation, alongside a detailed specification of all the terms, please refer to Supplementary, \cref{appendix:reissner_mindlin}.
For simplicity, it was assumed that the material properties are constant throughout the shell thickness---however, it should be noted that this is not a limitation of our method.
All modal analyses use $128$ knot spans per parametric direction with uniform material properties per scene, while the heat simulation operates directly on the $256$-span inpainted representation (see Supplementary, \cref{appendix:refsolverandtimeintegration}).

\begin{table}[t]
    \centering
    \caption{Structural material properties used in the Reissner-Mindlin modal analysis. All scenes use a single isotropic material with the standard shear correction factor $\kappa_s = 5/6$.
        The shell thickness $d$ matches the thermal value for the building and real-world scenes. The \textit{synthetic meshes} row gives the steel used for the end-to-end ground-truth validation (\cref{fig:e2e}) on the Suzanne, Spot, Bunny, and Armadillo meshes, run at native ground-truth scale.}
    \label{tab:scene_structural_materials}
    \begin{tabular}{llccccc}
        \toprule
        Scene(s)         & Material  & $L$ [\si{\meter}] & $E$ [\si{\giga\pascal}] & $\nu$ & $\rho$ [\si{\kilogram\per\meter\cubed}] & $d$ [\si{\meter}] \\
        \midrule
        BuildNet3D       & concrete  & ---               & 40                      & 0.20  & 2400                                    & 0.200             \\
        Building A       & concrete  & 15.0              & 40                      & 0.20  & 2400                                    & 0.200             \\
        Car              & steel     & 4.0               & 210                     & 0.30  & 7850                                    & 0.002             \\
        Woodshed         & wood      & 2.5               & 11                      & 0.35  & 600                                     & 0.020             \\
        Lion             & wool/felt & 0.3               & 0.003                   & 0.30  & 100                                     & 0.020             \\
        \midrule
        Synthetic meshes & steel     & ---               & 210                     & 0.30  & 7850                                    & 0.002             \\
        \bottomrule
    \end{tabular}
\end{table}

The generalized eigenvalue problem is solved with a shift-invert Lanczos iteration from SciPy~\cite{scipy}, using a sparse Cholesky factorization of the mass-shifted stiffness matrix---computed with CHOLMOD~\cite{chenCholmod2008} through \texttt{scikit-sparse}~\cite{scikit_sparse_0_5_0}---as the shift-invert operator; the spectral shift and its removal are detailed in Supplementary, \cref{appendix:reissner_mindlin}.

As for the thermal IGA simulation, we validate our IGA implementation of RM against a FEM-based baseline implemented in DOLFINx~\cite{barattaDOLFINx2023} (version 0.9.0)---see Supplementary, \cref{subsec:rm_fem_reference,suppmat:solvervalidation}.
The same FEM baseline is used in our end-to-end experiments on textureless meshes.

\subsection{Isogeometric Analysis Library}\label{implementation:IGA_lib}

The IGA framework, used for both the heat simulation and the Reissner-Mindlin modal analysis, was implemented within this work by leveraging the same primitives used in the differentiable renderer.
All tensor quantities---such as basis function evaluations, quadrature point batches, local element matrices, and control-points---are expressed via Dr.Jit~\cite{Jakob2020DrJit} primitives.
Similarly, the JIT-compiled, GPU-resident, and AD-enabled stack used to optimize the spline geometry is also used to assemble the IGA stiffness and mass matrices.
After assembling all local matrices for each patch, a local-to-global DOF ordering is used to construct the sparse
matrix, completing operator assembly.

\section{Data availability}

Source data for the Tables as well as meshes and simulation results are provided via GitHub (\url{https://github.com/Schindler-EPFL-Lab/FORGE-SIM}).

\section{Code Availability}

The source code used to optimize the models, assemble the simulation, generate the results, and produce the figures will be available via GitHub (\url{https://github.com/Schindler-EPFL-Lab/FORGE-SIM}).
The code is publicly available under a PolyForm Noncommercial License 1.0.0.

\bmhead{Supplementary information}

This article has supplementary materials.

\bmhead{Acknowledgements}

This research was supported by Innosuisse---Swiss Innovation Agency under Grant No. 105.237.1 IP-ICT, titled Insulated: Integrated Solution for Lean and Abridged Thermal Evaluation with Digital Twins, a collaborative project between EPFL and Schindler AG.

We want to thank Dr. Giuliano Guarino for help with mathematical aspects of the Reissner-Mindlin shell theory.

\section*{Declarations}

\bmhead{Competing interests}
The authors declare no competing interests.

\bibliographystyle{bst/sn-mathphys-num}
\bibliography{Shape-IGA,davor}%

\newpage
\begin{center}
    \noindent\textbf{\large Supplementary Materials}

    \vspace{0.3 cm}

    \noindent\textbf{Spline-Based Boundary Representations for Sparse View Reconstruction and Simulation Using Isogeometric Analysis}

    \vspace{0.3 cm}

\end{center}

\setcounter{section}{0}
\setcounter{subsection}{0}
\setcounter{subsubsection}{0}
\setcounter{figure}{0}
\setcounter{table}{0}
\setcounter{equation}{0}
\renewcommand{\thefigure}{S\arabic{figure}}
\renewcommand{\thetable}{S\arabic{table}}
\renewcommand{\theequation}{S\arabic{equation}}

\section{Shape Optimization}\label{secA1}

Let $R: \mathbb{R}^d \to \mathbb{R}^{p \times c} $ be a rendering function that takes as input scene parameters $\mathbf{\theta}$ and produces an output image with $p$ pixels and $c$ channels.
Shape optimization is the task of adjusting scene parameters so that the rendered image $\mathbf{y}$ matches the reference image $\mathbf{r}$.
This can formally be stated as an optimization problem:
\begin{equation}
    \minimize_\mathbf{\theta} \, \mathcal{L} \left (R (\mathbf{\theta}), \mathbf{r} \right)
\end{equation}
In shape optimization, a subset of scene parameters $\textbf{x} \subset \mathbf{\theta}$ that influence the shape are optimized, e.g. vertex positions of a triangular mesh.
It should be noted that the connectivity (topology) is not changed in the process.

Since the number of parameters is high, the gradient $\partial_{\mathbf{x}} \mathcal{L} (R(\mathbf{\theta}))$ is required for efficient optimization, so both the loss function and the rendering function must be differentiable almost everywhere.
Here, the gradient consists of two distinct components: shading gradients and silhouette gradients.
Shading gradients are defined on the whole surface and are typically small in magnitude.
On the other hand, silhouette gradients are defined on the visible surface boundary (silhouette) and have large magnitudes, if the reference and rendered silhouettes do not match.
As shown by \citet{nicoletLargeStepsInverse2021a}, the latter are so strong that naive gradient descent fails outright.
While using appropriate regularization can help, it still often leads to self-intersections.

\subsection{Large Steps}\label{subsubsec_large_steps}

Gradient preconditioning was introduced in \citet{nicoletLargeStepsInverse2021a} and it solved the problem of large sparse gradients that cause self-intersections by introducing a latent variable $\textbf{v}$
\begin{equation}
    \mathbf{v}(\mathbf{x}) = (\mathbf{I} + \lambda \mathbf{L}) \, \mathbf{x},
\end{equation}
which, for gradient descent on the latent variable, yields the position update
\begin{equation}
    \mathbf{x} \gets \mathbf{x} - \delta (\mathbf{I} + \lambda \mathbf{L})^{-2} \frac{\partial \mathcal{L}}{\partial \mathbf{x}},
\end{equation}
with $\mathbf{L}$ the combinatorial Laplacian, $\lambda$ a tunable parameter that determines how strongly the gradients are diffused, and $\delta$ the step size.
In practice, \citet{nicoletLargeStepsInverse2021a} pair this latent parametrization with the Adam optimizer using uniform second-moment estimates, which weakens the effective preconditioning to a single power of $(\mathbf{I} + \lambda \mathbf{L})^{-1}$.
Note that this update requires solving linear systems at each step. However, since the combinatorial Laplacian does not change throughout the optimization, sparse Cholesky factorization is performed at the beginning, requiring only forward and backward substitutions at each optimization step. This scheme was also found to improve the stability of spline-based shape optimization~\cite{worchelDifferentiableRenderingParametric2023}.

As noted by \citet{nicoletLargeStepsInverse2021a}, despite its advantages, preconditioning is not a regularization term and the optimized shape typically reconstructs less fine detail when compared to a regularized baseline.
This can mostly be remedied by carefully choosing the smoothing strength and reducing it as the optimization progresses.
The improved stability allows for much larger step sizes and often more than makes up for a slight reduction in detail in triangular mesh optimization.
However, in the context of spline optimization we found that these smoothing effects coupled with the inherent smoothness of splines can lead to excessive smoothing (see Methods in the main text).

\subsection{Differentiable Tessellation}\label{subsubsec_diff_tessellation}

Since spline surfaces are not natively supported in hardware, it is necessary to introduce an intermediate representation that can be rendered efficiently while preserving differentiability.
Spline primitives that we use for shape optimization are formally introduced in \cref{appendix:subsec_b_splines}.
We follow the method proposed by \citet{worchelDifferentiableRenderingParametric2023} and tessellate the spline surface.

Let $\mathbf{x}_s \in \mathbb{R}^{n_t \times 3}$ be the vertex positions of a tessellation of a multi-patch spline surface and let $\mathbf{c} \in \mathbb{R}^{n_c \times 3}$ be a vector of its control points. A differentiable renderer, such as Mitsuba~\cite{Mitsuba3}, computes the gradient
\begin{equation}
    \frac{\partial \mathcal{L}}{\partial \mathbf{x}_s}(R(\mathbf{\theta}), \mathbf{r}) = \left( \frac{\partial R}{\partial \mathbf{x}_s}(\mathbf{\theta}) \right)^T \frac{\partial \mathcal{L}}{\partial R} (R(\mathbf{\theta}), \mathbf{r}),
\end{equation}
while automatic differentiation ensures that the gradients are propagated to the appropriate control points
\begin{equation}
    \frac{\partial \mathcal{L}}{\partial \mathbf{c}}(R(\mathbf{\theta}), \mathbf{r}) = \left( \frac{\partial \mathbf{x}_s}{\partial \mathbf{c}}(\mathbf{c}) \right)^T \frac{\partial \mathcal{L}}{\partial \mathbf{x}_s} (R(\mathbf{\theta}), \mathbf{r}).
\end{equation}

At this point the parametrization from the previous subsection can be applied to the control points while taking into account that the control points form quads instead of triangles~\cite{worchelDifferentiableRenderingParametric2023}.

\section{Numerical Quadrature}\label{appendix:numerical_quadrature}

Evaluating integrals in an efficient manner is the building block for a wide range of algorithms in applied mathematics and engineering, ranging from assembly routines in finite element analysis to signal processing.
The integral is typically given by a sufficiently regular integrable function $f: \mathbb{R}^d \to \mathbb{R}$, with $d$ being the dimensionality of the input.
The integral can then be evaluated analytically by hand, using symbolic computation, or by means of numerical quadrature.
Analytic evaluation is the most efficient when applicable, while symbolic computation can produce highly complex expressions that are prohibitively expensive to evaluate in spite of being exact.
Thus, numerical quadrature is the only computationally tractable option for complex integrands.

Numerical quadrature is based on estimating the integral of $f$, denoted $I(f)$, by evaluating $f$ at a finite set of points
\begin{equation}\label{eq_quadrature}
    I(f) :=\int_\Omega f(\boldsymbol{x})\, dV \approx \sum_{i=1}^{n_q} w_i f(\boldsymbol{x}_i),
\end{equation}
where $\{\boldsymbol{x}_i\}_{i=1}^{n_q}$ is the set of evaluation points, referred to as quadrature nodes, and $\{w_i\}_{i=1}^{n_q}$ is the set of corresponding contributions, referred to as quadrature weights.
The pair $(\boldsymbol{x}_i, w_i)$ constitutes a quadrature point.
Quadrature nodes can lie outside of the domain and weights are arbitrary real numbers.
However, enforcing that the nodes lie entirely inside the domain of integration and that the weights are strictly positive is important for the stability of downstream applications, such as FEA simulation.

\subsection{Simple Quadrature Rules}\label{subsec_simple_quadrature}

In one spatial dimension ($d=1$), the two simplest quadrature formulas for integrating $f$ over an interval $[a, b]$ are the mid-point and the trapezoidal rule.
They approximate $I(f)$ as
\begin{equation}
    I(f) \approx I_M(f) = (b-a) f\left(\frac{a+b}{2} \right), \qquad I(f) \approx I_T(f) = \frac{b-a}{2}(f(a) + f(b)).
\end{equation}

Sampling the function only at the midpoint or at the endpoints is insufficient for all but the simplest integrands, motivating the need for subdivision: the interval is partitioned into $n$ sub-intervals, typically of the same length $h=(b-a) / n$.
The simple rules can then be applied on each sub-interval, resulting in $n$ function evaluations for the midpoint rule and $n+1$ for the trapezoidal.
These are referred to as composite quadrature rules.
The composite midpoint rule is used within the scope of the paper and is given as
\begin{equation}
    I(f) \approx I_M^n(f) = \frac{b-a}{n} \sum_{i=0}^{n-1} f\left(a + \frac{b-a}{n} \left(i+\frac{1}{2}\right) \right).
\end{equation}

The integral estimate is improved as $n$ increases with an algebraic order $\mathcal{O}(n^{-2})$.
The trapezoidal rule has an equivalent formula, but since it does not perform better on arbitrary integrands, it is not considered.

\subsection{Gauss--Legendre Quadrature}\label{subsec_gauss_legendre}

For the case of one spatial dimension ($d=1$), Gaussian quadrature is the most important family of quadrature methods.
It is defined by the optimality condition: given a positive weight function $\omega: \mathbb{R} \to \mathbb{R}^+$, Gaussian quadrature of order $n_q$ generates a set of $n_q$ quadrature points for integrands of form $\omega(x) f(x)$ such that the polynomial degree of exactness is theoretically maximal ($2n_q - 1$) for $f$.
Practically, this means that if $f$ is a polynomial function of degree $2n_q-1$, it can be integrated exactly.
More formally,
\begin{equation}
    I(f) := \int_a^b \omega(x) \, f(x) \, dx, \quad I_G^{n_q}(f) := \sum_{i=1}^{n_q} w_i f(x_i).
\end{equation}

Different choices of the weight function yield different schemes tailored to the integrand.
The simplest choice $\omega(x) = 1$ gives the Gauss--Legendre quadrature.
It is defined on a reference interval $[-1, 1]$.
The reference nodes $\{\hat x_i\}_{i=1}^{n_q}$ are the zeros of the Legendre polynomial of order $n_q$, denoted $P_{n_q}(x)$, with the corresponding reference weights $\hat w_i = 2 / \left[ (1-\hat x_i^2) (P_{n_q}'(\hat x_i))^2 \right]$.

The reference interval $[-1, 1]$ is mapped to an arbitrary one $[a, b]$, with $b > a$ and $a, b \in \mathbb{R}$, using an affine transform, yielding nodes and weights on $[a, b]$
\begin{equation}
    x_i = \frac{b+a}{2} + \frac{b-a}{2} \hat x_i, \quad w_i = \frac{b-a}{2} \hat w_i, \quad \forall i \in \{1, ..., n_q\}.
\end{equation}

\noindent Gauss--Legendre quadrature has the following properties~\cite{stoer_1980_intro}:
\begin{enumerate}
    \item \emph{Nodes inside $[a, b]$}:
          Quadrature nodes lie entirely inside of the interval $[a, b]$.
          In particular, $a < x_1 < \dots < x_{n_q} < b.$
    \item \emph{Positive weights}:
          The quadrature weights are strictly positive, i.e., $w_i > 0, \forall i \in \{1, \dots, n_q\}$.
    \item \emph{Polynomial exactness}:
          Polynomials of degree up to $2n_q-1$ are integrated exactly.
    \item \emph{High-order convergence}:
          If the function $f$ cannot be integrated exactly, then the error depends on the size of the integration interval and the $2n_q$-th derivative of $f$:
          \begin{equation}
              I(f) - I_{G}^{n_q}(f) = \frac{f^{(2n_q)}(\xi)}{(2n_q)!} \int_a^b \prod_{i=1}^{n_q} (x-x_i)^2 \, dx,
          \end{equation}
          for some $\xi \in (a, b)$.
          This assumes that the $2n_q$-th derivative of $f$ is well-defined, otherwise the formula does not hold.
    \item \emph{Spectral convergence for analytic functions}:
          If $f$ is analytic on interval $[a, b]$ and in its complex neighborhood, then the error decays geometrically as $C \rho^{-2n_q}$, where $C$ is a constant that depends on the interval length and $f$, while $\rho > 1$ depends on the size of the complex neighborhood (Bernstein ellipse) in which $f$ is analytic.
\end{enumerate}

\noindent The polynomial exactness property is particularly important as it allows for efficient integration of expressions that arise in FEA.
It is particularly important because failing to integrate polynomial terms can lead to under-integration, a phenomenon where a polynomial subspace is ``not seen'' by the quadrature scheme, potentially introducing large errors.
Therefore, to preserve the consistency of the weak formulation of a PDE, the quadrature order has to be at least $p+1$, where $p$ is the polynomial degree of the space.
This follows from mass-matrix entries being products of the polynomial basis functions, yielding a total degree $2p$.

\subsection{Quadrature in Higher Dimensions}\label{subsec_tensor_quad}

In higher dimensions ($d \geq 2$), there is no universal way of constructing an optimal quadrature rule for arbitrarily shaped domains.
In case of certain common shapes such as simplices and hypercubes closed form expressions can be obtained.

For hypercubes, the situation is particularly simple: tensor-product Gauss--Legendre quadrature can be applied which is optimal for integrating tensor-product polynomial basis functions typically defined on such domains.
In $d=2$ for a reference $[-1, 1]^2$ hypercube we have nodes $\boldsymbol{\hat x}_{i, j} := (\hat x_i, \hat x_j)$ and the corresponding weights $\hat w_{i, j} := \hat w_i \hat w_j$.
Quadrature order can be different along the first and the second direction.
Tensor-product Gauss--Legendre quadrature can be directly applied to computing integrals over spline surfaces, noting that they are $C^\infty$ inside of each knot span.
This means that the integral over a patch is first decomposed into a sum of integrals over individual knot spans, each of which is then mapped onto the reference square and evaluated with the tensor-product rule.

Note that the composite midpoint rule can also be applied, but its low order of convergence means that a large number of quadrature points are required to achieve the same level of accuracy as Gauss--Legendre quadrature.
In $d=1$ the difference in number of quadrature points can already be one or more orders of magnitude, and the midpoint rule is often computationally prohibitive already for $d=2$ if high accuracy is required.

\section{B-Splines}\label{appendix:subsec_b_splines}

B-splines are the fundamental primitive that we use for both shape optimization and isogeometric analysis.
We introduce them in a manner similar to \citet{hughesIsogeometricAnalysisCAD2005a} and \citet{dedeIsogeometricAnalysisSecond2015}.
These two papers serve as an overarching reference until the end of this section.

\subsection{Knot Vector}\label{subsubsec_knot_vector}

B-splines are a set of $n$ piece-wise polynomial basis functions of degree $p$ defined on an interval in parametric space $\xi \subset \mathbb{R}$, by a \textit{knot vector}.
The knot vector is a set of coordinates in parametric space $\Xi = \{\xi_1, \xi_2, ..., \xi_{n+p+1} \}$, $\xi_{i} \leq \xi_{i+1}$ that determines the shape of the  basis functions (see \cref{subsubsec_basis_functions} for the definition of B-spline basis functions).

The knot vector is defined by first choosing $s + 1$ distinct coordinates in parametric space $\Xi^*=\{\xi_1^*, \xi_2^*, ..., \xi_{s+1}^* \}$, $\xi_i^* < \xi_{i+1}^*$ that are called \textit{unique knots} and form $s$ \textit{knot spans}.
The full knot vector is assembled by repeating each unique knot by its \textit{multiplicity}.
Knot multiplicity determines the smoothness of the basis functions at the knot position: the number of continuous derivatives there equals $p$ minus the knot's multiplicity.

A knot vector is called \textit{open} (clamped) if the first and last unique knots are repeated $p+1$ times, while other unique knots, called \textit{interior} knots, can have a multiplicity of up to $p$.
\textit{Open uniform} (OU) knot vectors are common in CAD applications and further impose that all unique knots are equally spaced and interior knots have a multiplicity of 1.

This choice yields a set of basis functions that smoothly approximates the values with $C^{p-1}$ continuity for all $\xi \in (\xi_1^*, \xi_{s+1}^*)$, while it is \textit{interpolatory} at the edge of the interval: a single basis function is non-zero at those two points.
There are a total of $n = s + p$ basis functions.

\subsection{Basis Functions}\label{subsubsec_basis_functions}

The B-spline basis functions are defined in a recursive manner based on the knot vector; basis functions of degree $p=0$ are given as
\begin{equation}
    N_{i, 0} (\xi) =
    \begin{cases}
        1, & \text{if} \,\, \xi_{i} \leq \xi < \xi_{i+1}, \\
        0, & \text{otherwise},
    \end{cases}
\end{equation}
while for degree $p \geq 1$ they are computed as a weighted combination of two basis functions of degree $p-1$
\begin{equation}
    N_{i,p}(\xi) = \frac{\xi - \xi_i}{\xi_{i+p} - \xi_{i}} N_{i, p-1} (\xi) + \frac{\xi_{i+p+1} - \xi}{\xi_{i+p+1} - \xi_{i+1}} N_{i+1, p-1}(\xi),
\end{equation}
where fractions with a zero denominator are taken to be zero.
A stable numerical implementation, based on the Cox-de-Boor algorithm, is given in the NURBS book \cite{pieglNURBSBook1997}, (Algorithm A2.2).
An analogous algorithm exists for computing all non-zero derivatives (up to order $p$) of the B-spline basis functions (Algorithm A2.3). B-spline basis functions have 3 important properties:
\begin{enumerate}
    \item Partition of unity: the sum of basis functions at a given $\xi$ equals $1$ for all $\xi \in \left[\xi_1, \xi_{n+p+1} \right]$.
    \item Compact support: each basis function $N_{i, p}(\xi)$ is non-zero on $\xi \in [\xi_i, \xi_{i+p+1}]$ and zero elsewhere.
    \item Non-negativity: $N_{i, p}\geq 0, \forall\xi \in \left[\xi_1, \xi_{n+p+1} \right]$.
\end{enumerate}

\par
\begin{figure}
    \centering
    \includegraphics[width=3.3in]{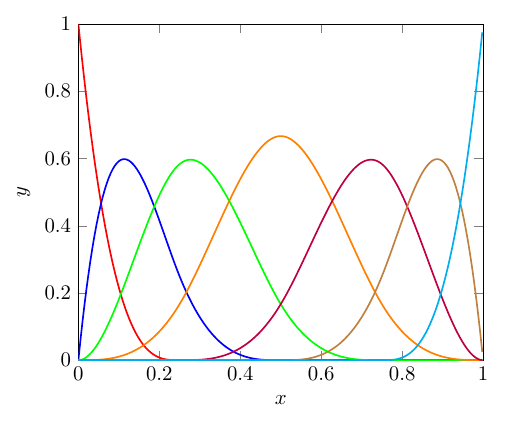}
    \caption{Cubic B-spline basis functions. Knot vector is $\Xi = \{0.0, 0.0, 0.0, 0.0, 0.25, 0.5, 0.75, 1.0, 1.0, 1.0, 1.0\}$}
    \label{fig_bspline_basis}
\end{figure}

\begin{figure}
    \centering
    \includegraphics[width=3.3in]{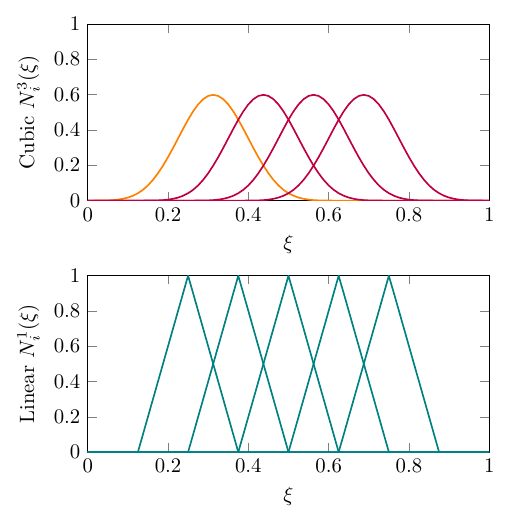}
    \caption{Local support of a cubic B-spline basis function: each basis function is non-zero on at most $p+1$ knot spans. Knot vector is $\Xi = \{0.0, 0.0, 0.0, 0.0, 0.25, 0.5, 0.75, 1.0, 1.0, 1.0, 1.0\}$}
    \label{fig_bspline_support}
\end{figure}

 Cubic B-spline basis functions for a uniform open knot vector with 4 knot spans are shown in \cref{fig_bspline_basis}. Observe that there are at most $p+1$ non-zero (active) basis functions in a particular knot span and that each basis function is locally supported (non-zero) on at most $p+1$ knot spans (see \cref{fig_bspline_support}). Two basis functions interact if they are both non-zero in a given knot span. It follows that each basis function interacts with at most $2p$ other basis functions.

\subsection{B-spline curves}\label{subsubsec_bspline_curves}

\par Once defined, taking a linear combination of B-spline basis functions weighted by values $\{\boldsymbol{c}_1, \boldsymbol{c}_2,..., \boldsymbol{c}_n \}$ constructs a smooth approximation of these values
\begin{equation}
    C(\xi) = \sum_{i=1}^{n} \boldsymbol{c}_i N_{i, p}(\xi)
\end{equation}
$C(\xi)$ represents a mapping $C: \xi \to \mathbb{R}^d$ with $d$ being the number of components of each value.
If these values represent coordinates in 3D space, then they define a B-spline curve and are commonly referred to as \textit{control points}.
If an open knot vector is used, the first and the last control point are guaranteed to coincide with the value of the curve at the edges of the parametric interval, i.e. $C(\xi_1) = \boldsymbol{c}_1$ and $C(\xi_{n+p+1}) = \boldsymbol{c}_{n}$.

The first derivative of a degree $p$ curve can be computed in two ways: by computing the basis functions of degree $p-1$ and combining them with control point differences \cite{pieglNURBSBook1997}, or by directly computing the derivatives of the degree $p$ basis functions.
The first approach yields
\begin{equation}
    C'(\xi) = p \sum_{i=1}^{n-1} \frac{\boldsymbol{c}_{i+1} - \boldsymbol{c}_i}{\xi_{i+p+1} - \xi_{i+1}} N_{i+1, p-1} (\xi)
\end{equation}
Note that the basis functions of degree $p-1$ share the same knot vector as degree $p$ basis functions, which causes the first and the last basis functions to evaluate to $0$.
While this formulation is useful in certain cases, computing the derivatives of basis functions results in simpler expressions and generalizes to higher-order.
Furthermore, basis function derivatives are used in isogeometric analysis (see \cref{appendix:iga}).
The $k$-th derivative of the approximation is given as
\begin{equation}
    C^{(k)}(\xi) = \sum_{i=1}^{n} \boldsymbol{c}_i N^{(k)}_{i, p}(\xi)
\end{equation}

\subsubsection{Knot Insertion}\label{sup_mat:subsubsec_knot_insertion}

A curve composed of multiple line segments can be refined by reducing the length of each line segment and increasing the number of segments, allowing it to more smoothly represent the shape and capture more detail.
An equivalent of such a procedure in the context of B-splines is \textit{h-refinement}.
The h-refinement procedure is based on repeatedly applying the \textit{knot insertion algorithm}.
Starting from the original knot vector, a knot is inserted at an arbitrary position in the parametric domain $\bar \xi \in (\xi_1, \xi_{n+p+1})$, increasing the number of basis functions by 1 without changing the geometry of the curve.

\par Let $\Xi = \{\xi_i\}_{i=1}^{n+p+1}$ be an arbitrary knot vector defining $n$ basis functions and let $\{\boldsymbol{c}_i\}_{i=1}^{n}$ be the values that are smoothly approximated.
The new knot is inserted into the interval $[\xi_k, \xi_{k+1})$.
Then, the new values $\{ \boldsymbol{\bar c}_i\}_{i=1}^{n+1}$ are computed from the original ones as
\begin{equation}
    \boldsymbol{\bar c}_i = (1 - \alpha_i) \boldsymbol{c}_{i-1} + \alpha_i \boldsymbol{c}_i, \quad i \in \{1, 2, ..., n+1\},
\end{equation}
with $\{\alpha_i\}_{i=1}^{n+1}$ being a set of weights defined as
\begin{equation}
    \alpha_i =
    \begin{cases}
        1,                                          & 1 \leq i \leq k - p,     \\
        \frac{\bar \xi - \xi_i}{\xi_{i+p} - \xi_i}, & k - p + 1 \leq i \leq k, \\
        0,                                          & k + 1 \leq i \leq n + 1,
    \end{cases}
\end{equation}
If the inserted knot coincides with any of the unique knots, then the continuity of the spline is reduced at that knot.
This is avoided in h-refinement as continuity properties of the smooth approximation must be preserved.

\par Selecting an appropriate set of knot positions that increase resolution where it is needed is not a straightforward task \cite{hughesIsogeometricAnalysisCAD2005a}.
We restrict our attention to the simplest case; the knot vector is open uniform and each knot span is subdivided into $d$ equal-length segments.
This increases the number of basis functions by almost $d$ times, reducing the size of the local support of each basis function by $d$ times in the parametric domain.

\subsection{B-Spline Surfaces}\label{subsec_bspline_surfaces}

\par Let $\Xi = \{\xi_1, \xi_2,... , \xi_{n+p+1} \}$ and $\mathcal{H} = \{\eta_1, \eta_2, ..., \eta_{m+q+1} \}$ be two knot vectors representing two B-splines, the first one of degree $p$ with $n$ basis functions, and the second one of degree $q$ with $m$ basis functions.
To smoothly approximate the values defined on a two-dimensional (2D) rectangular domain $\hat \Omega =  [\xi_1, \xi_{n+p+1}] \times [\eta_1, \eta_{m+q+1}]$, a \textit{bi-variate tensor product B-spline} basis is defined as
\begin{equation}
    B_{i, j}(\xi, \eta) = N_{i, p}(\xi) \, M_{j, q}(\eta),
\end{equation}
where $i \in \{1, 2, ..., n\}$ denotes the $i$-th basis function along the $\xi$ direction and $j \in \{1, 2, ..., m\}$ denotes the $j$-th basis function along the $\eta$ direction.
The parametric domain is partitioned into a grid of $s_\xi \times s_\eta$ knot spans.
A B-spline surface is defined by a grid of control points $\boldsymbol{c}_{i, j}$
\begin{equation}
    S(\xi, \eta) = \sum_{i=1}^n \sum_{j=1}^m \boldsymbol{c}_{i, j} \, N_{i, p}(\xi) \, M_{j, q}(\eta).
\end{equation}
If uniform open knot vectors are used for both directions, the resulting surface is $C^\infty$ within knot spans and $C^{\min \{p, q \} - 1}$ across interior knot lines, and it is interpolatory at the four corners.

\subsubsection{Properties}

Since the bi-variate B-spline is a tensor product of two B-splines, there are $(p+1) \times (q+1)$ active basis functions in a given knot span, with $p$ being the polynomial degree of the B-spline along $\xi$ and $q$ the degree of the B-spline along $\eta$.
Following the same logic as in \cref{subsubsec_basis_functions}, each basis function interacts with at most $(2p+1) \times (2q+1) - 1$ other basis functions.
For $p=q=3$ and open uniform knot vectors along both directions, most basis functions interact with 48 other basis functions.
Since they are non-zero on 4 intervals, other basis functions they interact with extend over a $7 \times 7$ knot span region.
This is much more than in a triangle mesh, where each vertex interacts with 6 other vertices (on average) that are directly adjacent to it.
The fact that B-spline basis functions have a larger local support than triangle vertices leads to more diffuse gradients when performing shape optimization as well as discouraging abrupt deformation.

Knot insertion is performed in the same manner as for uni-variate B-splines, shown in \cref{sup_mat:subsubsec_knot_insertion}, but multiple control points are added at the same time.
Inserting a knot into $\Xi$ adds $m$ control points as the updated control grid has $(n+1) \times m$ control points.
The situation is analogous when inserting a knot into $\mathcal{H}$.
This represents a \textit{global refinement} scheme as the entire row/column of the control point grid is updated.

\subsubsection{Differential Geometry}\label{subsubsec_diff_geometry}

Computing the surface normals requires evaluating the partial derivatives with respect to parametric directions
\begin{equation}
    \begin{aligned}
        \partial_{\xi} S (\xi, \eta) = \sum_{i=1}^n \sum_{j=1}^m \boldsymbol{c}_{i, j} \, N_{i, p}'(\xi) \, M_{j, q}(\eta), \\
        \partial_{\eta} S (\xi, \eta) = \sum_{i=1}^n \sum_{j=1}^m \boldsymbol{c}_{i, j} \, N_{i, p}(\xi) \, M_{j, q}'(\eta),
    \end{aligned}
\end{equation}
For the surface mapping $S: \hat \Omega \to \mathbb{R}^3$ the Jacobian $\mathbf{F}: \hat \Omega \to \mathbb{R}^{3 \times 2}$ is defined by the aforementioned partial derivatives as column vectors.
The first fundamental form of the mapping $\mathbf{G}: \hat \Omega \to \mathbb{R}^{2 \times 2}$ follows directly from the Jacobian
\begin{equation}
    \mathbf{G}(\xi, \eta) = \left(\mathbf{F}(\xi, \eta) \right)^T  \,\mathbf{F}(\xi, \eta)
\end{equation}
and $g: \hat \Omega \to \mathbb{R}$ is the area scaling factor used to evaluate the integrals over the surface by integrating over $\hat \Omega $
\begin{equation}
    \begin{aligned}
        g(\xi, \eta) = \sqrt{\det \left( \mathbf{G}(\xi, \eta) \right)}, \\
        dA(\xi, \eta) = g(\xi, \eta) \, d\xi \, d\eta.
    \end{aligned}
\end{equation}

The unit normal is a normalized cross product of the two partial derivatives
\begin{equation}
    \boldsymbol{\hat n}(\xi, \eta) = \frac{\partial_\xi S(\xi, \eta) \times \partial_\eta S(\xi, \eta)}{||\partial_\xi S(\xi, \eta) \times \partial_\eta S(\xi, \eta) ||}.
\end{equation}

Finally, computing curvature requires evaluating the second derivatives of the surface mapping and together they form the Hessian $\mathbf{H}: \hat \Omega \to \mathbb{R}^{3 \times 2 \times 2}$
\begin{equation}
    \mathbf{H} (\xi, \eta)= \begin{bmatrix}
        \partial_{\xi\xi} S(\xi, \eta)  & \partial_{\xi\eta} S(\xi, \eta)  \\
        \partial_{\eta\xi} S(\xi, \eta) & \partial_{\eta\eta} S(\xi, \eta) \\
    \end{bmatrix}.
\end{equation}
Carrying out a dot product with the normal along the first direction of the Hessian tensor yields the second fundamental form:
\begin{equation}
    \mathbf{II} (\xi, \eta) =
    \begin{bmatrix}
        \boldsymbol{\hat n}(\xi, \eta) \cdot \partial_{\xi\xi} S(\xi, \eta)  & \boldsymbol{\hat n}(\xi, \eta) \cdot \partial_{\xi\eta} S(\xi, \eta)  \\
        \boldsymbol{\hat n}(\xi, \eta) \cdot \partial_{\eta\xi} S(\xi, \eta) & \boldsymbol{\hat n}(\xi, \eta) \cdot \partial_{\eta\eta} S(\xi, \eta) \\
    \end{bmatrix},
\end{equation}
Principal curvatures $k_1$ and $k_2$ are the eigenvalues of the shape operator $\mathbf{S}: \hat\Omega \to \mathbb{R}^{2 \times 2}$
\begin{equation}
    \mathbf{S}(\xi, \eta) = \mathbf{G}^{-1}(\xi, \eta) \, \mathbf{II}(\xi, \eta).
\end{equation}
Willmore energy of a B-spline surface is a functional
\begin{equation}
    E_W  = \frac{1}{4} \int_{\xi_1}^{\xi_{n+p+1}} \int_{\eta_1}^{\eta_{m+q+1}} (k_1 - k_2)^2 \, g(\xi, \eta) \, d\xi d\eta.
\end{equation}

\subsection{Computing Willmore Energy}
\label{sup_amt:willmore}

Willmore energy is the central component in curvature regularization, but it cannot be evaluated analytically.
Leveraging the insight from  \cref{subsec_tensor_quad}, we can apply tensor-product quadrature along parametric directions $\xi$ and $\eta$.
However, to ensure optimal convergence and exactness, the integrand has to be smooth.
Thus, the integral over the parametric domain is split along knot spans.
For a knot span with indices $(i, j)$, $i$ and $j$ denoting the knot span indices along $\xi$ and $\eta$, respectively, the integral over the span is given by:
\begin{equation}\label{sec:method:reg:curvature:integration}
    E_{i, j} = \frac{1}{4} \int_{\xi_i^*}^{\xi_{i+1}^*} \int_{\eta_j^*}^{\eta_{j+1}^*} \left(k_1(\xi, \eta) - k_2(\xi, \eta)\right)^2 \, g(\xi, \eta) \, d\xi d\eta,
\end{equation}
where $g(\xi, \eta)$ is the metric, the square root of the determinant of the metric tensor $\mathbf{G}(\xi, \eta)$, and $\xi_i^*$ and $\eta_j^*$ are the unique knot values (see \cref{subsubsec_knot_vector}).

The integrand consists of a rational polynomial multiplied by a square-root term, making exact integration impractical.
To avoid under-integration---where an insufficient number of quadrature points fails to capture the polynomial components---the quadrature order must be chosen sufficiently high.
Owing to the tensor-product structure, the full polynomial degree can be treated in terms of the bi-degree $(p, p)$.
The polynomial component of \cref{sec:method:reg:curvature:integration} has bi-degree $(6p-6, 6p-6)$, which would suggest using $3p-2$ quadrature points per direction.
We found that Gauss--Legendre quadrature of order $2p$ is sufficient to avoid under-integration; increasing it further did not yield measurable accuracy improvements, likely due to single precision arithmetic.

The per-knot-span Willmore energy, approximated using tensor-product Gauss--Legendre quadrature of order $n$, is given by:
\begin{equation}
    E_{i, j} = \frac{1}{4} (\xi_{i+1}^* - \xi_{i}^*) (\eta_{j+1}^* - \eta_{j}^*) \sum_{k=1}^{n} \sum_{l=1}^{n} \frac{\hat w_{k}}{2} \frac{\hat w_{l}}{2} \left(k_1(\xi_k, \eta_l) - k_2(\xi_k, \eta_l)\right)^2 \, g(\xi_k, \eta_l),
\end{equation}
where $n = 2p$ is the quadrature order, $\{{\xi_k}\}_{k=1}^n$ and $\{{\eta_l}\}_{l=1}^n$ are the sets of Gauss--Legendre points on the intervals $[\xi_{i}^*, \xi_{i+1}^*]$ and $[\eta_{j}^*, \eta_{j+1}^*]$, respectively, and $\{{\hat w_{k}}\}_{k=1}^n$ and $\{{\hat w_{l}}\}_{l=1}^n$ are the corresponding reference weights on $[-1, 1]$ (see \cref{subsec_gauss_legendre}).

The total Willmore energy $\mathcal{L}_{\text{W}}$ is obtained by summing the contributions over all knot spans.
The computational cost of this method is an order of magnitude higher than Laplacian regularization, so evaluating the surface at as few points as possible is important for efficient optimization.
The total number of integrand evaluations scales as $(2ps)^2$, which remains below $256^2$ for $p=3$ and $s \leq 32$, values used throughout most of the optimization.
The midpoint rule would require more evaluations, and we also found it is susceptible to severe under-integration, failing to properly penalize high curvature; for $s=64$, avoiding under-integration would require increasing the number of evaluations to $512^2$.
Gauss--Legendre quadrature thus yields a significant speedup without compromising accuracy.

\section{Isogeometric Analysis}\label{appendix:iga}

\par Isogeometric analysis~\cite{hughesIsogeometricAnalysisCAD2005a}, commonly referred to as IGA, is a discretization approach to solving \textit{Partial differential equations} (PDEs), closely related to \textit{Finite element analysis} (FEA).
In FEA, one proceeds by defining a mesh comprised of a large number of elements with pre-defined shape that form an approximation of the exact domain $\Omega$, followed by a discretization of the appropriate functional space.
A common choice for problems that require $H^1(\Omega)$ regularity is to define the solution as a linear combination of Lagrange interpolating polynomials, achieving $C^0$ continuity along the boundary.
In contrast, the geometry in IGA is defined by spline control points and \textit{the same} basis functions that define the geometry are used to represent the solution.
This means that the geometry can be represented exactly if the domain is described by splines, removing the aforementioned approximation error inherent to FEA.

\par IGA bears some resemblance to \textit{isoparametric} FEM; it uses the basis functions used to discretize the solution to represent the local element geometry.
The crucial difference is that the geometric mapping is piecewise-polynomial, local to each element, and is primarily intended to reduce the error associated with approximating $\Omega$.
Furthermore, isoparametric FEM has $C^0$ continuity, as it still uses interpolating polynomials, while IGA typically exhibits $C^{p-1}$ continuity, with $p$ being the polynomial degree of the spline basis.

\par If the solution is expected to be smooth on the domain, IGA has a clear advantage in computational efficiency.
Consider a 1D problem discretized into $n$ elements in the context of FEM and $n$ knot spans in the context of IGA, respectively.
The number of \textit{degrees of freedom} (DOFs) for FEM scales as $O(np)$, while for splines with maximum continuity it is merely $O(n+p) \approx O(n)$ for small $p$ and large $n$.
As the number of spatial dimensions increases the difference becomes exponentially larger: the number of DOFs scales as $O(n^d p^d)$ for FEM while it scales as $O((n+p)^d) \approx O(n^d)$ for IGA.
Note that FEM basis functions interact with fewer other basis functions on average compared to spline basis functions that typically interact with $\mathcal{O}((2p)^d)$ other basis functions.
The combined effect is that IGA matrices are significantly smaller than FEA matrices for a given number of elements/knot spans while the IGA matrices usually have many more off-diagonal terms, resulting in higher bandwidth.
From a functional space approximation point of view, if the solution is sufficiently regular, IGA achieves the same accuracy with significantly fewer degrees of freedom.

\par Another advantage of IGA lies in the spline bases of choice, e.g., the commonly used B-splines introduced in \cref{appendix:subsec_b_splines}.
They are a non-negative basis that forms a partition of unity, yielding mass matrices with non-negative entries---strictly positive for basis functions with overlapping support.
The higher continuity of the basis is especially useful for problems that require higher than $C^0$ continuity, such as Kirchhoff-Love shell theory~\cite{guarinoIBCMShells2024}, as well as problems that only require $C^0$ continuity but are prone to locking, such as Reissner-Mindlin shell theory~\cite{bensonIsogeometricShellAnalysis2010}.
The latter will be discussed in more detail in \cref{appendix:reissner_mindlin}.

\section{Heat Equation}\label{appendix:heat_equation}

\subsection{Formal Statement}

\par Let $\Omega$ be a closed subset of $\mathbb{R}^3$ and let $\partial\Omega$ be a Lipschitz continuous boundary of $\Omega$. Furthermore, let the boundary be decomposed into two components $\partial \Omega = \partial \Omega_D \cup \partial \Omega_N$, $\partial\Omega_D \cap \partial \Omega_N = \emptyset$, where $\partial\Omega_D$ denotes the component where Dirichlet boundary conditions are imposed, and $\partial \Omega_N$ denotes the component with Neumann boundary conditions.
Then, the 3D heat equation is a linear, second-order partial differential equation
\begin{equation}
    \begin{cases}
        C(\boldsymbol{x}) \, \partial_t u(\boldsymbol{x}, t) -   \nabla \cdot \left (\mathbf{K}(\boldsymbol{x}) \, \nabla u(\boldsymbol{x}, t) \right) = f(\boldsymbol{x}, t), & \boldsymbol{x} \in \Omega, t >0       \\
        u(\boldsymbol{x}, 0) = u_0(\boldsymbol{x}),                                                                                                                            & \boldsymbol{x} \in \Omega             \\
        u(\boldsymbol{x}, t) = u_D(\boldsymbol{x}, t),                                                                                                                         & \boldsymbol{x} \in \partial \Omega_D, \\
        \left(\mathbf{K}(\boldsymbol{x}) \, \nabla u(\boldsymbol{x}, t)\right) \cdot \boldsymbol{\hat n}(\boldsymbol{x}) = F_N(u, \boldsymbol{x}, t)                           & \boldsymbol{x} \in \partial \Omega_N
    \end{cases}
\end{equation}
where $u(\boldsymbol{x}, t)$ is the temperature at each $\boldsymbol{x} \in \Omega$ and time $t \in [0, T]$ [\si{\kelvin}], $C(\boldsymbol{x})$ is the spatially varying volumetric heat capacity  [\si{\joule\per\cubic\meter\per\kelvin}], $\mathbf{K}(\boldsymbol{x})$ is a spatially varying thermal conductivity tensor [\si{\watt\per\meter\per\kelvin}], $f(\boldsymbol{x}, t)$ is the spatially varying, time-dependent volumetric source term [\si{\watt\per\cubic\meter}], $u_0(\boldsymbol{x})$ is the initial condition (temperature at $t=0$), $u_D (\boldsymbol{x}, t)$ is the time-dependent Dirichlet boundary condition, and $F_N(u, \boldsymbol{x}, t)$ is the non-linear, spatially varying, time-dependent incoming Neumann flux [\si{\watt\per\square\meter}], with $\boldsymbol{\hat n}$ the outward unit normal.
The Dirichlet boundary condition is assumed to be compatible with the initial condition, i.e. $u_D(\boldsymbol{x}, 0) = u_0(\boldsymbol{x})$.

\par Since we will not be using Dirichlet boundary conditions, to simplify the discussion we will assume that the problem has pure Neumann boundary conditions, i.e. $\partial\Omega_N = \partial\Omega$, $\partial\Omega_D = \emptyset$.

\subsection{Boundary Fluxes}

\par The 3 fundamental forms of heat transport are conduction, convection and radiation.
Conduction occurs in the bulk of the material and is explicitly captured in the 3D heat equation, while convection and radiation terms appear only on the boundary of the domain.
In the simplest case, convection is modeled by Newton's cooling law.
Since both are fluxes, they are assigned to the (incoming) Neumann flux
\begin{equation}\label{eq:full_neumann_flux}
    \begin{aligned}
        F_N (u, \boldsymbol{x}, t) & =  h(\boldsymbol{x}) \left (u_\infty - u(\boldsymbol{x}, t) \right)                                  \\
                                   & + \epsilon(\boldsymbol{x}) \, \sigma \left(u_\infty^4 - u(\boldsymbol{x}, t)^4 \right)               \\
                                   & + \boldsymbol{I} \cdot \boldsymbol{\hat n}(\boldsymbol{x}) \, v(\boldsymbol{x}, \boldsymbol{\hat I})
    \end{aligned}
\end{equation}
with $h(\boldsymbol{x})$ being the spatially varying heat transfer coefficient [\si{\watt\per\square\meter\per\kelvin}], $\epsilon(\boldsymbol{x})$ the spatially varying emissivity, $\sigma$ is the Stefan-Boltzmann constant [\si{\watt\per\square\meter\per\kelvin\tothe{4}}], $\boldsymbol{I}$ is directional insolation [\si{\watt\per\square\meter}].
Finally, $v(\boldsymbol{x}, \boldsymbol{\hat I})$ is the visibility function that returns 1 if a ray with origin $\boldsymbol{x}$ and direction $\boldsymbol{\hat I}$ does not intersect the object, and returns 0 if it does.

\subsection{Weak Formulation}\label{appendix:heat_weak_formulation}

\par To find the solution within the framework of IGA, we begin by introducing the function space $V = H^1(\Omega)$.
Multiplying both sides by $\varphi \in V$, integrating both sides, and performing integration by parts, we obtain
\begin{equation}
    \begin{aligned}
         & \int_\Omega C(\boldsymbol{x}) \, \partial_t u(\boldsymbol{x}, t) \,\varphi(\boldsymbol{x}) \, dV +\int_\Omega  \nabla \varphi(\boldsymbol{x}) \cdot \left (\mathbf{K}(\boldsymbol{x}) \, \nabla u(\boldsymbol{x}, t) \right) \, dV \\
         & = \int_\Omega f(\boldsymbol{x}, t) \, \varphi(\boldsymbol{x}) \, dV + \int_{\partial \Omega} F_N(u, \boldsymbol{x}) \, \varphi(\boldsymbol{x}) \, dS
    \end{aligned}
\end{equation}

Our simulation domain is a thin shell formed by the reconstructed spline surface, which is a closed manifold $\mathcal{M}\subset\mathbb{R}^3$ consisting of 6 differentiable patches and represents the outer surface of the domain.
Let $\boldsymbol{y} \in \mathcal{M} \subset \mathbb{R}^3$ denote a point on the manifold.
Each $\boldsymbol{y} \in \mathcal{M}$ is then assigned a thickness, expressed as a positive, continuous, bounded function $d: \mathcal{M} \to (0, d_{\text{max}}]$.
Then, the interior surface can be represented as an offset surface along the unit normal $n(\boldsymbol{y})$ at a given $\boldsymbol{y} \in \mathcal{M}$.
The domain $\Omega$ can then be defined as
\begin{equation}
    \Omega = \left\{\boldsymbol{y} + \zeta \, \boldsymbol{\hat n}(\boldsymbol{y}) \mid
    \boldsymbol{y} \in \mathcal{M}, \; \zeta \in [-d(\boldsymbol{y}), 0]\right\}, \quad
    \boldsymbol{x}(\boldsymbol{y}, \zeta) = \boldsymbol{y} + \zeta \, \boldsymbol{\hat n}(\boldsymbol{y}).
\end{equation}

The thickness is assumed to be much smaller than the size of the manifold.
Additionally, it is necessary to assume that the thickness never exceeds the radius of curvature at any point on the manifold to ensure that the offset surface is self-intersection free.
The thin shells considered in this work satisfy this assumption comfortably, and the curvature regularization employed during reconstruction further discourages high-curvature regions.
Thus, if we write the solution as a Taylor series with respect to $\zeta$ at $\zeta = 0$
\begin{equation}
    \begin{aligned}
        u(\boldsymbol{x}, t) = u(\boldsymbol{y}, \zeta, t)
         & \approx u(\boldsymbol{y}, 0,  t) + \partial_\zeta u(\boldsymbol{y}, 0,  t) \, \zeta + \mathcal{O}(\zeta^2) \\
         & \approx u(\boldsymbol{y}, 0,  t) + \mathcal{O}(\zeta),
    \end{aligned}
\end{equation}
we obtain that neglecting thickness introduces an error proportional to $d_{\text{max}}$ along $\zeta$.
This justifies treating the 3D heat equation over a volume as a 2D heat equation over the manifold $\mathcal{M}$ if the thickness is sufficiently small compared to the size of the manifold.

More formally, we approximate integrals over $\Omega$ as
\begin{equation}
    \int_{\Omega} f(\boldsymbol{x})\,dV
    \approx \int_\mathcal{M} \int_{-d(\boldsymbol{y})}^{0} f(\boldsymbol{y}, \zeta)\,d\zeta dS
    \approx \,\int_\mathcal{M} d(\boldsymbol{y}) \, f(\boldsymbol{y})\,dS.
\end{equation}
Finally, we obtain the weak formulation of the heat equation on manifold $\mathcal{M}$
\begin{equation}\label{eq:weak_formulation_manifold}
    \begin{aligned}
          & \int_\mathcal{M} d(\boldsymbol{y}) \, C(\boldsymbol{y}) \, \partial_t u(\boldsymbol{y}, t) \,\varphi(\boldsymbol{y}) \, dS +
        \int_\mathcal{M} d(\boldsymbol{y}) \, \nabla \varphi(\boldsymbol{y}) \cdot \left (\mathbf{K}(\boldsymbol{y}) \, \nabla u(\boldsymbol{y}, t) \right) \, dS \\
        = & \int_\mathcal{M} d(\boldsymbol{y}) \, f(\boldsymbol{y}, t) \, \varphi(\boldsymbol{y}) \, dS +
        \int_\mathcal{M} F_N(u, \boldsymbol{y}) \, \varphi(\boldsymbol{y}) \, dS.
    \end{aligned}
\end{equation}
The Neumann boundary condition becomes an additional source term, while other terms gain an explicit dependence on thickness.
Note that it represents the sum of fluxes on the outer and inner surfaces.
For simplicity, we assume that the interior surface is insulating, i.e., has zero contribution to the flux.

\subsection{Galerkin Discretization}

It remains to discretize the vector space $V$ with an appropriate finite set of basis functions.
Using the standard Galerkin formulation, $V_h \subset V$ is introduced, a finite-dimensional subspace of dimension $N$ spanned by basis functions $\{\varphi_i(\boldsymbol{y})\}_{i=1}^N$.
The solution in the semi-discrete form can then be expressed as
\begin{equation}
    u_h(\boldsymbol{y}, t) = \sum_{i=1}^N c_i(t) \, \varphi_i  (\boldsymbol{y}),
\end{equation}
with $\{c_i(t)\}_{i=1}^N$ denoting the set of time-dependent coefficients $c_i: [0, T] \to \mathbb{R}$ that encode the solution.
Substituting this ansatz into the weak formulation \eqref{eq:weak_formulation_manifold}
\begin{equation}
    \begin{aligned}
        \sum_{i=1}^N & \left(\int_\mathcal{M} d(\boldsymbol{y}) C(\boldsymbol{y}) \, \varphi_i (\boldsymbol{y}) \varphi_h(\boldsymbol{y}) \, \partial_t c_i(t) \, dS
        + \int_\mathcal{M} d(\boldsymbol{y}) \nabla \varphi_h (\boldsymbol{y}) \cdot \left (\mathbf{K}(\boldsymbol{y}) \nabla \varphi_i(\boldsymbol{y}) \right) c_i(t) \, dS \right) \\
        =            & \int_\mathcal{M} d(\boldsymbol{y}) \, f(\boldsymbol{y}, t) \, \varphi_h(\boldsymbol{y}) \, dS
        + \int_\mathcal{M} F_N(u, \boldsymbol{y}) \, \varphi_h(\boldsymbol{y}) \, dS,
    \end{aligned}
\end{equation}
where $\varphi_h \in V_h$.
This equation has to hold for every $\varphi_h$, which introduces a set of $N$ equations, formed by substituting $\varphi_h(\boldsymbol{y}) = \varphi_j(\boldsymbol{y}), \forall j \in \{1, \dots, N\}$, for each $t \in [0, T]$.
We introduce the stiffness matrix $\mathbf{A} \in \mathbb{R}^{N \times N}$, the mass matrix $\mathbf{M} \in \mathbb{R}^{N \times N}$, the Neumann flux vector $\boldsymbol{f} \in \mathbb{R}^N$, the source term $\boldsymbol{s} \in \mathbb{R}^N$, and the solution vector $\boldsymbol{c}: [0, T] \to \mathbb{R}^N$.
Their entries are given by
\begin{equation}
    \begin{aligned}
        \mathbf{M}_{i,j}                 & = \int_\mathcal{M} d(\boldsymbol{y}) \, C(\boldsymbol{y}) \, \varphi_i(\boldsymbol{y}) \, \varphi_j(\boldsymbol{y}) \, dS,                             \\
        \mathbf{A}_{i, j}                & = \int_\mathcal{M} d(\boldsymbol{y}) \, \nabla \varphi_i(\boldsymbol{y}) \cdot (\mathbf{K}(\boldsymbol{y}) \, \nabla \varphi_j(\boldsymbol{y})) \, dS, \\
        \boldsymbol{s}_i(t)              & = \int_\mathcal{M} d(\boldsymbol{y}) \, f(\boldsymbol{y}, t) \, \varphi_i(\boldsymbol{y}) \, dS,                                                       \\
        \boldsymbol{f}_i(\boldsymbol{c}) & = \int_\mathcal{M} F_N \! \left(\sum_{j=1}^{N} c_j \, \varphi_j(\boldsymbol{y}), \, \boldsymbol{y}\right) \, \varphi_i(\boldsymbol{y}) \, dS.
    \end{aligned}
\end{equation}
The system can then simply be written as
\begin{equation}\label{eq:heat_semi_discrete}
    \mathbf{M} \, \partial_t\boldsymbol{c}(t) + \mathbf{A} \boldsymbol{c}(t) = \boldsymbol{s}(t) + \boldsymbol{f}(\boldsymbol{c}(t)).
\end{equation}

\subsection{Fully Discrete System}

Finally, the continuous time variable of the semi-discrete system \eqref{eq:heat_semi_discrete} has to be discretized.
Following standard practice, the time interval $[0,T]$ is partitioned into $N_t$ sub-intervals $[t_j, t_{j+1}], j \in \{0, \dots, N_t-1\}$, where the solution at the first time step $t_0$ is the initial temperature.
Observe that the function defining the initial temperature might not be represented exactly in the basis $\{\varphi_i \}_{i=1}^N$, but we assume that this is the case and that the coefficients at initial time, denoted $\boldsymbol{c}^{0}$, are available.
Integrating over an arbitrary sub-interval $[t_j, t_{j+1}]$ yields
\begin{equation}
    \mathbf{M} \left( \boldsymbol{c^{j+1}} - \boldsymbol{c^j} \right) +
    \mathbf{A} \int_{t_j}^{t_{j+1}}\boldsymbol{c}(t) \, dt
    = \int_{t_j}^{t_{j+1}}\boldsymbol{s}(t) \, dt + \int_{t_j}^{t_{j+1}}\boldsymbol{f}(\boldsymbol{c}(t)) \, dt.
\end{equation}

The choice of how the integral over $[t_j, t_{j+1}]$ is evaluated gives rise to different methods with different computational demands, stability conditions, and orders of accuracy.
The cheapest option is to use an explicit time stepping scheme, with the CFL stability condition dictating that the allowed step size $\Delta t_j = t_{j+1} - t_j$ is bounded from above by $c_0 h^2$, where $c_0$ is a constant that does not depend on the discretization and $h$ is the size of the smallest element.
Given the spatially varying material properties, the CFL condition can easily require very small step sizes, requiring an excessively large number of time steps.
Therefore, to avoid the possible pitfalls, we choose the Crank--Nicolson scheme that is unconditionally stable for any $\Delta t_j$.
Additionally, it preserves the total energy of the system up to arithmetic error.
It approximates the integral using the trapezoidal rule, introducing $\mathcal{O}(\Delta t_j^2)$ error that is multiplied by another $\Delta t_j$ originating from the time step, thus giving second-order accuracy at final time $T$.
\begin{equation}
    \begin{aligned}
         & \mathbf{M} \left( \boldsymbol{c^{j+1}} - \boldsymbol{c^j} \right) + \frac{1}{2} \Delta t_j \mathbf{A} \left (\boldsymbol{c}^{j+1} + \boldsymbol{c}^j \right) + \mathcal{O}(\Delta t_j^3)                                                                     \\
         & = \frac{1}{2} \Delta t_j \left( \boldsymbol{s}(t_{j+1}) + \boldsymbol{s}(t_j) \right) + \frac{1}{2} \Delta t_j \left( \boldsymbol{f} \left( \boldsymbol{c}^{j+1} \right) + \boldsymbol{f} \left(\boldsymbol{c}^j \right) \right) + \mathcal{O}(\Delta t_j^3)
    \end{aligned}
\end{equation}

Observe that to compute the coefficients at next time $\boldsymbol{c}^{j+1}$ it is necessary to solve a (possibly non-linear) system of equations
\begin{equation}
    \left(\mathbf{M} + \frac{\Delta t_j}{2}  \mathbf{A} \right) \boldsymbol{c}^{j+1}
    = \left( \mathbf{M} - \frac{\Delta t_j}{2}  \mathbf{A} \right) \boldsymbol{c}^j
    + \frac{1}{2} \Delta t_j \left(\boldsymbol{s}(t_{j+1}) + \boldsymbol{s}(t_j) + \boldsymbol{f} \left(\boldsymbol{c}^{j+1} \right) + \boldsymbol{f} \left(\boldsymbol{c}^j \right) \right)
\end{equation}

\subsection{Solving The System}
If the Neumann flux contains only the linear advection term, the system is linear and another matrix can be introduced to the left-hand side.
For the full flux from \eqref{eq:full_neumann_flux}, this is not the case, and we are left with a non-linear right-hand side vector.
Therefore, solving the system once is not sufficient to obtain a correct $\boldsymbol{c}^{j+1}$.

One option is using the Newton-Raphson method, but in this context it is simpler to use fixed-point iterations.
Instead of computing $\boldsymbol{c}^{j+1}$ directly, we start from an initial guess $^{(0)}\boldsymbol{c}^{j+1}$ and obtain an improved guess $^{(1)}\boldsymbol{c}^{j+1}$ by solving the system of equations.
A good initial guess is required, and the coefficients do not change considerably for reasonable step sizes $\Delta t_j$, making $\boldsymbol{c}^j$ a very good candidate.
Repeating this procedure yields a sequence $\{^{(k)}\boldsymbol{c}^{j+1}\}_{k=0}^{k_\text{max}}$, where $k_\text{max}$ is the number of fixed-point iterations.
Typically, it terminates when two successive estimates differ by less than a prescribed tolerance, i.e., $\norm{^{(k_\text{max})}\boldsymbol{c}^{j+1} - ^{(k_\text{max}-1)}\boldsymbol{c}^{j+1}} < \text{tol} \norm{^{(k_\text{max})}\boldsymbol{c}^{j+1}}$.
The last iterate is then taken as the correct solution $\boldsymbol{c}^{j+1} = ^{(k_\text{max})}\boldsymbol{c}^{j+1}$.
The tolerance should be at least 2 orders of magnitude lower than the expected accuracy with $\text{tol} = 10^{-7}$ being a reasonable value.

The system can be solved either with direct or iterative solvers.
The left-hand side (LHS) matrix $\mathbf{M} + \Delta t_j/2 \, \mathbf{A}$ is a sparse, symmetric positive-definite (SPD) matrix: $\mathbf{M}$ is sparse SPD, while $\mathbf{A}$ is sparse symmetric positive \emph{semi}-definite---with pure Neumann boundary conditions the constant functions lie in its kernel---so the sum inherits definiteness from $\mathbf{M}$.
Sparsity stems from the compact support of the B-spline basis functions.
Furthermore, the condition number of $\mathbf{M}$ is $\mathcal{O}(1)$ while the condition number of $\mathbf{A}$ scales as $\mathcal{O}(h^{-2})$.
Noting that for reasonable time steps $\Delta t \approx C h$, the overall condition number of the LHS matrix is approximately $\mathcal{O}(h^{-1})$.

The direct solver of choice is the sparse Cholesky factorization while conjugate gradients (CG) is the iterative solver of choice, taking an expected $\mathcal{O}(h^{-1/2})$ iterations per solve.
We found that the CHOLMOD-based \texttt{cholespy} Cholesky solver~\cite{nicoletLargeStepsInverse2021a, chenCholmod2008} is generally faster---the LHS matrix is constant across time steps, so it is factored once and only forward and backward substitutions are performed per step---and it is generally recommended for 2D problems due to the effective nested dissection algorithm that keeps fill-in minimal with reasonable factorization times.
The CG solver was also implemented with simple inverse lumped mass matrix preconditioning, but was found to be notably slower in the beginning and when the materials are more conductive, corresponding to larger entries of $\mathbf{K}$.

\subsection{IGA Tensor Assembly}

Finally, it remains to examine how the previously introduced matrices and vectors are computed, i.e., how the (non-linear) system is assembled.
The first step is to decompose the integration domain;
the assembly is performed for each patch separately, as each patch is the image of the parametric domain under its surface mapping.
This recasts the integrals into ones over a rectangular domain, $[0, 1]^2$ in particular, allowing the use of efficient tensor-product Gauss--Legendre quadrature (see \cref{appendix:numerical_quadrature}).

Considering the integral over a single patch, and denoting it again with $\mathcal{M}$ for simplicity, observe that $\boldsymbol{y} = S (\xi, \eta)$.
We define the following pullbacks to the parametric domain:

\begin{equation}
    \hat{u}_h(\xi,\eta) = \sum_{j=1}^{N} c_j \hat{\varphi}_j(\xi,\eta), \quad
    \hat{\varphi}_i(\xi,\eta) := \varphi_i(S(\xi,\eta)), \quad
    \hat{d}(\xi,\eta) := d(S(\xi,\eta)),
\end{equation}

\begin{equation}
    \hat{C}(\xi,\eta) := C(S(\xi,\eta)), \quad
    \hat{\mathbf{K}}(\xi,\eta) := \mathbf{K}(S(\xi,\eta)), \quad
    \hat{f}(\xi,\eta,t) := f(S(\xi,\eta),t).
\end{equation}
Recalling the differential geometry concepts (\cref{subsubsec_diff_geometry}) we have that the gradient of the basis functions is
\begin{equation}
    \nabla \varphi(\boldsymbol{y}) = \mathbf{F}(\xi, \eta) \, \mathbf{G}^{-1}(\xi, \eta) \,\hatnabla \hat{\varphi}(\xi, \eta),
    \qquad
    \hatnabla =
    \begin{bmatrix}
        \partial_\xi \\
        \partial_\eta
    \end{bmatrix}.
\end{equation}

The matrices and vectors are then given by
\begin{equation}
    \begin{aligned}
        \mathbf{M}_{i,j}                 & = \int_0^1  \int_0^1 \hat{d}(\xi,\eta) \, \hat{C}(\xi,\eta) \, \hat{\varphi}_i(\xi,\eta) \, \hat{\varphi}_j(\xi,\eta) \, g(\xi,\eta)\, d\xi \, d\eta,                                                                   \\
        \mathbf{A}_{i,j}                 & = \int_0^1 \int_0^1 \hat{d}(\xi,\eta)\, (\hatnabla \hat{\varphi}_i)^T \left( \mathbf{G}^{-1} \mathbf{F}^T \hat{\mathbf{K}} \mathbf{F} \mathbf{G}^{-1} \right) \hatnabla \hat{\varphi}_j \, g(\xi,\eta) \, d\xi\, d\eta, \\
        \boldsymbol{s}_i(t)              & = \int_0^1 \int_0^1 \hat{d}(\xi,\eta) \, \hat{f}(\xi,\eta,t) \,\hat{\varphi}_i(\xi,\eta) \, g(\xi,\eta) \, d\xi \, d\eta,                                                                                               \\
        \boldsymbol{f}_i(\boldsymbol{c}) & = \int_0^1 \int_0^1 F_N\!\left( \hat{u}_h(\xi,\eta),
        S(\xi,\eta)
        \right)
        \hat{\varphi}_i(\xi,\eta)\,
        g(\xi,\eta)\,
        d\xi\, d\eta
    \end{aligned}
\end{equation}

Finally, to simplify the stiffness matrix expression, the conductivity tensor is replaced with an isotropic conductivity, a scalar function $k: \mathcal{M} \to \mathbb{R}$, with an accompanying $\hat k(\xi, \eta) := k(S(\xi, \eta))$, reducing the stiffness matrix expression to
\begin{equation}
    \mathbf{A}_{i,j} = \int_0^1 \int_0^1 \hat{d}(\xi,\eta) \, \hat k(\xi, \eta) \, \hatnabla \hat{\varphi}_i (\xi, \eta) \cdot \left( \mathbf{G}^{-1}(\xi, \eta) \hatnabla \hat{\varphi}_j(\xi, \eta) \right) \, g(\xi,\eta) \, d\xi\, d\eta
\end{equation}

Each integral is decomposed over knot spans along the $\xi$ and $\eta$ directions, and taking the mass matrix entries as an example, we have
\begin{equation}
    \mathbf{M}_{i,j} = \sum_{k=1}^s \sum_{l=1}^s \int_{\xi_k^*}^{\xi_{k+1}^*}  \int_{\eta_l^*}^{\eta_{l+1}^*} \hat{d}(\xi,\eta) \, \hat{C}(\xi,\eta) \, \hat{\varphi}_i(\xi,\eta) \, \hat{\varphi}_j(\xi,\eta) \, g(\xi,\eta)\, d\xi \, d\eta,
\end{equation}
where $s$ is the number of knot spans in the knot vectors along both directions while $\{\xi_1^*, \allowbreak \dots, \allowbreak \xi_{s+1}^*\}$ and $\{\eta_1^*, \allowbreak \dots, \allowbreak \eta_{s+1}^*\}$ are the unique knots.
This arrangement allows for efficient integrations because the B-spline basis functions are polynomials inside of a given knot span.
Furthermore, only $(p+1)\times (q+1)$ basis functions are active in each knot span.
These basis functions are identified and a local mass matrix is assembled for each knot span.
Based on indices, the local matrices are then inserted into the global sparse matrix.
Additional care has to be taken when converting per-patch indices into global indices on the entire surface, to ensure correct coupling along the edges.
The same procedure applies to the stiffness matrix and RHS vectors.

It remains to numerically evaluate the integrals along both directions.
Since the thickness and heat capacity fields are represented in the same spline basis as the geometry (as produced by inpainting), mass matrix entries are the product of four polynomials of bi-degree $(p, q)$, meaning that the total bi-degree is $(4p, 4q)$.
Therefore, tensor product Gauss--Legendre quadrature of bi-order $(2p+1, 2q+1)$ should be applied to avoid under-integration.
The metric arises from the geometric description and is thus disregarded---it cannot introduce under-integration, only reduce the accuracy of the end result.
Observe that this is excessive on knot-spans with uniform material properties, as the bi-degree in such cases is only $(2p, 2q)$, reducing the amount of computation required by roughly a factor of $3$.
Similar reasoning applies to the other quantities of interest.
Since assembly time does not dominate overall execution, we keep the quadrature bi-order $(2p+1, 2q+1)$.

Taking the mass matrix as an example, a local (per knot-span) matrix can be computed as follows:
\begin{equation}
    \begin{aligned}
        \mathbf{M}^{\text{loc}}_{i', j'} = (\xi^*_{k+1} - \xi^*_k)(\eta^*_{l+1} - \eta^*_l) & \sum_{a=1}^{2p+1} \sum_{b=1}^{2q+1} \frac{\hat w_a}{2} \frac{\hat w_b}{2}                                                                             \\
                                                                                            & \times \hat{d}(\xi_a, \eta_b) \, \hat{C}(\xi_a, \eta_b) \, \hat{\varphi}_{i'}(\xi_a, \eta_b) \, \hat{\varphi}_{j'}(\xi_a, \eta_b) \, g(\xi_a, \eta_b)
    \end{aligned}
\end{equation}
where $\varphi_{i'}, \varphi_{j'} \in \mathcal{B}_{k, l}$, with $\mathcal{B}_{k, l}$ denoting the set of polynomial basis functions active in the parametric region $[\xi^*_k, \xi^*_{k+1}] \times [\eta^*_l, \eta^*_{l+1}]$, and $(\xi_a, \eta_b)$ the corresponding mapped Gauss--Legendre nodes.
The same expression is evaluated on each such rectangular region, making vectorized assembly, and thus leveraging GPU parallelism, very simple to achieve.
However, note that this does increase the memory footprint;
the intermediate array of local matrices has $s^2 (p+1)^2 (q+1)^2$ entries while the sparse matrix has approximately $(s+p)(s+q)$ rows with at most $(2p+1) (2q+1)$ entries per row.
For $p=q=3$, this is approximately $5$ times more storage, on top of the sparse matrix.
A possible mitigation is to batch the assembly, but we did not find it necessary for our problem sizes.

Note that the assembly procedure is fully differentiable as it uses the same data structures as Mitsuba \cite{Mitsuba3}.
This is only destroyed at the last step, as the assembled operators are exported to sparse CSR matrices from scipy \cite{scipy} and handed to the linear solver described above.

\section{Reissner-Mindlin Modal Analysis}\label{appendix:reissner_mindlin}

\subsection{Formal Statement}

\par Let $\Omega \subset \mathbb{R}^3$ be a closed three-dimensional body whose mechanical response we wish to model and let $\partial\Omega$ be its Lipschitz continuous boundary.
Furthermore, let the boundary be decomposed into two components $\partial\Omega = \partial\Omega_D \cup \partial\Omega_N$, $\partial\Omega_D \cap \partial\Omega_N = \emptyset$, where $\partial\Omega_D$ denotes the component where Dirichlet (prescribed displacement) conditions are imposed, and $\partial\Omega_N$ denotes the component with Neumann (prescribed traction) conditions.
Assuming the displacements are small, a linearized elasticity formulation can be applied.
Then, the dynamics are governed by the momentum equation, which is a linear, second-order partial differential equation:
\begin{equation}\label{eq:rm_3d_momentum}
    \begin{cases}
        \rho(\boldsymbol{x}) \, \partial_{tt} \boldsymbol{u}^{\text{3D}}(\boldsymbol{x}, t) - \nabla \cdot \boldsymbol{\sigma}(\boldsymbol{x}, t) = \boldsymbol{f}^{\text{3D}}(\boldsymbol{x}, t), & \boldsymbol{x} \in \Omega,           \\
        \boldsymbol{u}^{\text{3D}}(\boldsymbol{x}, t) = \boldsymbol{u}_D(\boldsymbol{x}, t),                                                                                                       & \boldsymbol{x} \in \partial\Omega_D, \\
        \boldsymbol{\sigma}(\boldsymbol{x}, t) \, \boldsymbol{\hat n}(\boldsymbol{x}) = \boldsymbol{t}(\boldsymbol{x}, t),                                                                         & \boldsymbol{x} \in \partial\Omega_N,
    \end{cases}
\end{equation}
where $\boldsymbol{u}^{\text{3D}}(\boldsymbol{x}, t)$ is the displacement vector, $\rho(\boldsymbol{x})$ is the density [\si{\kilogram\per\cubic\meter}], $\boldsymbol{f}^{\text{3D}}(\boldsymbol{x}, t)$ represents the body forces [\si{\newton\per\cubic\meter}], $\boldsymbol{\sigma} = \lambda \, \mathrm{tr}(\boldsymbol{\varepsilon}) \, \mathbf{I} + 2\mu \, \boldsymbol{\varepsilon}$ is the Cauchy stress tensor for a linear isotropic material with Lam\'{e} parameters $\lambda$, $\mu$ and linearized strain tensor $\boldsymbol{\varepsilon} = \frac{1}{2}(\nabla \boldsymbol{u}^{\text{3D}} + (\nabla \boldsymbol{u}^{\text{3D}})^T)$, $\boldsymbol{u}_D$ is the prescribed boundary displacement, $\boldsymbol{t}$ is the prescribed surface traction [\si{\newton\per\square\meter}], and $\boldsymbol{\hat n}$ is the outward unit normal to $\partial\Omega$.

Solving this problem directly on a thin three-dimensional domain is computationally expensive: the mesh must be able to resolve the thickness direction, introducing a large number of degrees of freedom and ill-conditioned systems.

\subsection{Dimensional Reduction}

\par Similarly to the dimensional reduction performed for the heat equation (\cref{appendix:heat_equation}), we describe the physical domain $\Omega$ as an offset from the closed mid-surface manifold $\mathcal{M} \subset \mathbb{R}^3$.
Each point $\boldsymbol{y} \in \mathcal{M}$ is assigned a thickness, expressed as a positive, continuous, and bounded function $d: \mathcal{M} \to (0, d_{\text{max}}]$.
The domain is then defined along the unit normal $\boldsymbol{\hat n}(\boldsymbol{y})$ as
\begin{equation}
    \Omega = \left\{\boldsymbol{y} + \zeta_{\boldsymbol{y}} \, \boldsymbol{\hat n}(\boldsymbol{y}) \;\middle|\;
    \boldsymbol{y} \in \mathcal{M}, \; \zeta_{\boldsymbol{y}} \in \left[-\tfrac{d(\boldsymbol{y})}{2}, \tfrac{d(\boldsymbol{y})}{2}\right]\right\}.
\end{equation}
Note that our method reconstructs the outer surface of the shell while the standard shell formulations use the mid-surface.
In principle this would require adapting the shell theory, but this is beyond the scope of this work, so we accept the approximation error, noting that the thickness is small compared to the size of the manifold.

Unlike the heat equation where a scalar temperature field is sought, elastodynamics requires solving for a 3D displacement field.
The two main formulations are Kirchhoff-Love shell theory and Reissner-Mindlin shell theory, with the key distinguishing factor being the regularity of the solution, the number of unknowns, and the allowed shell thickness.

Kirchhoff-Love theory is best suited to thin shells (thickness $\approx 0.01$ of shell dimension);
it requires $C^1$ continuity of the functional space, either by construction or through use of appropriate penalty terms \cite{guarinoKLInteriorPenalty2024}, and thus requires only 3 positional displacement variables per control point.
However, since our representation has only $C^0$ continuity along the shared edges, this formulation is inadmissible without more advanced treatment.
On the other hand, Reissner-Mindlin (RM) shell theory requires only $C^0$ continuity and is designed for shells of moderate thickness, which aligns better with our datasets.
It requires 3 positional and 3 rotational displacement variables per control point, making the resulting system much larger, as well as introducing a plethora of locking phenomena.

We choose the Reissner-Mindlin (RM) shell theory as it is compatible with $C^0$ continuity for the solution field as well as being applicable to shells of moderate thickness.
The RM kinematic assumption asserts that material fibers that were initially normal to the undeformed mid-surface remain straight after deformation, but are not required to remain normal: shearing is allowed.
The fiber length must be preserved (inextensibility).
Under these assumptions, the 3D displacement is approximated by a first-order expansion in the thickness coordinate $\zeta$:
\begin{equation}\label{eq:rm_through_thickness}
    \boldsymbol{u}^{\text{3D}}(\boldsymbol{y}, \zeta, t) \approx \boldsymbol{v}(\boldsymbol{y}, t) + \zeta \, \boldsymbol{w}(\boldsymbol{y}, t),
\end{equation}
where $\boldsymbol{v}(\boldsymbol{y}, t)$ is the displacement of the mid-surface and $\boldsymbol{w}(\boldsymbol{y}, t)$ is the change of the director (the rotation of the normal fiber).
Both are two-dimensional fields defined on $\mathcal{M}$ alone, and $\zeta$ enters only as a linear parameter.

The elastic energy of the three-dimensional body can then be integrated analytically through the thickness.
Substituting \cref{eq:rm_through_thickness} into the 3D strain tensor and integrating $\int_{-d/2}^{d/2} (\cdot) \, d\zeta$, the terms independent of $\zeta$ produce the \emph{membrane energy} (in-plane stretching, scaled by $d$), the terms linear in $\zeta$ vanish by symmetry, and the terms quadratic in $\zeta$ produce the \emph{bending energy} (curvature change, scaled by $d^3/12$).
The through-thickness shear, which is absent in the thinner Kirchhoff-Love theory, produces the \emph{transverse shear energy} (scaled by $d$, with a correction factor $\kappa_s = 5/6$).
Similarly, the kinetic energy integrates to a translational inertia term (scaled by $\rho d$) and a rotary inertia term (scaled by $\rho d^3/12$).
This reduction yields a two-dimensional variational problem on $\mathcal{M}$ with two independent fields: the displacement $\boldsymbol{v}$ and the director change $\boldsymbol{w}$.
The full derivation is lengthy and technical, and we refer the interested reader to \cite{guarinoRMStabilization2025} for a complete derivation in the Appendices.

For modal analysis, we set $\boldsymbol{f}^{\text{3D}} = \boldsymbol{0}$ and seek harmonic solutions $\boldsymbol{v}(\boldsymbol{y}, t) = \boldsymbol{v}(\boldsymbol{y}) e^{i\omega t}$, reducing the problem to the generalized eigenvalue problem
\begin{equation}\label{eq:rm_eigenvalue}
    \mathbf{K} \boldsymbol{\phi} = \lambda \, \mathbf{M} \boldsymbol{\phi},
\end{equation}
where $\mathbf{K}$ is the shell stiffness matrix, $\mathbf{M}$ is the mass matrix, $\lambda = \omega^2$ are the squared angular natural frequencies ($\omega = 2\pi f$), and $\boldsymbol{\phi}$ are the mode shapes.
For a free-free shell, i.e., with no Dirichlet or Neumann boundary conditions, the first six eigenvalues are zero, corresponding to three translational and three rotational rigid body modes.

\subsection{Tensor Notation}

\par The director change $\boldsymbol{w}$ is parametrized by a rotation vector $\boldsymbol{\omega}: \mathcal{M} \to \mathbb{R}^3$, expressed in global Cartesian coordinates:
\begin{equation}
    \boldsymbol{w} = \boldsymbol{\omega} \times \boldsymbol{\hat n}.
\end{equation}
The cross product automatically satisfies the inextensibility constraint $\boldsymbol{w} \cdot \boldsymbol{\hat n} = 0$, ensuring that the director remains a unit vector to first order.

The per-patch mid-surface is parametrized by a B-spline $\boldsymbol{y} = \boldsymbol{S}(\xi, \eta)$, with $(\xi, \eta)$ being the parametric coordinates.
We adopt the tensor notation in favor of matrix notation used for the heat equation as it is standard in the field and simplifies the expressions.
The covariant tangent vectors, unit normal, and normal derivatives are
\begin{equation}
    \boldsymbol{a}_1 = \frac{\partial \boldsymbol{S}}{\partial \xi}, \quad
    \boldsymbol{a}_2 = \frac{\partial \boldsymbol{S}}{\partial \eta}, \quad
    \boldsymbol{\hat n} = \frac{\boldsymbol{a}_1 \times \boldsymbol{a}_2}{|\boldsymbol{a}_1 \times \boldsymbol{a}_2|}, \quad
    \boldsymbol{\hat n}_{,\alpha} = \frac{\partial \boldsymbol{\hat n}}{\partial \xi^\alpha}.
\end{equation}
The metric tensor and its inverse are
\begin{equation}
    g_{\alpha\beta} = \boldsymbol{a}_\alpha \cdot \boldsymbol{a}_\beta, \qquad
    g^{\alpha\beta} = \left[g_{\alpha\beta}\right]^{-1}, \qquad
    g = \det(g_{\alpha\beta}).
\end{equation}

The parametric derivative of the director change is
\begin{equation}
    \boldsymbol{w}_{,\alpha} = \boldsymbol{\omega}_{,\alpha} \times \boldsymbol{\hat n} + \boldsymbol{\omega} \times \boldsymbol{\hat n}_{,\alpha}.
\end{equation}
Since $\boldsymbol{\omega}$ is expressed in global Cartesian coordinates (constant basis vectors), $\boldsymbol{\omega}_{,\alpha}$ is simply the parametric derivative of the rotation coefficients---no Christoffel symbols appear.
The curvature of the surface enters through the cross products with $\boldsymbol{\hat n}$ and $\boldsymbol{\hat n}_{,\alpha}$.

\subsection{Strain Measures}

\par The through-thickness integration of the 3D strain tensor yields three covariant strain measures on the mid-surface.
The \emph{membrane strain} captures in-plane stretching (from the terms independent of $\zeta$):
\begin{equation}\label{eq:rm_membrane_strain}
    \varepsilon_{\alpha\beta} = \frac{1}{2}\left(\boldsymbol{a}_\alpha \cdot \boldsymbol{v}_{,\beta} + \boldsymbol{a}_\beta \cdot \boldsymbol{v}_{,\alpha}\right).
\end{equation}

The \emph{bending strain} (curvature change, from the terms linear in $\zeta$) involves both the displacement and rotation fields:
\begin{equation}\label{eq:rm_bending_strain}
    \kappa_{\alpha\beta} = \frac{1}{2}\left(\boldsymbol{a}_\alpha \cdot \boldsymbol{w}_{,\beta} + \boldsymbol{a}_\beta \cdot \boldsymbol{w}_{,\alpha} + \boldsymbol{\hat n}_{,\alpha} \cdot \boldsymbol{v}_{,\beta} + \boldsymbol{\hat n}_{,\beta} \cdot \boldsymbol{v}_{,\alpha}\right).
\end{equation}
The terms involving $\boldsymbol{\hat n}_{,\alpha}$ couple the displacement to the curvature and are essential for curved shells; they vanish on flat plates where $\boldsymbol{\hat n}$ is constant.

The \emph{transverse shear strain} measures the deviation of the deformed director from the deformed surface normal:
\begin{equation}\label{eq:rm_shear_strain}
    \gamma_\alpha = \boldsymbol{a}_\alpha \cdot \boldsymbol{w} + \boldsymbol{\hat n} \cdot \boldsymbol{v}_{,\alpha} = \boldsymbol{\omega} \cdot \boldsymbol{L}_\alpha + \boldsymbol{\hat n} \cdot \boldsymbol{v}_{,\alpha},
\end{equation}
where $\boldsymbol{L}_\alpha = \boldsymbol{\hat n} \times \boldsymbol{a}_\alpha$ are the \emph{bending lever} vectors that convert rotation into transverse shear.

\subsection{Constitutive Relations}

\par For an isotropic linear elastic material with Young's modulus $E$, Poisson ratio $\nu$, and thickness $d$, the constitutive law is expressed in covariant components using plane-stress modified Lam\'e parameters:
\begin{equation}
    \mu = \frac{E}{2(1+\nu)}, \qquad
    \lambda = \frac{E\nu}{(1+\nu)(1-2\nu)}, \qquad
    \bar\lambda = \frac{2\lambda\mu}{\lambda + 2\mu}.
\end{equation}
The modification $\bar\lambda$ converts the three-dimensional elasticity tensor to plane stress.

The elastic energy density for a covariant strain tensor $e_{\alpha\beta}$ with thickness scale factor $s$ is
\begin{equation}\label{eq:rm_elastic_energy}
    W(e) = \frac{s}{2} \left(\bar\lambda \left(g^{\alpha\beta} e_{\alpha\beta}\right)^2 + 2\mu \, g^{\alpha\gamma} e_{\gamma\delta} \, g^{\delta\beta} e_{\beta\alpha}\right),
\end{equation}
where $g^{\alpha\beta}$ is the contravariant metric tensor used for index raising.
For membrane strains $s = d$, while for bending ones $s = d^3/12$.

The transverse shear energy density is
\begin{equation}
    W_{\text{shear}} = \frac{1}{2} \, \kappa_s \, \mu \, d \, g^{\alpha\beta} \gamma_\alpha \gamma_\beta,
\end{equation}
with $\kappa_s = 5/6$ being the Timoshenko shear correction factor.

The membrane and bending energy densities share the plane-stress elasticity tensor in contravariant form:
\begin{equation}\label{eq:rm_elasticity_tensor}
    \mathcal{D}^{\alpha\beta\gamma\delta} = \bar\lambda \, g^{\alpha\beta} g^{\gamma\delta} + \mu \left(g^{\alpha\gamma} g^{\beta\delta} + g^{\alpha\delta} g^{\beta\gamma}\right).
\end{equation}
Differentiating the energy densities with respect to the strains defines the membrane force $N^{\alpha\beta}$, bending moment $M^{\alpha\beta}$, and transverse shear force $Q^\alpha$:
\begin{subequations}\label{eq:rm_stress_resultants}
    \begin{align}
        N^{\alpha\beta} & = d \, \mathcal{D}^{\alpha\beta\gamma\delta} \, \varepsilon_{\gamma\delta}, \label{eq:rm_membrane_force}         \\
        M^{\alpha\beta} & = \frac{d^3}{12} \, \mathcal{D}^{\alpha\beta\gamma\delta} \, \kappa_{\gamma\delta}, \label{eq:rm_bending_moment} \\
        Q^\alpha        & = \kappa_s \, \mu \, d \, g^{\alpha\beta} \, \gamma_\beta. \label{eq:rm_shear_force}
    \end{align}
\end{subequations}

\subsection{Drilling Stabilization}

\par The rotation component along the surface normal, $\boldsymbol{\omega} \cdot \boldsymbol{\hat n}$, has no physical stiffness in the Reissner-Mindlin theory;
in fact it is assumed to be 0.
Thus, to avoid a non-physical rotation field, an \emph{objective drilling penalty} is introduced that ties this rotation to the physical in-plane rotation extracted from the displacement gradient,
\begin{equation}
    \omega_{\text{phys}}
    = \tfrac{1}{2}\,\boldsymbol{\hat n}\cdot\left(\boldsymbol{a}^{\alpha}\times\boldsymbol{v}_{,\alpha}\right),
\end{equation}
where $\boldsymbol{a}^{\alpha}=g^{\alpha\beta}\boldsymbol{a}_{\beta}$ are the contravariant tangent vectors.
Equivalently,
\begin{equation}
    \omega_{\text{phys}}
    = \tfrac{1}{2}\left[
        \left(\boldsymbol{\hat n}\times\boldsymbol{a}^{1}\right)\cdot\boldsymbol{v}_{,1}
        +
        \left(\boldsymbol{\hat n}\times\boldsymbol{a}^{2}\right)\cdot\boldsymbol{v}_{,2}
        \right].
\end{equation}

The drilling penalty energy density is then taken as
\begin{equation}
    W_{\text{drill}} = \frac{\alpha_{\text{drill}}}{2} \left(\boldsymbol{\omega} \cdot \boldsymbol{\hat n} - \omega_{\text{phys}}\right)^2,
\end{equation}
with $\alpha_{\text{drill}} = c_{\text{d}} \, \mu \, d$ for a dimensionless factor $c_{\text{d}} \sim \mathcal{O}(1)$; we use $c_{\text{d}} = 1$ throughout.
This formulation vanishes for rigid body motions, ensuring that the six zero eigenvalues of a free-free shell are preserved exactly.

\subsection{Weak Formulation}\label{subsec:rm_weak_formulation}

\par Let $\boldsymbol{u} = (\boldsymbol{v}, \boldsymbol{\omega})$ denote the generalized displacement, combining the mid-surface displacement and the rotation vector, and let $V = [H^1(\mathcal{M})]^6$ be the appropriate function space, reflecting three displacement and three rotation components in global Cartesian coordinates.
The total potential energy of the shell is
\begin{equation}\label{eq:rm_total_energy}
    \Pi(\boldsymbol{u}) = \int_\mathcal{M} \left(W(\varepsilon; d) + W(\kappa; d^3/12) + W_{\text{shear}} + W_{\text{drill}}\right) \sqrt{g} \, d\xi \, d\eta.
\end{equation}
For the free-vibration problem ($\boldsymbol{f}^{\text{3D}} = \boldsymbol{0}$), the variational statement reduces to the eigenvalue problem: find $(\lambda, \boldsymbol{u}) \in \mathbb{R} \times V$, $\boldsymbol{u} \neq \boldsymbol{0}$, such that
\begin{equation}\label{eq:rm_weak_form}
    a(\delta\boldsymbol{u}, \boldsymbol{u}) = \lambda \, m(\delta\boldsymbol{u}, \boldsymbol{u}), \qquad \forall \, \delta\boldsymbol{u} \in V,
\end{equation}
where $a: V \times V \to \mathbb{R}$ is the stiffness bilinear form and $m: V \times V \to \mathbb{R}$ is the mass bilinear form.

The stiffness bilinear form is obtained as the second variation of $\Pi$ and decomposes into membrane, bending, shear, and drilling contributions:
\begin{equation}\label{eq:rm_stiffness_decomposition}
    a(\delta\boldsymbol{u}, \boldsymbol{u}) = a_{\text{mem}} + a_{\text{bend}} + a_{\text{shear}} + a_{\text{drill}}.
\end{equation}
Using the stress resultants \cref{eq:rm_stress_resultants}, the physical contributions take the standard virtual work form:
\begin{equation}\label{eq:rm_internal_work}
    a_{\text{mem}} + a_{\text{bend}} + a_{\text{shear}} = \int_\mathcal{M} \left(\delta\varepsilon_{\alpha\beta} \, N^{\alpha\beta} + \delta\kappa_{\alpha\beta} \, M^{\alpha\beta} + \delta\gamma_\alpha \, Q^\alpha \right) \sqrt{g} \, d\xi \, d\eta,
\end{equation}
where $\delta\varepsilon_{\alpha\beta}$, $\delta\kappa_{\alpha\beta}$, $\delta\gamma_\alpha$ are the strain measures \cref{eq:rm_membrane_strain}--\cref{eq:rm_shear_strain} evaluated at the test function $\delta\boldsymbol{u}$.
Substituting the constitutive relations \cref{eq:rm_stress_resultants} yields the explicit bilinear forms:
\begin{subequations}\label{eq:rm_bilinear_forms}
    \begin{align}
        a_{\text{mem}}(\delta\boldsymbol{u}, \boldsymbol{u})   & = \int_\mathcal{M} d \, \mathcal{D}^{\alpha\beta\gamma\delta} \, \delta\varepsilon_{\alpha\beta} \, \varepsilon_{\gamma\delta} \, \sqrt{g} \, d\xi \, d\eta, \label{eq:rm_a_mem}     \\
        a_{\text{bend}}(\delta\boldsymbol{u}, \boldsymbol{u})  & = \int_\mathcal{M} \frac{d^3}{12} \, \mathcal{D}^{\alpha\beta\gamma\delta} \, \delta\kappa_{\alpha\beta} \, \kappa_{\gamma\delta} \, \sqrt{g} \, d\xi \, d\eta, \label{eq:rm_a_bend} \\
        a_{\text{shear}}(\delta\boldsymbol{u}, \boldsymbol{u}) & = \int_\mathcal{M} \kappa_s \, \mu \, d \, g^{\alpha\beta} \, \delta\gamma_\alpha \, \gamma_\beta \, \sqrt{g} \, d\xi \, d\eta. \label{eq:rm_a_shear}
    \end{align}
\end{subequations}
The drilling stabilization contributes
\begin{equation}\label{eq:rm_a_drill}
    a_{\text{drill}}(\delta\boldsymbol{u}, \boldsymbol{u}) = \int_\mathcal{M} \alpha_{\text{drill}} \left(\delta\boldsymbol{\omega} \cdot \boldsymbol{\hat n} - \delta\omega_{\text{phys}}\right) \left(\boldsymbol{\omega} \cdot \boldsymbol{\hat n} - \omega_{\text{phys}}\right) \sqrt{g} \, d\xi \, d\eta,
\end{equation}
where $\delta\omega_{\text{phys}}$ and $\omega_{\text{phys}}$ denote the physical in-plane rotation extracted from the displacement gradient of $\delta\boldsymbol{u}$ and $\boldsymbol{u}$, respectively.
The mass bilinear form $m$ is derived in the following section.
For a free-free shell, the stiffness form $a$ has a six-dimensional kernel corresponding to three translational and three rotational rigid body modes, yielding six zero eigenvalues in \cref{eq:rm_weak_form}.

\subsection{Mass Matrix}

\par The kinetic energy of the shell is obtained by integrating the 3D kinetic energy density through the thickness using the Reissner-Mindlin ansatz \cref{eq:rm_through_thickness}:
\begin{equation}\label{eq:rm_kinetic_energy}
    T = \frac{1}{2}\int_\mathcal{M} \left(\rho \, d \, |\dot{\boldsymbol{v}}|^2 + \frac{\rho \, d^3}{12} |\dot{\boldsymbol{w}}|^2 \right) \sqrt{g}\, d\xi\, d\eta.
\end{equation}
The first term is the translational inertia and the second is the rotary inertia.
Since $\boldsymbol{w} = \boldsymbol{\omega} \times \boldsymbol{\hat n}$, the rotary inertia can be rewritten as $|\dot{\boldsymbol{w}}|^2 = |\dot{\boldsymbol{\omega}}|^2 - (\dot{\boldsymbol{\omega}} \cdot \boldsymbol{\hat n})^2$, yielding the mass bilinear form
\begin{equation}\label{eq:rm_mass_bilinear}
    m(\delta\boldsymbol{u}, \boldsymbol{u}) = \int_\mathcal{M} \rho \, d \, \delta\boldsymbol{v} \cdot \boldsymbol{v} \, \sqrt{g} \, d\xi \, d\eta + \int_\mathcal{M} \frac{\rho \, d^3}{12} \left(\delta\boldsymbol{\omega} \cdot \boldsymbol{\omega} - (\delta\boldsymbol{\omega} \cdot \boldsymbol{\hat n})(\boldsymbol{\omega} \cdot \boldsymbol{\hat n})\right) \sqrt{g} \, d\xi \, d\eta.
\end{equation}
The projection $\delta_{ij} - \hat n_i \hat n_j$ in the rotary inertia assigns zero mass to the drilling rotation, which is physically correct since this component does not correspond to any through-thickness motion.
The Galerkin discretization of \cref{eq:rm_mass_bilinear} gives the mass matrix entries:
\begin{equation}
    M^{IJ}_{ij} =
    \begin{cases}
        \displaystyle\int_\mathcal{M} \rho \, d \, N_I \, N_J \, \delta_{ij} \, \sqrt{g} \, d\xi\, d\eta,                                    & i, j \in \{v_x, v_y, v_z\},                \\[8pt]
        \displaystyle\int_\mathcal{M} \frac{\rho \, d^3}{12} \, N_I \, N_J \, (\delta_{ij} - \hat n_i \hat n_j) \, \sqrt{g} \, d\xi\, d\eta, & i, j \in \{\omega_x, \omega_y, \omega_z\},
    \end{cases}
\end{equation}
where $N_I, N_J$ are basis functions and $\delta_{ij}$ is the Kronecker delta.
There are no displacement-rotation cross terms, since the kinetic energy \cref{eq:rm_kinetic_energy} decouples into translational and rotary contributions.

\subsection{Galerkin Discretization}

\par The displacement and rotation fields are approximated using the same B-spline basis functions that define the geometry (isogeometric analysis):
\begin{equation}
    \boldsymbol{v}_h(\xi, \eta) = \sum_{I=1}^N N_I(\xi, \eta) \, \boldsymbol{v}_I, \qquad
    \boldsymbol{\omega}_h(\xi, \eta) = \sum_{I=1}^N N_I(\xi, \eta) \, \boldsymbol{\omega}_I,
\end{equation}
with 6 degrees of freedom per control point: $[v_x, v_y, v_z, \omega_x, \omega_y, \omega_z]_I$.
Since all degrees of freedom are expressed in global Cartesian coordinates and share control points at patch boundaries, $C^0$ continuity of both displacement and rotation is automatically enforced without interface coupling terms.

For each basis function $I$ and Cartesian component $k \in \{x, y, z\}$, the strain contributions from displacement DOF $(I, k)$ are:
\begin{equation}
    \varepsilon_{\alpha\beta}^{I,k} = \frac{1}{2}\left(a_{\alpha,k} \frac{\partial N_I}{\partial \xi^\beta} + a_{\beta,k} \frac{\partial N_I}{\partial \xi^\alpha}\right), \quad
    \gamma_\alpha^{I,k} = \hat n_k \frac{\partial N_I}{\partial \xi^\alpha},
\end{equation}
\begin{equation}
    \kappa_{\alpha\beta}^{I,k} = \frac{1}{2}\left(\hat n_{\alpha,k} \frac{\partial N_I}{\partial \xi^\beta} + \hat n_{\beta,k} \frac{\partial N_I}{\partial \xi^\alpha}\right),
\end{equation}
where $a_{\alpha,k}$, $\hat n_k$, $\hat n_{\alpha,k}$ denote the $k$-th Cartesian component of $\boldsymbol{a}_\alpha$, $\boldsymbol{\hat n}$, $\boldsymbol{\hat n}_{,\alpha}$ respectively.

The strain contributions from rotation DOF $(I, k)$ are:
\begin{equation}
    \kappa_{\alpha\beta}^{I,3+k} = \frac{1}{2}\left(L_{\alpha,k} \frac{\partial N_I}{\partial \xi^\beta} + L_{\beta,k} \frac{\partial N_I}{\partial \xi^\alpha}\right)
    + \frac{1}{2}\left[(\boldsymbol{\hat n}_{,\beta} \times \boldsymbol{a}_\alpha)_k + (\boldsymbol{\hat n}_{,\alpha} \times \boldsymbol{a}_\beta)_k\right] N_I,
\end{equation}
\begin{equation}
    \gamma_\alpha^{I,3+k} = L_{\alpha,k} \, N_I,
\end{equation}
where $L_{\alpha,k}$ is the $k$-th component of $\boldsymbol{L}_\alpha = \boldsymbol{\hat n} \times \boldsymbol{a}_\alpha$.
The first term in $\kappa_{\alpha\beta}^{I,3+k}$ arises from $\boldsymbol{\omega}_{,\alpha} \times \boldsymbol{\hat n}$ (rotation gradient), while the second arises from $\boldsymbol{\omega} \times \boldsymbol{\hat n}_{,\alpha}$ (curvature-rotation coupling, present only on curved surfaces).

The drilling contribution requires a per-DOF drilling strain operator, obtained by expanding the drilling penalty for each DOF type:
\begin{equation}
    \psi^{I,k} = -\frac{1}{2}\left[\left(\boldsymbol{\hat n} \times \boldsymbol{a}^1\right)_k \frac{\partial N_I}{\partial \xi^1} + \left(\boldsymbol{\hat n} \times \boldsymbol{a}^2\right)_k \frac{\partial N_I}{\partial \xi^2}\right], \qquad
    \psi^{I,3+k} = \hat n_k \, N_I,
\end{equation}
where $(\boldsymbol{\hat n} \times \boldsymbol{a}^\alpha)_k$ denotes the $k$-th Cartesian component of the rotated contravariant tangent vector, with $\boldsymbol{a}^\alpha = g^{\alpha\beta}\boldsymbol{a}_\beta$.

Substituting the discrete fields into the bilinear forms \cref{eq:rm_bilinear_forms}--\cref{eq:rm_a_drill}, the stiffness matrix entry for DOFs $(I, m)$ and $(J, n)$, with $m, n \in \{1, \ldots, 6\}$, is
\begin{equation}\label{eq:rm_stiffness_entry}
    \begin{aligned}
        K_{(I,m),(J,n)} = \int_\mathcal{M} \bigg( & d \, \mathcal{D}^{\alpha\beta\gamma\delta} \, \varepsilon_{\alpha\beta}^{I,m} \, \varepsilon_{\gamma\delta}^{J,n}
        + \frac{d^3}{12} \, \mathcal{D}^{\alpha\beta\gamma\delta} \, \kappa_{\alpha\beta}^{I,m} \, \kappa_{\gamma\delta}^{J,n}                                        \\
                                                  & + \kappa_s \, \mu \, d \, g^{\alpha\beta} \, \gamma_\alpha^{I,m} \, \gamma_\beta^{J,n}
        + \alpha_{\text{drill}} \, \psi^{I,m} \, \psi^{J,n} \bigg) \sqrt{g} \, d\xi \, d\eta,
    \end{aligned}
\end{equation}
where all strain arrays carry the composite DOF index as defined above, with $\varepsilon_{\alpha\beta}^{I,m} = 0$ for rotation DOFs ($m > 3$).
This expression, together with the mass matrix entries, yields the discrete generalized eigenvalue problem $\mathbf{K}\boldsymbol{\phi} = \lambda\,\mathbf{M}\boldsymbol{\phi}$ from \cref{eq:rm_eigenvalue}.

\subsection{IGA Assembly}

The assembly follows the standard IGA procedure described for the heat equation (\cref{appendix:heat_equation}).
Each patch integral in \cref{eq:rm_stiffness_entry} is decomposed over knot spans, and local stiffness and mass matrices are assembled using tensor-product Gauss--Legendre quadrature with $2p + 1$ points per direction, where $p$ is the spline degree.
This rule is finer than the $p + 1$ points that would suffice for polynomial integrands because the integrands on a curved surface are rational: the inverse metric, the normal derivatives, and the area factor $\sqrt{g}$ all introduce non-polynomial dependence.
At each quadrature point, the per-DOF strain arrays $\varepsilon_{\alpha\beta}^{I,m}$, $\kappa_{\alpha\beta}^{I,m}$, $\gamma_\alpha^{I,m}$, and $\psi^{I,m}$ are evaluated for all active basis functions, and the integrand of \cref{eq:rm_stiffness_entry} is accumulated into the local matrix.
Local matrices are mapped to global DOF indices via the composite spline surface mapping and assembled into a global sparse matrix.

\par All four stiffness contributions---membrane, bending, shear, and drilling---are integrated with this same full quadrature rule; no reduced or selective integration and no assumed-strain treatment is applied.
Within each patch, the $C^{p-1}$ continuity of the spline basis substantially mitigates the shear locking that afflicts low-order $C^0$ elements, so the fully integrated cubic ($p = 3$) basis is used throughout, with the objective drilling stabilization ($c_{\text{d}} = 1$) active.
The residual membrane locking that can arise on very thin shells is controlled by knot refinement rather than by any modified integration or projection scheme.

\subsection{Eigenvalue Solve}

\par The generalized eigenvalue problem \cref{eq:rm_eigenvalue} is solved using a shift-invert Lanczos iteration \cite{scipy} to extract the lowest eigenvalues and corresponding mode shapes.
Since the stiffness matrix $\mathbf{K}$ is singular (its kernel consists of the six rigid body modes), a mass-proportional spectral shift is applied:
\begin{equation}
    \mathbf{K}_{\text{reg}} = \mathbf{K} + \epsilon \, \mathbf{M},
\end{equation}
shifting all the eigenvalues by $\epsilon$ without affecting the eigenvectors.
Note that the mass matrix is also rank deficient, with a null-space of dimension approximately one sixth of all degrees of freedom, due to only 2 physical rotations being admissible at a particular point on the surface.
Yet, the regularized matrix $\mathbf{K}_{\text{reg}}$ is symmetric positive definite, allowing a sparse Cholesky factorization (via \texttt{scikit-sparse}~\cite{scikit_sparse_0_5_0}'s CHOLMOD~\cite{chenCholmod2008}) to serve as the shift-invert operator in the \texttt{scipy}~\cite{scipy}'s shift-invert Lanczos eigensolver.
After solving, $\epsilon$ is subtracted from the computed eigenvalues to recover the original spectrum.

\subsection{Finite Element Reference Solver}\label{subsec:rm_fem_reference}

\par The FEM reference spectra are computed with an independent implementation of the same Reissner-Mindlin formulation in DOLFINx~\cite{barattaDOLFINx2023}, operating on triangle meshes: the faceted ground-truth meshes for the end-to-end validation (\cref{appendix:e2e}), and curved second-order tessellations of the reconstructed spline for the solver-agreement study.
The 6-DOF layout of the IGA solver is retained: displacement and rotation are global Cartesian vector fields, both discretized with continuous Lagrange elements of degree 3.
The strain measures are expressed coordinate-free through the surface normal $\boldsymbol{\hat n}$ and the tangential projector $\mathbf{I} - \boldsymbol{\hat n} \boldsymbol{\hat n}^T$, and are constructed to vanish exactly on rigid body motions, preserving the six zero modes even on faceted geometry.
The plane-stress constitutive relations, the shear correction factor $\kappa_s = 5/6$, the objective drilling penalty (with a per-element coefficient $E \, d^3 / h_e^2$, where $h_e$ is the local element size), and the projected rotary inertia all mirror the IGA solver.

\par Obtaining a locking-free reference spectrum with $C^0$ elements requires a more aggressive treatment than in the IGA case.
Following \citet{huReissnerMindlinBbar2020}, the reference applies an element-local assumed-strain ($\bar{B}$) projection.
Let $\{K\}$ be the elements of the triangulation and, for degree-$p$ displacement and rotation fields, let $W_h$ be the space of tensor fields that are polynomials of one degree lower on each element and discontinuous across element boundaries, $W_h|_K = \mathbb{P}_{p-1}(K)$.
The element-local $L^2$ projection $\Pi$ onto $W_h$ acts componentwise: for a strain field $s$, $\bar s = \Pi s \in W_h$ is defined on each element $K$ independently by
\begin{equation}\label{eq:rm_bbar_projection}
    \int_K \left(\bar s - s\right) q \, dS = 0 \qquad \forall\, q \in W_h|_K .
\end{equation}
The projection is applied to \emph{both} the membrane and the transverse-shear strains, replacing them by the assumed strains
\begin{equation}\label{eq:rm_bbar_strains}
    \bar\varepsilon_{\alpha\beta} = \Pi\, \varepsilon_{\alpha\beta}, \qquad
    \bar\gamma_\alpha = \Pi\, \gamma_\alpha,
\end{equation}
while the bending strain $\kappa_{\alpha\beta}$ and the drilling term are left compatible and fully integrated.
The stiffness bilinear form \cref{eq:rm_stiffness_decomposition} is thus replaced by
\begin{equation}\label{eq:rm_bbar_stiffness}
    \bar a(\delta\boldsymbol{u}, \boldsymbol{u}) = \bar a_{\text{mem}} + a_{\text{bend}} + \bar a_{\text{shear}} + a_{\text{drill}},
\end{equation}
where $\bar a_{\text{mem}}$ and $\bar a_{\text{shear}}$ are the membrane and shear forms \cref{eq:rm_a_mem} and \cref{eq:rm_a_shear} with the compatible strains $\varepsilon_{\alpha\beta}$, $\gamma_\alpha$ (in both the trial and the test slot) replaced by their assumed counterparts \cref{eq:rm_bbar_strains}, and $a_{\text{bend}}$, $a_{\text{drill}}$ are unchanged.
Because the same orthogonal projector acts on the test and trial strains, the projected stiffness is symmetric positive semi-definite; the six-dimensional rigid-body kernel is preserved exactly, since the compatible membrane and shear strains already vanish on rigid-body motions and hence so do their projections.
The displacement and rotation fields are cubic ($p = 3$) Lagrange elements, so $W_h$ is the discontinuous quadratic space: on a thin flat-plate benchmark, only this assumed-strain configuration yielded a clean, locking-free spectrum, whereas standard displacement elements and the unprojected and reduced-integration variants we tried either locked or admitted spurious modes.
The generalized eigenvalue problem is solved with SLEPc's shift-invert eigensolver backed by a MUMPS sparse factorization, using the same mass-proportional regularization $\mathbf{K} + \epsilon \, \mathbf{M}$ as the IGA solver.

\section{End-to-End Validation: Details}\label{appendix:e2e}

This section details the end-to-end validation summarized in the main text: simulation on the boundary representation recovered by our pipeline (IGA-based simulation on the reconstruction, with inpainted fields) against simulation on the ground-truth geometry (an independent FEM-based solve on the original source mesh, with exact fields).
Because the two procedures share neither geometry representation, mathematical discretization, nor field provenance, their agreement validates the complete image-to-simulation pipeline rather than any single component.

\subsection{Protocol and Datasets}

We use four standard computer graphics meshes that carry ground truth---Suzanne, Spot, the Stanford bunny (referred to as Bunny), and Armadillo.
For each, we (i) render a multi-view synthetic dataset from the source mesh with known camera poses, (ii) reconstruct the spline, (iii) inpaint the temperature and material fields exactly as for the real-world datasets, (iv) solve the heat and shell-modal problems on the recovered boundary representation using our IGA solver, and (v) solve the same problems on the source mesh with FEM simulation based on DOLFINx~\cite{barattaDOLFINx2023}.
Since the poses are known, the reconstruction recovers absolute scale; we therefore run at the native ground-truth scale, allowing the energies and frequencies to be compared directly.
The materials are summarized in \cref{tab:e2e_materials}: a two-material aluminum/steel checkerboard (an ${\sim}8\times$ diffusivity contrast) for the heat problem, and a single thin steel shell for the modal problem.

\begin{table}[ht]
    \centering
    \footnotesize
    \caption{
        Material properties for the synthetic end-to-end targets, identical on both solvers.
        Thermal diffusivity is $\alpha = k/(\rho\,c)$; shell thickness $d$ enters as a per-area scaling.
    }
    \label{tab:e2e_materials}
    \begin{tabular}{llccccc}
        \toprule
        Problem               & Role / material    & $k$ [\si{\watt\per\meter\per\kelvin}] & $\rho$ [\si{\kilogram\per\meter\cubed}] & $c$ [\si{\joule\per\kilogram\per\kelvin}] & $d$ [\si{\meter}] & $\alpha$ [\si{\meter\squared\per\second}] \\
        \midrule
        \multirow{2}{*}{Heat} & active / aluminum  & 237                                   & 2700                                    & 900                                       & 0.01              & $9.75\times10^{-5}$                       \\
                              & passive / steel    & 50                                    & 7850                                    & 500                                       & 0.01              & $1.27\times10^{-5}$                       \\
        \midrule
                              &                    & $E$ [\si{\giga\pascal}]               & $\nu$                                   & $\rho$ [\si{\kilogram\per\meter\cubed}]   & $d$ [\si{\meter}] & $\kappa_s$                                \\
        \cmidrule(lr){3-7}
        Modal                 & steel shell        & 210                                   & 0.30                                    & 7850                                      & 0.002             & $5/6$                                     \\
        \bottomrule
    \end{tabular}
\end{table}

\subsection{Ground-Truth Mesh Preparation}

For a reliable simulation comparison, the ground truth meshes must be closed manifolds with an appropriate tessellation.
Two of the source meshes are not directly usable: Suzanne ships as a head shell plus two detached eyeball caps that are recessed in their sockets (so the raw mesh is non-manifold and self-intersecting), and the Bunny has open holes on its base whose interior is never observed.
We weld duplicated seam vertices, drop detached interior bodies that lie inside the hull (the eyeballs, invisible to the reconstruction), cap the resulting openings, and smooth the caps into a natural lid; the eyelid and brow detail on the shell corresponding to the head are retained.
Armadillo and Spot are already closed manifold meshes and are thus left unchanged.
Crucially, this closure is performed before texturing, so that both the rendered images and the ground-truth field are defined on the same closed surface; remeshing for element quality is applied only at the simulation stage.
We verified that the closure changes Suzanne's surface area by less than $0.1\%$, confirming that the repair reproduces the recoverable geometry and does not bias the comparison.
Closing the holes in the Bunny model is very important as the base carries no valid ground-truth field and thus inflates the apparent field discrepancy from $1.1\%$ to $7.6\%$, introducing a clear data artifact.
Each ground-truth mesh is isotropically remeshed for element quality (worst-case equiangular skew below $0.6$; target edge length $0.012$ \si{\meter} for the heat solve and $0.025$ \si{\meter} for the modal solve) before the FEM solve.

\subsection{Field Provenance}

The heat problem requires an initial temperature field, while a two-material partition and a heat source are added to make it more challenging.
For the ground-truth solve, these are assigned exactly on the source mesh; while for the IGA simulation they are inpainted onto the spline from the rendered single-channel images, as described in the main-text (Section 4.3).
The temperature pattern is either obtained from a checkerboard pattern sampled in UV space for the UV-carrying mesh (Spot) or an equivalent checkerboard defined directly in world space for the meshes without UV coordinates (Suzanne, Bunny, Armadillo).
The thermal images are obtained by rendering this reference field.
This allows us to directly compare the sharp, exact field on the ground-truth mesh with the smooth recovered field on the spline.

\subsection{Reference Solver and Time Integration}
\label{appendix:refsolverandtimeintegration}

The ground-truth heat solve uses linear (P1) Lagrange elements while the ground-truth shell-modal solve uses a geometrically-exact Reissner--Mindlin formulation with cubic (P3) fields and an assumed-strain (B-bar) treatment of the transverse-shear and membrane terms to suppress locking, matching the (near) locking-free behavior of our IGA discretization.
A convergence study (varying mesh resolution and element degree) confirmed that P3 has converged; lower-order B-bar elements retained a few percent of spurious stiffness at our shell thickness.
On the IGA side, the same study revealed a residual membrane-locking tail at $64$ knot spans per direction on the thinnest configurations.
This vanishes very quickly under refinement, being removed entirely at $128$ spans per direction.
Thus, all IGA modal simulations use $128$ spans.
The heat time step is derived once from the ground-truth system's natural relaxation time and reused by both solvers, allowing for direct comparison across time steps.

\subsection{Solver Validation on Identical Geometry}
\label{suppmat:solvervalidation}

Before comparing across different geometries, we verify that the two solvers agree when the geometry is held fixed, so that the end-to-end discrepancies can be attributed to geometry recovery rather than to solver disagreement.

We validated our implementation of RM using a thin steel sphere ($E = 193.05\;\si{\giga\pascal}$, $\nu = 0.28$, $\rho = 8025.9\;\si{\kilogram\per\cubic\meter}$, $d = 1.5875\;\si{\milli\meter}$, $R = 0.1135\;\si{\meter}$), by comparing the results against established theoretical and experimental eigenfrequencies~\cite{liuVibrationKLShells2022}.
At $32 \times 32$ knot spans per patch, the implementation accurately captures the six degenerate rigid-body modes, which are cleanly isolated from the elastic spectrum: their eigenvalues lie roughly 11 orders of magnitude below that of the first elastic mode.
The sphere geometry is approximated by 6 B-spline patches with near $G^1$ continuity along the edges ($\approx 10^{-5}$ $L^2$ error), so only a small amount of geometric modelling error is present.

On the eight reconstructed scenes, running both our IGA solvers and DOLFINx ones on the recovered spline geometry, the two independent codes agree to within $0.3\%$ in the heat field and $0.5\%$ in the elastic frequencies (\cref{tab:e2e_solver_agreement}).
In case of heat simulation, each patch is tessellated with $512\times512\times2$ P1 triangles, while for modal simulation $64\times64\times2$ curved elements (geometric order $2$, evaluated from the spline) with P3 solution fields are used.

\begin{table}[ht]
    \centering
    \footnotesize
    \caption{Solver agreement on identical geometry: our IGA solver versus DOLFINx on the \emph{same} reconstructed spline, over all eight reconstructed scenes. Heat is the area-weighted relative $L^2$ of the temperature field (transient mean); modal is the mean relative error over the elastic frequencies.}
    \label{tab:e2e_solver_agreement}
    \begin{tabular}{lcc}
        \toprule
        Scene             & Heat field $L^2$ & Modal freq. (mean) \\
        \midrule
        Woodshed          & $0.025\%$        & $0.348\%$          \\
        Lion              & $0.044\%$        & $0.482\%$          \\
        Building~A Spring & $0.157\%$        & $0.360\%$          \\
        Car               & $0.296\%$        & $0.154\%$          \\
        BuildNet-1        & $0.163\%$        & $0.019\%$          \\
        BuildNet-2        & $0.203\%$        & $0.004\%$          \\
        BuildNet-3        & $0.108\%$        & $0.009\%$          \\
        BuildNet-4        & $0.194\%$        & $0.029\%$          \\
        \bottomrule
    \end{tabular}
\end{table}

The residuals observed in \cref{tab:e2e_solver_agreement} are an order of magnitude below the end-to-end discrepancies reported next, confirming that the latter cannot be accounted for by differences in numerical methods.

\subsection{End-to-End Heat Results}
\label{sup_mat:endtoendheat}

We omit the results presented in the main text concerning the relative heat errors at the initial and final time, heat generation rate, and the first operator eigenvalue.
However, we provide additional detail on how we recovered the difference in spatial evolution.

To probe the accuracy of field evolution, to which the conserved total energy is blind, we compute the area-weighted relative $L^2$ discrepancy between the two temperature fields, transferred across the surfaces by symmetric closest-point projection (\cref{tab:e2e_field_l2}).
The raw cross-field $L^2$ of the near-binary initial checkerboard is dominated by an edge-transfer artifact: a small offset present between meshes flips edge vertices $0\!\leftrightarrow\!1$, injecting large errors along every pattern boundary.
We isolate this by reporting the floor---each field's own symmetric round-trip through the other representation---and the floor-corrected excess, defined as $\sqrt{\mathrm{cross}^2 - \mathrm{floor}^2}$, representing the genuine discrepancy.
The round-trip alone causes a $7$--$24\%$ discrepancy on the sharp $t_0$ field, while the recovered hot-area fraction matches the ground truth to within $2\%$, proving that the inpainting recovers the pattern well and the raw cross-field metric is overly pessimistic.

This is further verified by the excess at the final time step ($t_N$), where we observe the floor has mostly disappeared as the field is significantly smoother after $1000$ diffusion steps.
This gives us an error of $1.1$--$6.9\%$.
Observe that both the initial and final errors are closely related to reconstruction quality.

\begin{table}[ht]
    \centering
    \footnotesize
    \setlength{\tabcolsep}{5pt}
    \caption{Cross-geometry temperature-field discrepancy (relative $L^2$, percent). ``Hot frac.'' is the mismatch in recovered hot-area coverage (artifact-immune). At each time we report the raw cross-field $L^2$, the transfer floor, and the floor-corrected excess; the diffused ($t_N$) excess is the genuine field discrepancy.}
    \label{tab:e2e_field_l2}
    \begin{tabular}{lccccccc}
        \toprule
                  &           & \multicolumn{3}{c}{Initial ($t_0$)} & \multicolumn{2}{c}{Final ($t_N$)}                                    \\
        \cmidrule(lr){3-5}\cmidrule(lr){6-7}
        Mesh      & Hot frac. & cross                               & floor                             & excess & cross & \textbf{excess} \\
        \midrule
        Suzanne   & 0.2       & 7.51                                & 7.25                              & 1.97   & 1.80  & \textbf{1.41}   \\
        Bunny     & 0.0       & 11.12                               & 8.68                              & 6.95   & 1.27  & \textbf{1.07}   \\
        Spot      & 0.3       & 17.84                               & 16.40                             & 7.02   & 4.36  & \textbf{3.35}   \\
        Armadillo & 1.7       & 27.89                               & 23.91                             & 14.36  & 7.06  & \textbf{6.89}   \\
        \bottomrule
    \end{tabular}
\end{table}

\subsection{End-to-End Modal Results}

\Cref{tab:e2e_modes} lists all fourteen elastic eigenfrequencies for each mesh, ground truth versus recovered, at $128$ IGA spans.
The fundamental mode is the least geometry-sensitive, yet it shows a large discrepancy, especially for Spot, whose thin legs are poorly approximated by the sparse views.
Higher panel-bending modes agree more closely in their frequency, but not necessarily in their shape.
Near-degenerate modes reorder between the two solvers, so a handful of per-mode errors are index-pairing artifacts rather than genuine frequency error; we therefore report the median over the fourteen modes in the main text, together with the mean best modal-assurance criterion (MAC), which quantifies the similarity of the modal deformation.

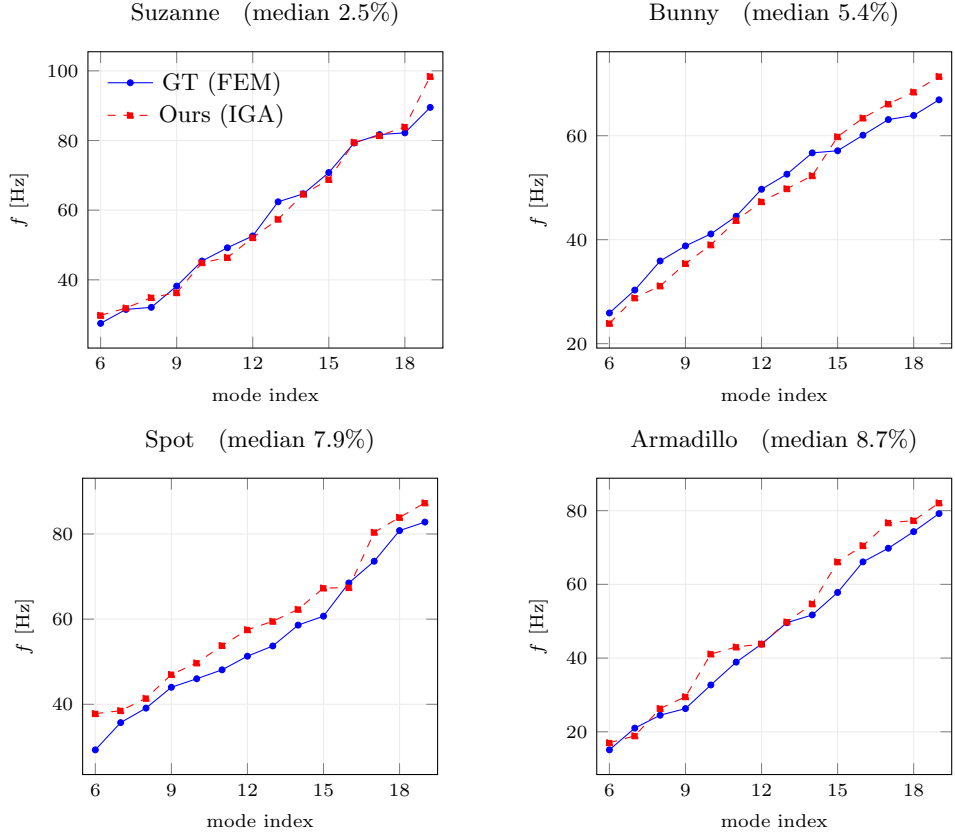
\begin{figure}[ht]
    \centering
    \begin{subfigure}{0.48\textwidth}
        \centering
        \begin{tikzpicture}
            \begin{axis}[width=\linewidth, height=5.5cm,
                    xmin=5.5, xmax=19.5, xtick={6,9,12,15,18},
                    xlabel={mode index}, ylabel={$f$ [\si{\hertz}]},
                    title={Suzanne \; (median $2.5\%$)}, title style={font=\small},
                    label style={font=\footnotesize}, tick label style={font=\footnotesize},
                    grid=both, grid style={gray!15},
                    legend pos=north west, legend style={font=\scriptsize, draw=none, fill=none}]
                \addplot[blue, mark=*, mark size=1.1pt] coordinates {(6,27.5)(7,31.5)(8,32.1)(9,38.2)(10,45.4)(11,49.2)(12,52.6)(13,62.4)(14,64.7)(15,70.8)(16,79.3)(17,81.7)(18,82.2)(19,89.5)}; \addlegendentry{GT (FEM)}
                \addplot[red, dashed, mark=square*, mark size=1.1pt] coordinates {(6,29.8)(7,31.9)(8,34.9)(9,36.3)(10,44.9)(11,46.4)(12,52.1)(13,57.4)(14,64.6)(15,68.8)(16,79.5)(17,81.4)(18,83.9)(19,98.4)}; \addlegendentry{Ours (IGA)}
            \end{axis}
        \end{tikzpicture}
    \end{subfigure}
    \hfill
    \begin{subfigure}{0.48\textwidth}
        \centering
        \begin{tikzpicture}
            \begin{axis}[width=\linewidth, height=5.5cm,
                    xmin=5.5, xmax=19.5, xtick={6,9,12,15,18},
                    xlabel={mode index}, ylabel={$f$ [\si{\hertz}]},
                    title={Bunny \; (median $5.4\%$)}, title style={font=\small},
                    label style={font=\footnotesize}, tick label style={font=\footnotesize},
                    grid=both, grid style={gray!15}]
                \addplot[blue, mark=*, mark size=1.1pt] coordinates {(6,25.9)(7,30.3)(8,35.9)(9,38.8)(10,41.1)(11,44.5)(12,49.7)(13,52.6)(14,56.7)(15,57.1)(16,60.1)(17,63.1)(18,63.9)(19,66.9)};
                \addplot[red, dashed, mark=square*, mark size=1.1pt] coordinates {(6,23.9)(7,28.8)(8,31.1)(9,35.4)(10,39.0)(11,43.7)(12,47.3)(13,49.8)(14,52.3)(15,59.8)(16,63.4)(17,66.1)(18,68.4)(19,71.4)};
            \end{axis}
        \end{tikzpicture}
    \end{subfigure}

    \vspace{0.3em}
    \begin{subfigure}{0.48\textwidth}
        \centering
        \begin{tikzpicture}
            \begin{axis}[width=\linewidth, height=5.5cm,
                    xmin=5.5, xmax=19.5, xtick={6,9,12,15,18},
                    xlabel={mode index}, ylabel={$f$ [\si{\hertz}]},
                    title={Spot \; (median $7.9\%$)}, title style={font=\small},
                    label style={font=\footnotesize}, tick label style={font=\footnotesize},
                    grid=both, grid style={gray!15}]
                \addplot[blue, mark=*, mark size=1.1pt] coordinates {(6,29.3)(7,35.7)(8,39.1)(9,44.0)(10,46.0)(11,48.1)(12,51.3)(13,53.7)(14,58.6)(15,60.7)(16,68.5)(17,73.6)(18,80.8)(19,82.8)};
                \addplot[red, dashed, mark=square*, mark size=1.1pt] coordinates {(6,37.8)(7,38.5)(8,41.4)(9,47.0)(10,49.7)(11,53.8)(12,57.5)(13,59.5)(14,62.3)(15,67.3)(16,67.4)(17,80.4)(18,83.9)(19,87.3)};
            \end{axis}
        \end{tikzpicture}
    \end{subfigure}
    \hfill
    \begin{subfigure}{0.48\textwidth}
        \centering
        \begin{tikzpicture}
            \begin{axis}[width=\linewidth, height=5.5cm,
                    xmin=5.5, xmax=19.5, xtick={6,9,12,15,18},
                    xlabel={mode index}, ylabel={$f$ [\si{\hertz}]},
                    title={Armadillo \; (median $8.7\%$)}, title style={font=\small},
                    label style={font=\footnotesize}, tick label style={font=\footnotesize},
                    grid=both, grid style={gray!15}]
                \addplot[blue, mark=*, mark size=1.1pt] coordinates {(6,15.1)(7,21.0)(8,24.5)(9,26.3)(10,32.7)(11,38.9)(12,43.8)(13,49.6)(14,51.7)(15,57.8)(16,66.1)(17,69.8)(18,74.3)(19,79.2)};
                \addplot[red, dashed, mark=square*, mark size=1.1pt] coordinates {(6,17.0)(7,18.9)(8,26.3)(9,29.5)(10,41.1)(11,43.0)(12,43.8)(13,49.8)(14,54.7)(15,66.1)(16,70.5)(17,76.7)(18,77.3)(19,82.1)};
            \end{axis}
        \end{tikzpicture}
    \end{subfigure}
    \caption{Recovered (IGA, red dashed) versus ground-truth (FEM, blue solid) elastic eigenspectra: the fourteen elastic frequencies against mode index (the paper convention, in which modes $0$--$5$ are rigid-body); the per-panel median relative error is given in each title. The two spectra track closely for every mesh, with the agreement tightening for the better-reconstructed shapes (Suzanne, Bunny) and loosening for the hardest ones (Spot, Armadillo).}
    \label{fig:e2e_parity}
\end{figure}

\Cref{fig:e2e_parity} overlays the recovered and ground-truth eigenspectra for the four meshes; the two track closely throughout.
The per-mode median error grows monotonically with reconstruction difficulty---$2.5\%$ (Suzanne), $5.4\%$ (Bunny), $7.9\%$ (Spot), $8.7\%$ (Armadillo)---mirroring the geometry error and the declining mode-shape MAC, and confirming that the residual spectral discrepancy is set by reconstruction quality rather than by the solver.

\begin{table}[ht]
    \centering
    \footnotesize
    \setlength{\tabcolsep}{5pt}
    \caption{
        The fourteen elastic eigenfrequencies [\si{\hertz}], ground truth (GT, FEM on the source mesh) versus ours (IGA on the recovered spline, $128$ spans).
        The six rigid-body modes are excluded; all runs use native ground-truth scale.
    }
    \label{tab:e2e_modes}
    \begin{tabular}{c cc cc cc cc}
        \toprule
               & \multicolumn{2}{c}{Suzanne} & \multicolumn{2}{c}{Bunny}   & \multicolumn{2}{c}{Spot}    & \multicolumn{2}{c}{Armadillo}                             \\
        \cmidrule(lr){2-3}\cmidrule(lr){4-5}\cmidrule(lr){6-7}\cmidrule(lr){8-9}
        Mode   & GT                          & Ours                        & GT                          & Ours                          & GT   & Ours & GT   & Ours \\
        \midrule
        6      & 27.5                        & 29.8                        & 25.9                        & 23.9                          & 29.3 & 37.8 & 15.1 & 17.0 \\
        7      & 31.5                        & 31.9                        & 30.3                        & 28.8                          & 35.7 & 38.5 & 21.0 & 18.9 \\
        8      & 32.1                        & 34.9                        & 35.9                        & 31.1                          & 39.1 & 41.4 & 24.5 & 26.3 \\
        9      & 38.2                        & 36.3                        & 38.8                        & 35.4                          & 44.0 & 47.0 & 26.3 & 29.5 \\
        10     & 45.4                        & 44.9                        & 41.1                        & 39.0                          & 46.0 & 49.7 & 32.7 & 41.1 \\
        11     & 49.2                        & 46.4                        & 44.5                        & 43.7                          & 48.1 & 53.8 & 38.9 & 43.0 \\
        12     & 52.6                        & 52.1                        & 49.7                        & 47.3                          & 51.3 & 57.5 & 43.8 & 43.8 \\
        13     & 62.4                        & 57.4                        & 52.6                        & 49.8                          & 53.7 & 59.5 & 49.6 & 49.8 \\
        14     & 64.7                        & 64.6                        & 56.7                        & 52.3                          & 58.6 & 62.3 & 51.7 & 54.7 \\
        15     & 70.8                        & 68.8                        & 57.1                        & 59.8                          & 60.7 & 67.3 & 57.8 & 66.1 \\
        16     & 79.3                        & 79.5                        & 60.1                        & 63.4                          & 68.5 & 67.4 & 66.1 & 70.5 \\
        17     & 81.7                        & 81.4                        & 63.1                        & 66.1                          & 73.6 & 80.4 & 69.8 & 76.7 \\
        18     & 82.2                        & 83.9                        & 63.9                        & 68.4                          & 80.8 & 83.9 & 74.3 & 77.3 \\
        19     & 89.5                        & 98.4                        & 66.9                        & 71.4                          & 82.8 & 87.3 & 79.2 & 82.1 \\
        \midrule
        median & \multicolumn{2}{c}{$2.5\%$} & \multicolumn{2}{c}{$5.4\%$} & \multicolumn{2}{c}{$7.9\%$} & \multicolumn{2}{c}{$8.7\%$}                               \\
        MAC    & \multicolumn{2}{c}{$0.53$}  & \multicolumn{2}{c}{$0.43$}  & \multicolumn{2}{c}{$0.34$}  & \multicolumn{2}{c}{$0.32$}                                \\
        \bottomrule
    \end{tabular}
\end{table}

\subsection{Geometry Error and the Reconstruction--Simulation Correlation}

\begin{table}[ht]
    \centering
    \footnotesize
    \caption{
        Geometric reconstruction error: recovered tessellation versus co-registered ground-truth surface, as distances in percent of the ground-truth bounding-box diagonal, except for the area ratio
    }
    \label{tab:e2e_geometry}
    \begin{tabular}{lccccc}
        \toprule
        Mesh      & Chamfer & Hausdorff & Mean surf. & RMS surf. & Area ratio \\
        \midrule
        Suzanne   & 0.207   & 3.112     & 0.106      & 0.338     & 1.0014     \\
        Bunny     & 0.273   & 2.131     & 0.131      & 0.326     & 0.9907     \\
        Spot      & 0.473   & 5.226     & 0.214      & 0.605     & 0.9859     \\
        Armadillo & 0.722   & 7.192     & 0.363      & 0.829     & 0.9893     \\
        \bottomrule
    \end{tabular}
\end{table}

Finally, \cref{tab:e2e_geometry} quantifies the geometric reconstruction error that drives the errors discussed above.
They are computed from distances between the rendering tessellation of the reconstructed boundary representation and the corresponding ground-truth mesh, and given in percentages of the bounding-box diagonal of the ground truth mesh.
Ordering the four meshes by their mean surface distance, the heat initial-energy error, the $\lambda_1$ error, the field excess at initial time, and the median modal-frequency error all increase monotonically with geometry error---the exceptions being the generation rate and the field excess at final time---while the mode-shape agreement (MAC) decreases monotonically with geometry error.
With only four meshes these results are indicative rather than statistically conclusive, but the consistent trend is direct evidence that reconstruction quality governs simulation fidelity to a degree that cannot be explained by solver differences alone.

\section{Initial Pose Optimization}\label{supp_mat:initial_pose}

The shape optimization procedure requires a reasonable initial guess for the shape's pose and scale: image-space gradients propagate poorly through silhouette boundaries when the proxy is far from the target, and downstream shape optimization cannot easily translate vertices or control points across long distances.
To produce such a guess, we initialize from a cubed-sphere B-spline proxy, apply pure-normal regularization to smooth out the patch seams of the cubed-sphere parametrization, and rescale the result so that its extent along each axis equals \SI{2}{\meter} (a unit-radius sphere).
The remainder of this section describes how the translation, isotropic scale, and orientation of this proxy are determined from the input views.

\subsection{Initial Position: Camera Convergence Point}

\par The initial translation is set to the \emph{convergence point} of the input cameras --- the point that minimizes the sum of squared distances to the optical axes of all views.
For scenes where the cameras converge on the object (the common case in object-centric capture), this yields a position close to the object centroid.
Performance is expected to degrade if the cameras are close to parallel, in which case the convergence point is poorly conditioned.

\par Let each camera $i \in \{1, \dots, N\}$ be defined by its world-space origin $\mathbf{p}_i \in \mathbb{R}^3$ and unit forward direction $\mathbf{d}_i \in \mathbb{R}^3$.
The orthogonal projection onto the plane normal to $\mathbf{d}_i$ is
\begin{equation}
    S_i = I - \mathbf{d}_i \mathbf{d}_i^\top,
\end{equation}
where $I$ is the $3 \times 3$ identity, and the squared distance from a point $\mathbf{c}$ to the line through $\mathbf{p}_i$ with direction $\mathbf{d}_i$ is $\| S_i (\mathbf{c} - \mathbf{p}_i) \|^2$.
The convergence point is therefore the minimizer of
\begin{equation}
    \mathbf{c}^\star = \argmin_{\mathbf{c} \in \mathbb{R}^3} \sum_{i=1}^{N} \| S_i \mathbf{c} - S_i \mathbf{p}_i \|^2,
\end{equation}
which is the ordinary least-squares solution of the overdetermined linear system $A \mathbf{c} = \mathbf{b}$ with
\begin{equation}
    A =
    \begin{pmatrix}
        S_1 \\
        S_2 \\
        \vdots \\
        S_N
    \end{pmatrix}
    \in \mathbb{R}^{3N \times 3},
    \qquad
    \mathbf{b} =
    \begin{pmatrix}
        S_1 \mathbf{p}_1 \\
        S_2 \mathbf{p}_2 \\
        \vdots \\
        S_N \mathbf{p}_N
    \end{pmatrix}
    \in \mathbb{R}^{3N}.
\end{equation}
The closed-form solution $\mathbf{c}^\star = (A^\top A)^{-1} A^\top \mathbf{b}$ is computed once, without rendering, and used as the initial translation of the proxy.

\subsection{Initial Scale: Golden-Section Silhouette Search}

\par Once the proxy is placed at $\mathbf{c}^\star$, its initial isotropic scale is selected by minimizing a silhouette loss in image space.
The proxy is rendered as a black diffuse surface against a constant white emitter under Mitsuba's \texttt{mono} variant, so the rendered image is a single-channel luminance and the loss against the reference masks reduces to a pure silhouette match.
The search is restricted to the bracket $[c_{\min} r_{\min}, c_{\max} r_{\min}]$, where $r_{\min}$ is the smallest distance from $\mathbf{c}^\star$ to any camera origin and $0 < c_{\min} < c_{\max} < 1$ are safety constants ($c_{\min} = 0.01$, $c_{\max} = 0.9$ in our implementation); the upper bound prevents the proxy from starting tangent to a camera, where the silhouette gradient degenerates.

\par The search proceeds in two phases.
A coarse linear sweep first evaluates the loss at a small number of evenly-spaced radii across the bracket, guarding against degenerate camera arrangements (e.g.\ ring rigs) in which the silhouette loss is locally flat and a pure golden-section search would converge to a meaningless point in the flat region.
A golden-section refinement is then run inside a two-step neighborhood of the best coarse sample, contracting the bracket until its width falls below $1\%$ of the original upper bound.
The midpoint of the final bracket is taken as the initial scale.

\subsection{Pose Refinement by Differentiable Silhouette Matching}

\par The translation and scale obtained above place the proxy in a region where the silhouette loss is differentiable with respect to all pose parameters.
A final refinement step then optimizes nine parameters jointly---a translation $\boldsymbol{t} \in \mathbb{R}^3$, a per-axis scaling $\boldsymbol{s} \in \mathbb{R}^3$, and Euler angles $\boldsymbol{\theta} \in \mathbb{R}^3$---combined into the standard TRS transform $T(\boldsymbol{t}) \, R(\boldsymbol{\theta}) \, S(\boldsymbol{s})$ applied to the proxy's global control points.
A small seeded Gaussian jitter is added to the initial translation and per-axis scale to break the perfect symmetry of the cubed-sphere proxy, which otherwise tends to trap the optimizer in saddle points.
The objective is the silhouette $L^1$ loss summed over all training views, and the parameters are optimized with L-BFGS using the differentiable Mitsuba renderer to backpropagate image-space gradients through the rasterized proxy.
The optimized $(\boldsymbol{t}, \boldsymbol{s}, \boldsymbol{\theta})$ define the bounding-box transform that initializes the subsequent shape and texture optimization.

\end{document}